\pdfoutput=1

\documentclass[11pt]{article}

\usepackage{acl}

\usepackage{times}
\usepackage{latexsym}
\usepackage{times}
\usepackage{latexsym}
\usepackage[figuresright]{rotating}
\usepackage{amsmath}
\usepackage{xcolor}
\usepackage[T1]{fontenc}

\usepackage[utf8]{inputenc}
\usepackage{microtype}
\usepackage{inconsolata}
\usepackage{booktabs}
\usepackage{graphicx}
\usepackage{subcaption}
\usepackage{multirow}
\usepackage{bm}
\usepackage{amsthm,amsmath,amssymb}
\usepackage{color}
\usepackage{colortbl}
\usepackage{tcolorbox}
\usepackage{xcolor}
\usepackage{mathrsfs}
\usepackage{CJKutf8}
\usepackage{hyperref} 
\usepackage{ruby}
\usepackage{xpinyin}

\usepackage[T1]{fontenc}

\usepackage[utf8]{inputenc}

\usepackage{microtype}
\definecolor{lightred}{HTML}{f4dee0} 
\definecolor{lightgreen}{HTML}{ecf4eb}

\newcommand{\mycustomsize}{\fontsize{10}{11}\selectfont}
\newcommand{\exafontsize}{\fontsize{10}{11}\selectfont}
\newtcbox{\red}[1][]{on line,boxsep=0.5pt,left=0pt,right=0pt,top=0pt,bottom=0pt,colframe=white,colback=red!25!white,#1}

\title{Does Mapo Tofu Contain Coffee?\\ Probing LLMs for Food-related Cultural Knowledge}

\newcommand{\cuhksz}{$^1$}
\newcommand{\sribd}{$^2$}
\newcommand{\huji}{$^3$}
\newcommand{\uestc}{$^4$}

\newcommand{\ku}{$^5$}
\newcommand{\tubingen}{$^6$}

\author{Li Zhou\cuhksz$^,$\sribd, Taelin Karidi\huji, \textbf{Wanlong Liu}\uestc, Nicolas Garneau\ku, Yong Cao\tubingen,\\ \textbf{Wenyu Chen}\uestc, \textbf{Haizhou Li}\cuhksz$^,$\sribd, \textbf{Daniel Hershcovich}\ku \\
{\cuhksz}The Chinese University of Hong Kong, Shenzhen
{\sribd}Shenzhen Research Institute of Big Data  \\
{\huji}Hebrew University of Jerusalem
{\uestc}University of Electronic Science and Technology of China, 
\\
{\ku}University of Copenhagen, 
{\tubingen}University of Tübingen, Tübingen AI Center \\
\texttt{\small \{lizhou21, haizhouli\}@cuhk.edu.cn}, 
\texttt{\small taelin.karidi@mail.huji.ac.il}, 
\texttt{\small liuwanlong@std.uestc.edu.cn},\\
\texttt{\small \{nicolas.garneau, dh\}@di.ku.dk},
\texttt{\small yongcao2018@gmail.com},
\texttt{\small cwy@uestc.edu.cn}
}

\begin{document}
\maketitle
\begin{abstract}
Recent studies have highlighted the presence of cultural biases in Large Language Models (LLMs), yet often lack a robust methodology to dissect these phenomena comprehensively. Our work aims to bridge this gap by delving into the \textsc{Food} domain—a universally relevant yet culturally diverse aspect of human life. We introduce \textsc{FmLAMA}, a multilingual dataset centered on food-related cultural facts and variations in food practices. 
We analyze LLMs across various architectures and configurations, evaluating their performance in both monolingual and multilingual settings. 
By leveraging templates in six different languages, we investigate how LLMs interact with language-specific and cultural knowledge. 
Our findings reveal that 
(1) LLMs demonstrate a pronounced bias towards food knowledge prevalent in the United States; 
(2) Incorporating relevant cultural context significantly improves LLMs' ability to access cultural knowledge; 
(3) The efficacy of LLMs in capturing cultural nuances is highly dependent on the interplay between the probing language, the specific model architecture, and the cultural context in question.
This research underscores the complexity of integrating cultural understanding into LLMs and emphasizes the importance of culturally diverse datasets to mitigate biases and enhance model performance across different cultural domains.

\end{abstract}

\section{Introduction} 
\label{sec:intro}
\begin{figure}
    \centering
    \includegraphics[width=1\linewidth, trim=0cm 0cm 0cm 0cm, clip]{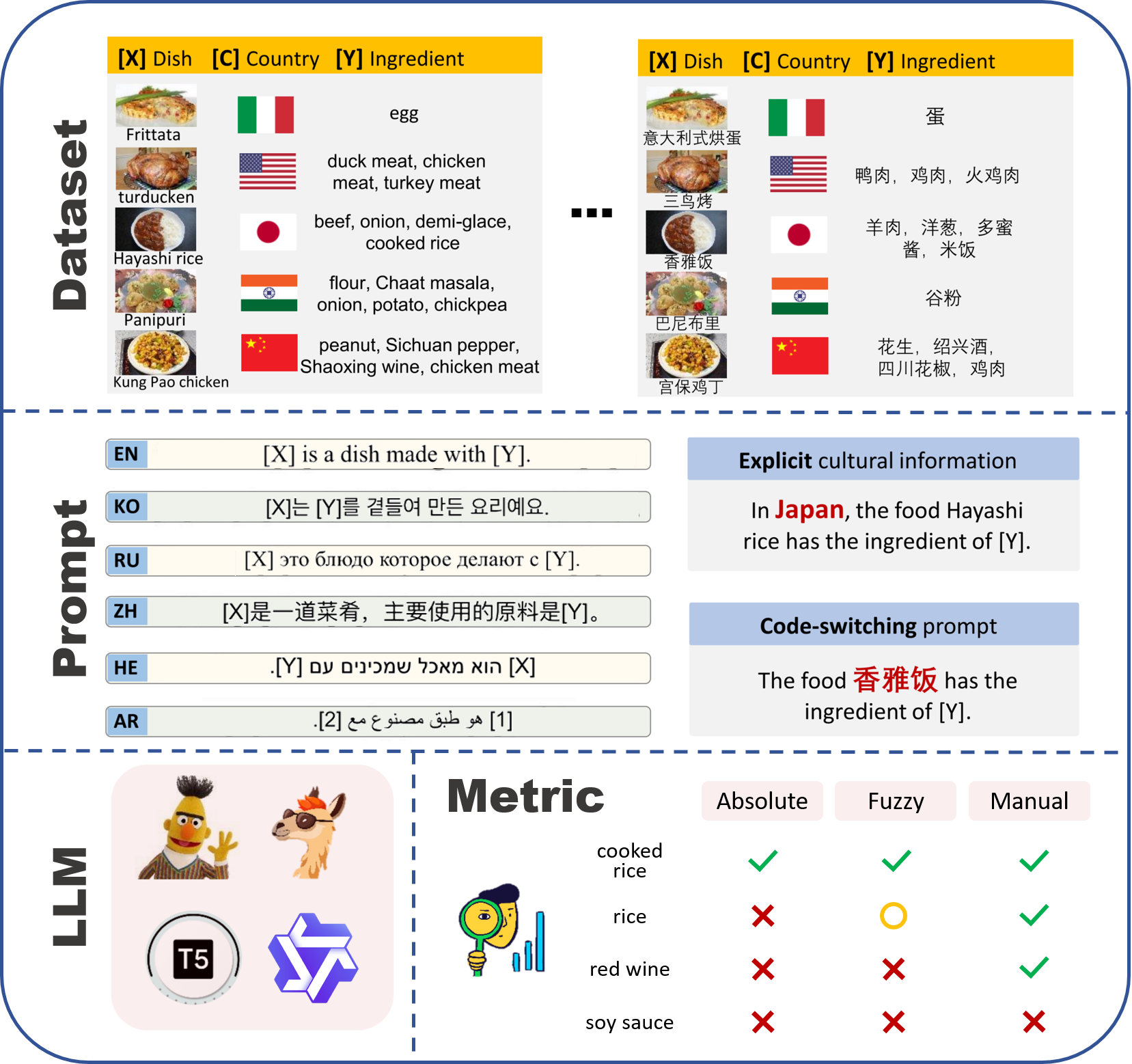}
    \caption{Summary of the various aspects of our work.
    }\label{fig: overview}
\end{figure}

Asking a French person for the recipe of \textit{Beef Bourguignon} in English might yield an immediate and precise response, while the same query might pose challenges to a Chinese individual unless posed as \begin{CJK}{UTF8}{gbsn}\xpinyin*{勃艮第牛肉}\end{CJK} (its literal translation). 
In China, the dish is also commonly referred to by its broader description, \begin{CJK}{UTF8}{gbsn}\xpinyin*{红酒炖牛肉}\end{CJK} (\textit{Red Wine Stewed Beef}), highlighting the main ingredients and cooking technique, albeit without specifying a regional origin. 
Employing \begin{CJK}{UTF8}{gbsn}\xpinyin*{法式红酒炖牛肉}\end{CJK} (\textit{French-style Red Wine Stewed Beef}) with an adjectival description can indicate adherence to French culinary traditions.
This practice illustrates how cultural and linguistic nuances shape knowledge transmission.
In cross-cultural communication, when direct translations are unavailable, speakers often switch between languages—a phenomenon known as code-switching—to better convey meaning~\cite{aguilar-solorio-2020-english, dogruoz-etal-2021-survey}.
This variability underscores the challenges language models encounter in navigating cross-cultural culinary contexts and highlights the difficulty of responding with cultural knowledge, which involves using specific language, engaging in code-switching, and integrating cultural context.

Trained on vast datasets, Large Language Models (LLMs) encode a wide array of knowledge, but also face challenges with various biases, such as those related to gender~\cite{savoldi-etal-2021-gender, kaneko-etal-2022-gender}, belief~\cite{sogaard-2021-lockes, gonzalez-etal-2021-interaction, lent-sogaard-2021-common}, and culture~\cite{cao-etal-2023-assessing, deshpande-etal-2022-stereokg, yin-etal-2022-geomlama, mukherjee-etal-2024-global, singh2024global}. 
In this paper, we adopt food-domain knowledge data as cultural semantic proxy, “country of origin” as cultural demographic proxy~\cite{adilazuarda-etal-2024-towards}, and address the following research questions: (1) What cultural biases exist within the cultural knowledge contained in LLMs? (2) How can we best prompt and evaluate LLMs to elicit culture-specific knowledge? 
Specifically, we first design an automated construction method for the dataset \textsc{FmLAMA}  (\S\ref{sec:fmlama_construction}). Next, we introduce the cultural knowledge probing method (\S\ref{sec:probing_method}), encompassing the probing task, template and metric. We then conduct experiments (\S\ref{sec:experiments}) and analysis (\S\ref{sec:analysis})  to address our proposed research questions. Finally, we perform an error analysis on the probing results (\S\ref{sec:evaluation}).
Figure~\ref{fig: overview} provides an overview of our work's various aspects, and our \textbf{contributions} are:

\begin{itemize}
\item We present \textsc{FmLAMA}, a pioneering dataset focused on the food domain, which is inherently rich in cultural diversity. This dataset is a multifaceted tool for probing LLMs across cultures and languages.
\item 
We propose novel metrics aimed at evaluating LLMs' capacity to accurately and sensitively probe cultural knowledge, utilizing a combination of automatic and manual evaluation methods.
\item We analyze the impact of integrating cultural context and language specificity in prompts, offering insights to optimize LLMs for equitable cross-cultural knowledge retrieval.
\end{itemize}
Our methodology for automated collection of cultural knowledge corpora 
extends the analysis potential in other domains, broadening the scope of research on cultural biases in LLMs.
\footnote{Data and code are available in \url{https://github.com/lizhou21/FmLAMA-master}.}

\section{Related Work}
\paragraph{Cultural knowledge datasets.}
Cultural knowledge, encompassing the customs, beliefs, traditions, and practices of a culture, is crucial yet challenging to encapsulate. While some researchers focus on manually curating cultural knowledge datasets, others evaluate LLMs' performance on culturally related tasks. \citet{yin-etal-2022-geomlama} and \citet{palta-rudinger-2023-fork} have developed benchmarks such as geo-diverse prompts and food-custom datasets (FORK) to probe cultural biases in commonsense reasoning systems. However, manual dataset construction is inefficient and hard to scale, prompting a shift towards automated methods. For instance, StereoKG~\cite{deshpande-etal-2022-stereokg} offers a scalable knowledge graph that blends cultural knowledge with stereotypes, and \textsc{Candle}~\cite{nguyen2023extracting} extracts cultural commonsense knowledge from the web, organizing it into clusters. Despite these advances, the variability in data representation—from sentences to triplets using \texttt{OpenIE}—poses challenges for consistency and noise control in knowledge probing. \citet{keleg-magdy-2023-dlama} aims to mitigate this by selecting culturally diverse factual triples from Wikidata, focusing mainly on explicit country information.
In contrast, our work proposes an automated, efficient approach to constructing a cultural knowledge dataset in a uniform triplet format, addressing the limitations of existing methods and focusing on implicit cultural knowledge. Appendix~\ref{app-sec:example} shows a comparison of our dataset with existing ones.

%




\paragraph{Knowledge probing.}
Deciphering the knowledge encoded by LLMs poses significant challenges due to their opaque nature , early benchmarks like LAMA \cite{petroni-etal-2019-language} sought to quantify the factual knowledge in English LLMs, while ParaRel \cite{elazar-etal-2021-measuring} highlighted their consistency issues. Subsequent efforts as mLAMA~\cite{kassner-etal-2021-multilingual} and mParaRel~\cite{fierro-sogaard-2022-factual} expanded these benchmarks multilingually, though such methods often focus on single-word entities, limiting their depth of assessment. To address these shortcomings, newer studies~\cite{shin-etal-2020-autoprompt, zhong-etal-2021-factual, meng-etal-2022-rewire} have evolved towards eliciting more comprehensive factual knowledge, including multi-word entities, with \citet{jiang-etal-2020-x} developing algorithms for multi-token predictions. LPAQA~\cite{10.1162/tacl_a_00324} further refines this by optimizing prompt discovery for more accurate knowledge probing. Our work builds on this foundation, targeting multi-token probing within the food domain, characterized by complex expressions like \textit{Trigonella foenum-graecum}. 











\section{\textsc{FmLAMA} Construction}
\label{sec:fmlama_construction}
To assess whether LLMs encode and access cultural information, we develop \textsc{FmLAMA}, a multicultural, multilingual dataset focusing on culinary knowledge. 
The designed framework can be adapted to other cultural domains.


\paragraph{Step \#1: Obtain countries set.} Following \citet{zhou-etal-2023-cultural}, we use countries of food origin to delineate cultural groups. This method leverages countries as proxies for cultural identity, encapsulating diverse traditions, values, and norms that reflect the breadth of human civilizations across geographical boundaries~\cite{minkov2012national, peterson2018traversing}.


\paragraph{Step \#2: Acquire food instances.} 
We utilize SPARQL to query Wikidata, extracting a vast array of food-related data. This approach exploits Wikidata's RDF triple structure to gather detailed information on food instances, offering a rich source of comprehensive food knowledge.

\textbf{i. Class.} For our food-focused dataset, we concentrate on the \href{https://www.wikidata.org/wiki/Q746549}{dish} class and employ two approaches to find food instances:
\begin{itemize}
{\exafontsize
\item Explicit instance of \href{https://www.wikidata.org/wiki/Q746549}{dish}, e.g., \href{https://www.wikidata.org/wiki/Q828043}{\textit{bouillabaisse}}.
\item Inferred through a hierarchy, e.g., {\href{https://www.wikidata.org/wiki/Q1191278}{\textit{Blanquette de veau}}
$\xrightarrow{\mathrm{\text{subclass of}}}$\href{https://www.wikidata.org/wiki/Q2920963}{stew}$\xrightarrow{\mathrm{\text{subclass of}}}$\href{https://www.wikidata.org/wiki/Q746549}{\textit{dish}}}.}
\end{itemize}
This enables comprehensive inclusion of food instances, represented as $I \xrightarrow{(\text{instance of} | \text{subclass of})+}$ \href{https://www.wikidata.org/wiki/Q746549}{\textit{dish}}, where `$|$' denotes ``or'', and `$+$' is ``one or more''.


\textbf{ii. Cultural group.} We organize food instances by their origin, applying these strategies:
\begin{itemize}
{\exafontsize
\item Directly specified in Wikidata, e.g., \href{https://www.wikidata.org/wiki/Q828043}{\textit{bouillabaisse}} $\xrightarrow{\text{country of origin}}$ \href{https://www.wikidata.org/wiki/Q142}{France}.
\item Through the associated cuisine category, e.g., \href{https://www.wikidata.org/wiki/Q1298577}{\textit{mapo doufu}} $\xrightarrow{\text{cuisine}}$ \href{https://www.wikidata.org/wiki/Q10876842}{Chinese cuisine} $\xrightarrow{\text{country}}$ \href{https://www.wikidata.org/wiki/Q148}{China}.}
\end{itemize}
We exclude dishes with multiple origin countries to maintain cultural specificity.

\textbf{iii. Properties included.} The property ``\href{https://www.wikidata.org/wiki/Property:P527}{has part(s)}'' identifies food ingredients for a dish. Additional properties like ``\href{https://www.wikidata.org/wiki/Property:P186}{made from material}'' and ``\href{https://www.wikidata.org/wiki/Property:P18}{image}'' are collected to support future research (e.g., multimodal), though they are not utilized in this study. 
In each instance, the descriptive language for all attributes remains consistent.

Ultimately, our constructed dataset, \textsc{FmLAMA}, comprises 33,600 dish instances, detailed by name, origin, ingredients, and optionally, materials and images, encompassing 128 cultural country groups and 250 languages. The average number of ingredients in \textsc{FmLAMA} dataset is 2.04. Examples and statistics are provided in Appendix~\ref{app:dataset}.

\section{Cultural Knowledge Probing} 
\label{sec:probing_method}

\subsection{Probing task}
\label{sec:probing_tasks}

We adopt Word Prediction (WP) as the knowledge-probing task, following previous work~\cite{fierro-sogaard-2022-factual, wu-etal-2023-plms, fierro-etal-2024-mulan}.
Firstly, we manually design a prompt template $t$ focused on the core attribute ``has part(s)'', illustrating a connection between a dish (subject) and its ingredient(s) (object).
WP is usually implemented as a candidate retrieval problem.
The candidate set consists of all objects in the focused, filtered sub-dataset of \textsc{FmLAMA}.\footnote{As the size of the filtered sub-dataset increases, the candidate object set also expands, leading to greater difficulty in probing. Consequently, in this paper, results obtained by probing across different filtered sub-datasets cannot be used for horizontal comparison.}
The primary objective is to utilize LLMs to obtain the probability of each candidate $C$ and subsequently rank the predicted objects based on these probabilities.

\paragraph{MASK operation.}
Using subject-object tuples {([X], [Y])} as queries, we probe LLMs by replacing the subject and masking the object.
Considering that each candidate object is tokenized into $d$ subtokens $\left\{ c_1, \cdots, c_d \right\} $ by LLMs correspondingly, we apply [MASK] token of varying lengths to the objects within each query. 
$D$ queries are constructed for each dish subject $X$ based on the same template $t$, $D$ is the maximum number of object tokens, each query $Q_{d}^{X}$ is defined as:
\begin{equation}
    \mycustomsize
    Q_{d}^{X}=t\left( X,\left[ \mathrm{MASK} \right] *d \right) , d\in \left[ 1,D \right]
    \vspace{-3pt}.
    \label{eq:query}
\end{equation}

\paragraph{Probability acquisition.} We use Mean Pooling\footnote{Mean Pooling has been shown to be superior to other pooling methods for multi-token probing~\cite{wu-etal-2023-plms}.}
method to obtain the prediction probability of each candidate. 
Specifically, for candidate object $C=\left\{ c_1, \cdots, c_d \right\} $ of length $d$, its probability is obtained from the likelihoods associated with the [MASK] tokens in $Q_{d}^{X}$, and the predicted probability of $C$ is calculated as the average of the 
probabilities of composing its subtokens:
\begin{equation}
\mycustomsize
    P\left( Q_{d}^{X},C \right) =\frac{1}{d}\sum_{i=1}^d{p\left( Q_{d}^{X},\left[ \mathrm{MASK} \right] _i=c_i \right)}
    \vspace{-3pt},
    \label{eq:subtokens}
\end{equation}
where $p\left( \cdot \right)$ is obtained after the softmax operation.


\subsection{Probing template}
\label{sec:probing_templates}

Considering LLMs produce varied predictions based on prompt framing~\cite{elazar-etal-2021-measuring, wang2023assessing}, we craft \textit{five} templates conveying identical meanings in each prompt language. For example, ``[X] is a dish made with [Y]'', ``[X] is a type of food that includes [Y]''.

\paragraph{Cultural information.}
To explore the impact of introducing explicit cultural context on LLMs' ability to access cultural knowledge, we enhance the basic templates by integrating country information. For instance, ``In [C], [X] is a dish made with [Y]'', where [C] denotes the country of origin for the dish [X], [Y] indicates the ingredient object.

\paragraph{Multilingual prompts.}
To explore how different prompt language settings affect LLMs' cultural knowledge probing abilities, we craft prompts in six languages written by native speakers, including English (\textit{en}), Chinese (\textit{zh}), Arabic (\textit{ar}), Korean (\textit{ko}), Russian (\textit{ru}), and Hebrew (\textit{he}). These languages span 4 different language families -- Indo-European (English, Russian), Semitic (Hebrew, Arabic), Altaic (Korean), and Sino-Tibetian (Chinese), and are spoken by more than 2.356 billion speakers. Furthermore, these languages represent cultural diversity, being spoken on different continents by groups with rich and distinct cultural backgrounds.
These prompt illustrations and language details are depicted in Appendix~\ref{app:templates}.

\paragraph{Code-switching.}
To simulate real-world scenarios where mixed-language expression often occurs, we implement code-switching prompts, varying the main language ($\mathrm{ML}$) and subject language ($\mathrm{SL}$).
Specifically, we define the code-switching prompt setting as $\mathrm{P}(\mathrm{ML}, \mathrm{SL}) $, where $\mathrm{ML}, \mathrm{SL} \in \{en, zh, ar, ko, ru, he\}$, and $\mathrm{ML} \neq \mathrm{SL}$. 
Our objective is to ensure that the language of the predicted object remains consistent with the main language (ML).
For instance, the prompt ``\begin{CJK}{UTF8}{gbsn}\xpinyin*{勃艮第牛肉}\end{CJK} is a dish made with [Y]'' follows the $\mathrm{P}(en,zh)$ setting, with the expected output for [Y] in English.

\subsection{Probing Metric}
\label{sec:probing_metric}
Despite considerable prior work on knowledge probing, even studies employing a similar LAMA-style approach lack a standardized evaluation criterion. Although our experiments solely focus on a single relationship, that is, the ingredients of a food item, our probing task poses greater challenges for LLMs: 
(1) The number of objects in each instance is not fixed,
(2) the number of food instances contained in each cultural group varies,
(3) the verbalization of ingredients is not unique
(e.g., in Chinese, both \begin{CJK}{UTF8}{gbsn}\xpinyin*{盐}\end{CJK} and \begin{CJK}{UTF8}{gbsn}\xpinyin*{食盐}\end{CJK} can denote \textit{salt}), and
(4) ingredients can be flexible (e.g., in the Italian dish \textit{frico}, listed with \textit{cheese} as an ingredient in Wikidata, either \textit{mozzarella} or \textit{feta} are actually valid).
(5) The reference ingredients label in Wikidata is incomplete. 

Given these constraints, we introduce two automatic metrics: the absolute-match metric, Mean Average Precision (mAP), and the fuzzy-match metric, Mean Word Similarity (mWS), along with a Manual Evaluation Score (MES) that incorporates both LLM-simulated and real human assessments for implementation.
All mAP, mWS, and MES metrics are in the range $[0, 1]$.

\paragraph{Mean Average Precision.}
mAP is widely used in information retrieval settings, assessing the relevance of predicted objects (in our case ingredients) only when they precisely match the reference ones. 
The precision at rank $k$ (P@k) for a given food instance $i$ is defined as follows:
\begin{equation}
\mycustomsize
    \mathrm{P@k}={\frac{\left| \mathrm{ing}_i \cap \mathrm{topk}_i \right|}{k}}
    \vspace{-3pt},
    \label{eq:precision_at_k}
\end{equation}
\noindent
where $\mathrm{ing}_i$ is the reference ingredients set, and $\mathrm{topk}_i$ signifies the set of top-$k$ objects with the highest predicted probability of belonging to food item $i$ by LLMs.
Then the average precision of food item $i$ is computed as follows:
\begin{equation}
\mycustomsize
    \mathrm{AP}_i=\frac{1}{\left| \mathrm{ing}_i \right|}\sum_{k=1}^n{\mathrm{P}@\mathrm{k}\times \mathrm{rel}@\mathrm{k}}
    \vspace{-3pt},
\end{equation}
\noindent
where $n$ refers to the size of the candidate object set, and $\mathrm{rel}@\mathrm{k}$ is 1 if the object at rank $k$ is relevant to food item $i$, otherwise 0.
Finally, we compute the mAP in the following way:
\begin{equation}
\mycustomsize
    \mathrm{mAP}=\frac{1}{\left| G \right|}\sum_{i\in G}{\mathrm{AP}_i}
    \vspace{-3pt},
\end{equation}
\noindent
where $G$ represents a dish group we are focusing on (i.e. a subset of \textsc{FmLAMA}).

\paragraph{Mean Word Similarity.}
mWS is defined based on the semantic similarity between predicted and reference objects.
First, we define the similarity score $S\left( i,g \right)$ for each ingredient $g$ within each dish 
$i$. Only the predicted objects in the top-$l$ rankings of the model prediction that are most similar to $g$ contribute to the evaluation score, where $l=\left| \mathrm{ing}_i \right|$. 
$S\left( i,g \right)$ is defined as follows:
\begin{equation}
    \mycustomsize
    S@l\left( i,g \right) =\underset{p\in \mathrm{topl}_i}{\max}\left[ \cos \left( w_g,w_p \right) \right] , g\in \mathrm{ing}_i 
    \vspace{-3pt},
\end{equation}
where $w_g$ and $w_p$ are the \texttt{Fasttext}~\cite{bojanowski-etal-2017-enriching, joulin2016fasttext} vectors for the ingredient $g$ and the predicted object $p$, respectively, and $\cos \left( \cdot \right) $ is the cosine similarity function.
Then we can compute the probing similarity $WS_i$ for each food instance $i$ and mWS for the targeted food group as follows:

\begin{eqnarray}
    \mycustomsize
    \mathrm{WS}_i&=&\frac{1}{\left| \mathrm{ing}_i \right|}\sum_{g\in \mathrm{ing}_i}{S@l\left( i,g \right)} \\
    \mycustomsize \mathrm{mWS}&=&\frac{1}{\left| G \right|}\sum_{i\in G}{\mathrm{WS}_i}
\end{eqnarray}

\begin{table*}[ht]
  \begin{subtable}{1\linewidth}
    \centering
    \scalebox{0.6}{
\begin{tabular}{@{}lr|r|rrr|rr|lll|r@{}}
\toprule
                                       &                                       &                                         & \multicolumn{3}{c|}{\textbf{Encoder-only LLMs}}                                                                 & \multicolumn{2}{c|}{\textbf{Encoder-Decoder LLMs}}                       & \multicolumn{3}{c|}{\textbf{Decoder-only LLMs}}                                                                                                            &                                               \\ \cmidrule(lr){4-11}
\multirow{-2}{*}{\textbf{Origin}}      & \multirow{-2}{*}{\textbf{Count}}      & \multirow{-2}{*}{\textbf{Dumb Base}}    & \textbf{Bb}                         & \textbf{Bl}                         & \textbf{mB}                         & \textbf{T5}                        & \textbf{mT5}                        & \multicolumn{1}{r}{\textbf{Qwen2}}                & \multicolumn{1}{r}{\textbf{Llama2}}               & \multicolumn{1}{r|}{\textbf{Llama3}}               & \multirow{-2}{*}{\textbf{Avg.}}               \\ \midrule
\textbf{Italy}                         & 215 (13.9\%)                          & \underline{18.14}                                   & 8.12±1.14                           & 7.52±0.43                           & 9.31±1.33                           & 4.54±0.56                          & 6.73±1.34                           & 26.82±4.36                                        & 19.37±3.68                                        & 29.62±3.30                                         & 14.00                                         \\
\textbf{U.S.}                          & 285 (18.4\%)                          & 11.10                                   & \textbf{19.98±2.00}                 & \textbf{18.83±1.95}                 & \textbf{19.91±5.88}                 & \textbf{10.24±2.62}                & \textbf{16.50±4.86}                 & 32.21±2.05                                        & 21.64±5.75                                        & 36.58±3.58                                         & \underline{21.99}                \\
\textbf{Turkey}                        & 98 (6.3\%)                            & 12.90                                   & 11.80±1.03                          & 9.84±3.81                           & 14.98±3.89                          & 6.10±1.57                          & 10.25±2.91                          & 25.22±1.46                                        & 22.96±4.04                                        & 30.01±2.30                                         & 16.39                                         \\
\textbf{Japan}                         & 186 (12.0\%)                          & 9.35                                    & 11.10±1.19                          & 9.14±1.81                           & 11.52±1.89                          & 5.45±1.90                          & 6.38±1.00                           & 23.15±2.01                                        & 20.24±3.10                                        & 25.47±1.73                                         & 14.06                                         \\
\textbf{France}                        & 175 (11.3\%)                          & 16.50                                   & 9.33±1.24                           & 7.51±0.90                           & 8.80±2.34                           & 3.86±1.62                          & 4.57±1.11                           & 29.56±2.38                                        & 21.95±4.83                                        & 30.01±2.19                                         & 14.45                                         \\
\textbf{U.K.}                          & 83 (5.4\%)                            & \textbf{18.67}                                   & 12.47±1.77                          & 13.72±2.20                          & 14.20±4.01                          & 6.16±2.89                          & 9.88±3.06                           & 29.06±2.47                                        & 21.82±3.43                                        & 32.15±4.06                                         & 17.43                                         \\
\textbf{Mexico}                        & 57 (3.7\%)                            & 9.40                                    & 8.25±1.04                           & 10.35±2.88                          & 8.87±2.18                           & 3.41±0.49                          & 5.96±2.36                           & 25.14±1.58                                        & 22.03±5.65                                        & 28.97±3.54                                         & 14.12                                         \\
\textbf{India}                         & 132 (8.5\%)                           & 11.57                                   & \underline{18.42±1.35} & \underline{16.94±3.10} & \underline{17.51±1.75} & \underline{9.18±2.98} & 7.80±1.15                           & \underline{35.82±1.79}               & \underline{33.88±5.00}               & \underline{37.62±1.89}                & \textbf{22.15}                                \\
\textbf{Germany}                       & 57 (3.7\%)                            & 14.02                                   & 8.54±1.77                           & 7.84±1.01                           & 9.00±2.39                           & 4.38±1.48                          & 6.68±1.56                           & 29.05±2.63                                        & 20.42±5.56                                        & 28.86±3.78                                         & 14.35                                         \\
\textbf{China}                         & 97 (6.3\%)                            & 8.60                                    & 15.06±2.45                          & 14.94±1.37                          & 15.24±5.31                          & 7.80±1.97                          & \underline{12.39±2.62} & 28.96±3.15                                        & 21.02±4.78                                        & 30.39±2.70                                         & 18.23                                         \\
\textbf{Iran}                          & 21 (1.4\%)                            & 12.60                                   & 9.16±1.75                           & 8.19±1.51                           & 13.56±3.16                          & 9.00±2.04                          & 6.21±0.15                           & \textbf{40.26±8.19}                               & \textbf{35.36±9.81}                               & \textbf{45.72±3.65}                                & 20.93                                         \\
\textbf{Greece}                        & 21 (1.4\%)                            & 15.07                                   & 4.09±1.10                           & 3.48±1.45                           & 4.85±1.37                           & 3.81±1.09                          & 1.32±0.15                           & 31.68±8.85                                        & 15.88±2.54                                        & 31.60±3.17                                         & 12.09                                         \\
\textbf{Spain}                         & 95 (6.1\%)                            & 16.05                                   & 9.18±0.93                           & 7.18±1.04                           & 7.42±1.51                           & 3.98±1.37                          & 3.33±0.46                           & 25.21±5.07                                        & 17.00±4.23                                        & 25.36±1.68                                         & 12.33                                         \\
\textbf{Russia}                        & 27 (1.7\%)                            & 10.64                                   & 4.72±0.98                           & 7.10±2.81                           & 6.44±1.97                           & 1.41±0.60                          & 2.21±0.33                           & 14.83±4.80                                        & 10.47±0.98                                        & 11.77±0.59                                         & 7.37                                          \\
\textbf{ALL}                           & 1549 (100.0\%)                        & 13.29                                   & 12.56±1.06                          & 11.58±1.29                          & 12.96±2.84                          & 6.32±1.42                          & 8.68±2.03                           & 28.56±2.27                                        & 21.89±4.06                                        & 30.96±2.09                                         & 16.69                                         \\ \midrule  \midrule
\rowcolor[HTML]{EFEFEF} 
\multicolumn{2}{l|}{\cellcolor[HTML]{EFEFEF}\textbf{Coefficient of Variation (CV)}} & 24.15  & 40.87                               & 40.76                               & 36.89                               & 43.29                              & 54.29                               & \multicolumn{1}{r}{\cellcolor[HTML]{EFEFEF}20.39} & \multicolumn{1}{r}{\cellcolor[HTML]{EFEFEF}28.27} & \multicolumn{1}{r|}{\cellcolor[HTML]{EFEFEF}23.84} & \multicolumn{1}{r}{\cellcolor[HTML]{EFEFEF}-} \\ \bottomrule
\end{tabular}}
    \caption{Performance results evaluated using \textbf{mAP} (\%). CV indicates the extent of cultural bias in each model's performance.}
  \end{subtable}
  
  \medskip 
  
  \begin{subtable}{1\linewidth}
    \centering
    \scalebox{0.69}{
\begin{tabular}{@{}l|rrr|rr|lll|r@{}}
\toprule
\multirow{2}{*}{\textbf{Origin}}      & \multicolumn{3}{c|}{\textbf{Encoder-only LLMs}}                                                                                & \multicolumn{2}{c|}{\textbf{Encoder-Decoder LLMs}}                                  & \multicolumn{3}{c|}{\textbf{Decoder-only LLMs}}                                                                                                      & \multirow{2}{*}{\textbf{Avg.}} \\ \cmidrule(lr){2-9}
                                      & \textbf{Bb}                              & \textbf{Bl}                              & \textbf{mB}                              & \textbf{T5}                              & \textbf{mT5}                             & \multicolumn{1}{r}{\textbf{Qwen2}}       & \multicolumn{1}{r}{\textbf{Llama2}}      & \multicolumn{1}{r|}{\textbf{Llama3}}                           &                                \\ \midrule
\multicolumn{1}{l|}{\textbf{Italy}}   & 0.3813±0.02                              & 0.3624±0.01                              & 0.3573±0.02                              & 0.3129±0.02                              & 0.3195±0.01                              & 0.4930±0.02                              & 0.4510±0.02                              & \multicolumn{1}{l|}{0.5057±0.01}                              & 0.3979                         \\
\multicolumn{1}{l|}{\textbf{U.S.}}    & \textbf{0.4461±0.02}    & \textbf{0.4358±0.02}    & \textbf{0.4219±0.06}    & \underline{0.3516±0.02} & \textbf{0.4076±0.03}    & 0.5147±0.02                              & 0.4656±0.04                              & \multicolumn{1}{l|}{0.5411±0.02}                              & \textbf{0.4481}                         \\
\multicolumn{1}{l|}{\textbf{Turkey}}  & 0.4130±0.02                              & 0.3886±0.03                              & 0.3954±0.04                              & 0.3387±0.03                              & 0.3419±0.02                              & 0.4689±0.01                              & 0.4612±0.03                              & \multicolumn{1}{l|}{0.5022±0.01}                              & 0.4137                         \\
\multicolumn{1}{l|}{\textbf{Japan}}   & 0.3778±0.02                              & 0.3501±0.03                              & 0.3446±0.03                              & 0.3059±0.03                              & 0.2932±0.01                              & 0.4416±0.01                              & 0.4304±0.02                              & \multicolumn{1}{l|}{0.4436±0.01}                              & 0.3734                         \\
\multicolumn{1}{l|}{\textbf{France}}  & 0.3750±0.02                              & 0.3660±0.02                              & 0.3497±0.04                              & 0.3279±0.02                              & 0.3367±0.01                              & 0.5201±0.02                              & 0.4697±0.03                              & \multicolumn{1}{l|}{0.5099±0.01}                              & 0.4069                         \\
\multicolumn{1}{l|}{\textbf{U.K.}}    & 0.3708±0.02                              & 0.3775±0.02                              & 0.3583±0.04                              & 0.3031±0.02                              & 0.3320±0.03                              & 0.4532±0.03                              & 0.4291±0.01                              & \multicolumn{1}{l|}{0.4620±0.04}                              & 0.3857                         \\
\multicolumn{1}{l|}{\textbf{Mexico}}  & 0.3509±0.01                              & 0.3654±0.03                              & 0.3255±0.03                              & 0.3287±0.01                              & 0.3262±0.02                              & 0.4547±0.02                              & 0.4762±0.04                              & \multicolumn{1}{l|}{0.4863±0.03}                              & 0.3892                         \\
\multicolumn{1}{l|}{\textbf{India}}   & \underline{0.4402±0.01} & \underline{0.4193±0.04} & \underline{0.4135±0.02} & 0.3380±0.04                              & 0.3141±0.01                              & 0.5106±0.00                              & \underline{0.5140±0.03} & \multicolumn{1}{l|}{0.5190±0.02}                              & 0.4336                         \\
\multicolumn{1}{l|}{\textbf{Germany}} & 0.3699±0.03                              & 0.3492±0.03                              & 0.3698±0.03                              & 0.2987±0.02                              & 0.3358±0.01                              & 0.4987±0.02                              & 0.4758±0.05                              & \multicolumn{1}{l|}{0.4983±0.01}                              & 0.3995                         \\
\multicolumn{1}{l|}{\textbf{China}}   & 0.3676±0.02                              & 0.3532±0.02                              & 0.3317±0.05                              & 0.3368±0.03                              & \underline{0.3670±0.03} & 0.4521±0.04                              & 0.4285±0.02                              & \multicolumn{1}{l|}{0.4671±0.02}                              & 0.3880                         \\
\multicolumn{1}{l|}{\textbf{Iran}}    & 0.3950±0.03                              & 0.3747±0.03                              & 0.3972±0.03                              & 0.3384±0.04                              & 0.3303±0.02                              & \textbf{0.5510±0.06}    & \textbf{0.5498±0.07}    & \multicolumn{1}{l|}{\textbf{0.6086±0.03}}    & \underline{0.4431}                         \\
\multicolumn{1}{l|}{\textbf{Greece}}  & 0.3854±0.03                              & 0.3707±0.03                              & 0.3461±0.02                              & \textbf{0.3569±0.04}    & 0.3257±0.01                              & \underline{0.5415±0.08} & 0.4459±0.02                              & \multicolumn{1}{l|}{\underline{0.5527±0.02}} & 0.4156                         \\
\multicolumn{1}{l|}{\textbf{Spain}}   & 0.3761±0.01                              & 0.3386±0.02                              & 0.3290±0.02                              & 0.3215±0.02                              & 0.2923±0.01                              & 0.4630±0.03                              & 0.4247±0.03                              & \multicolumn{1}{l|}{0.4641±0.01}                              & 0.3762                         \\
\multicolumn{1}{l|}{\textbf{Russia}}  & 0.3686±0.01                              & 0.3778±0.03                              & 0.3286±0.01                              & 0.3386±0.01                              & 0.2912±0.02                              & 0.4272±0.04                              & 0.3969±0.02                              & \multicolumn{1}{l|}{0.4053±0.00}                              & 0.3668                         \\ \midrule
\multicolumn{1}{l|}{\textbf{ALL}}     & 0.3959±0.01                              & 0.3803±0.02                              & 0.3699±0.03                              & 0.3276±0.02                              & 0.3380±0.01                              & 0.4864±0.01                              & 0.4573±0.03                              & \multicolumn{1}{l|}{0.4981±0.01}                              & 0.4067                         \\ \bottomrule
\rowcolor[HTML]{EFEFEF} 
\multicolumn{1}{l|}{\cellcolor[HTML]{EFEFEF}\textbf{Corr.}}   & 0.35          & 0.59         & 0.38             & 0.57            & 0.83      & \multicolumn{1}{r}{\cellcolor[HTML]{EFEFEF}0.86} & \multicolumn{1}{r}{\cellcolor[HTML]{EFEFEF}0.71} & \multicolumn{1}{r|}{\cellcolor[HTML]{EFEFEF}0.92} & \multicolumn{1}{r}{\cellcolor[HTML]{EFEFEF}-} \\ \bottomrule
\end{tabular}}
    \caption{Performance results evaluated using \textbf{mWS} with \texttt{Fasttext}. Corr. represents the correlation between the FastText and BERT base embedding evaluation results.}
  \end{subtable}
  
  \caption{\textbf{Automatic Evaluation}: Probing performance comparison with English prompts and \textsc{FmLAMA}-\textit{en} sub-dataset. ``B/mB'' respectively represent abbreviations for BERT and mBERT, and ``b/l'' stands for base/large. \textbf{Bold} and \underline{underline} indicate the best and second-best performing cultural groups, respectively, for each model (within each column).
  The average \texttt{Pearson correlation} between mAP and mWS across LLMs is 0.72.}
  \label{tab:en_without_main}
\end{table*}

\begin{table}[ht]
\centering
\scalebox{0.9}{
\begin{tabular}{@{}lrrrr@{}}
\toprule
\textbf{}    & \textbf{Bb}  & \textbf{Bl}    & \textbf{mB}     & \textbf{T5}     \\ \midrule
\textbf{mAP} & 0.60         & 0.49           & 0.48            & 0.56            \\
\textbf{mWS} & 0.48         & 0.38           & 0.36            & 0.41            \\ \midrule \midrule
             & \textbf{mT5} & \textbf{Qwen2} & \textbf{Llama2} & \textbf{Llama3} \\ \midrule
\textbf{mAP} & 0.38         & 0.05           & 0.04            & 0.13            \\
\textbf{mWS} & -0.06        & 0.05           & -0.04           & 0.02            \\ \bottomrule
\end{tabular}}
\caption{Correlation between Country Dish Counts and Probing Performance.}
\label{tab:count-corr}
\end{table}

\begin{table*}[ht]
\centering
\scalebox{0.9}{
\begin{tabular}{lrr|lrr|lrr}
\toprule
\textbf{Origin} & \textbf{GPT-4o} & \textbf{Human} & \textbf{Origin}  & \textbf{GPT-4o} & \textbf{Human} & \textbf{Origin} & \textbf{GPT-4o} & \textbf{Human} \\ \midrule
\textbf{Italy}  & 52.40           & 55.70          & \textbf{U.K.}    & 64.12           & \textbf{72.33} & \textbf{Iran}   & 53.39           & 51.70          \\
\textbf{U.S.}   & \textbf{70.06}  & \underline{ 70.26}    & \textbf{Mexico}  & 47.05           & 51.95          & \textbf{Greece} & 47.68           & 51.23          \\
\textbf{Turkey} & 55.19           & 53.59          & \textbf{India}   & \underline{ 64.22}     & 63.72          & \textbf{Spain}  & 43.02           & 53.16          \\
\textbf{Japan}  & 50.76           & 56.07          & \textbf{Germany} & 51.71           & 59.02          & \textbf{Russia} & 31.42           & 44.86          \\ \cline{7-9} 
\textbf{France} & 52.03           & 58.28          & \textbf{China}   & 61.52           & 68.96          & \textbf{All}    & 56.58           & -              \\ \bottomrule
\end{tabular}
}
\caption{\textbf{Manual Evaluation: }Llama3 probing results evaluated using \textbf{MES}. The human evaluation score is the average MES from three evaluators per country group. Since evaluators vary by country group, the overall result (ALL) is shown as ``-''.}
\label{tab:manual_evaluation}
\end{table*}

\paragraph{Manual Evaluation Score.} MES considers only the top-$l$ predicted ingredients for evaluation in each dish, similar to the mWS metric, where $l$ is the number of reference ingredients.
A predicted ingredient is considered correct if it satisfies any of the following criteria: (1) \textit{Direct Match}: The ingredient is exactly in the Wikidata reference; (2) \textit{Substitutability}: The ingredient can replace one in the reference; (3) \textit{Missing Traditional Ingredient}: The ingredient is traditionally or commonly used in the dish, but not listed in the reference. 
Given a manual evaluator $\mathrm{M}\left( \cdot \right)$, 
the MES is defined as follows:
\begin{eqnarray}
\mycustomsize
    \mathrm{S}_i&=&\frac{1}{l}\sum_{p\in \mathrm{topl}_i}{\mathrm{M}\left( p,\mathrm{ing}_i \right)} \\
\mycustomsize
    \mathrm{MES}&=&\frac{1}{\left| G \right|}\sum_{i\in \mathrm{G}}{\mathrm{S}_i}
\end{eqnarray} 
where $\mathrm{M}\left( \cdot \right)$ is 1 if the top-$l$ predicted ingredient $p$ is evaluated as meeting the criteria; otherwise, 0.

\section{Experiments: Existing Cultural Biases}
\label{sec:experiments}

\subsection{Experimental setup}

\paragraph{Baselines}
We explore encoder-only LLMs, including BERT~\cite{devlin-etal-2019-bert} and mBERT, encoder-decoder LLMs, such as T5~\cite{raffel2020exploring} and mT5~\cite{xue2020mt5}, as well as decoder-only LLMs like Qwen2~\cite{yang2024qwen2}, Llama2~\cite{touvron2023llama}, and Llama3~\cite{llama3modelcard}. 
Of these, BERT and T5 are English monolingual LLMs, while the remaining five are multilingual LLMs.
\footnote{Monolingual LLM configurations, as well as results for the five languages on their respective filtered sub-datasets, are provided in the Appendix~\ref{app:language_results}.}
Additionally, we employ a `dumb' baseline, in which the model is assumed to consistently predict the top-10 most common ingredients for each dish.

\paragraph{Cultural groups}

We identify the top 30 countries with the most dishes across six languages, take their intersection to ensure sufficient data, consider geographical diversity, and ultimately narrow our focus to 14 cultural groups, as shown in Table~\ref{tab:en_without_main}.

\paragraph{Metrics}
We evaluate all LLMs using two automated metrics. Due to cost constraints, we conduct manual evaluation only on the LLM that achieve the highest probing performance in the automated metrics. Specifically, for the evaluator $\mathrm{M}\left( \cdot \right)$ used to compute MES, we utilize GPT-4o~\cite{openai2024gpt4technicalreport}\footnote{Version: gpt-4o-2024-08-06} as a simulated evaluator and recruit real human evaluators from Prolific.\footnote{\url{https://www.prolific.co/}} 
For each country group, we hire 3 evaluators familiar with the related cuisine, with a pay rate of £9/hour, considered mid-level on the platform.
Details on the GPT-4o evaluation prompts and the human evaluation platform are provided in the Appendix~\ref{app:manual-evaluation}.

\subsection{Automatic evaluation results}
The probing results based on English prompt on the filtered dataset \textsc{FmLAMA}-\textit{en} are illustrated in Table~\ref{tab:en_without_main}.
The average Pearson correlation between mAP and mWS across all LLMs is 0.72, indicating a strong positive relationship and demonstrating that the experimental results are consistent across both automatic metrics.
Overall, without fine-tuning, decoder-only LLMs exhibit significantly better cultural knowledge recall in such challenging tasks compared to encoder-only and encoder-decoder LLMs, highlighting their superior capacity as knowledge bases~\cite{petroni-etal-2019-language}.
Besides, results across all models (Avg. Column) indicate that the groups from the U.S. and India, whose official languages include English, perform the best generally. 
They consistently rank in the top 3 in both automatic evaluation metrics.
This pattern is especially pronounced in the results for encoder-only LLMs and encoder-decoder LLMs. 
Apart from the Iran group\footnote{See detailed analysis in Appendix~\ref{app:Iran}.}, these two cultural groups also demonstrate the best performance in decoder-only LLMs. Especially, for the `dumb' baseline, the U.S. probing performance is not the best, suggesting that the data distribution does not influence the LLM's cultural judgment capabilities. This further strengthens the persuasiveness of our experiment.

Additionally, (1) to assess the variation in model performance across different cultural groups, we calculate the Coefficient of Variation based on the absolute-match mAP metric. This serves as an indicator of the degree of cultural bias in each model's performance. Our results show that decoder-only LLMs exhibit less cultural bias compared to other model types. Among these, the newer version, Llama3, demonstrates greater consistency in performance than Llama2. Similarly, the non-English-dominant Qwen2 LLM also shows smaller variations in cultural bias.
(2) To explore the impact of different embedding methods in mWS, we use BERT Base with mean pooling to evaluate probing performance under the mWS metric. Results (see Table~\ref{tab: en_without_BERT_mWS}) show that the U.S. cultural group achieves the highest performance, while decoder-only LLMs exhibit significant improvement, consistent with findings using FastText. We calculate the correlation between mWS evaluations using BERT Base and FastText for each LLM. The results reveal a strong correlation for decoder-only LLMs, as shown in Table~\ref{tab:en_without_main}.
(3) To explore the correlation between training data size and cultural probing performance, we use country dish counts as a proxy for Wikidata's cultural data size and calculate their correlation with probing performance. The correlation results are show as Table~\ref{tab:count-corr}. Based on the probing performance evaluated by the two automated metrics, LLMs like Bb, B1, and T5 consistently exhibit little correlations with cultural data size. In contrast, decoder-only LLMs show almost no correlations, indicating limited sensitivity to cultural data size as represented by dish counts. 
Therefore, in more advanced LLMs, cultural competency isn’t directly tied to the training sample size of specific cultural groups.

\begin{figure}[t]
    \centering
    \includegraphics[height=5cm,width=1\linewidth, trim=0cm 0.5cm 0cm 0cm, clip]{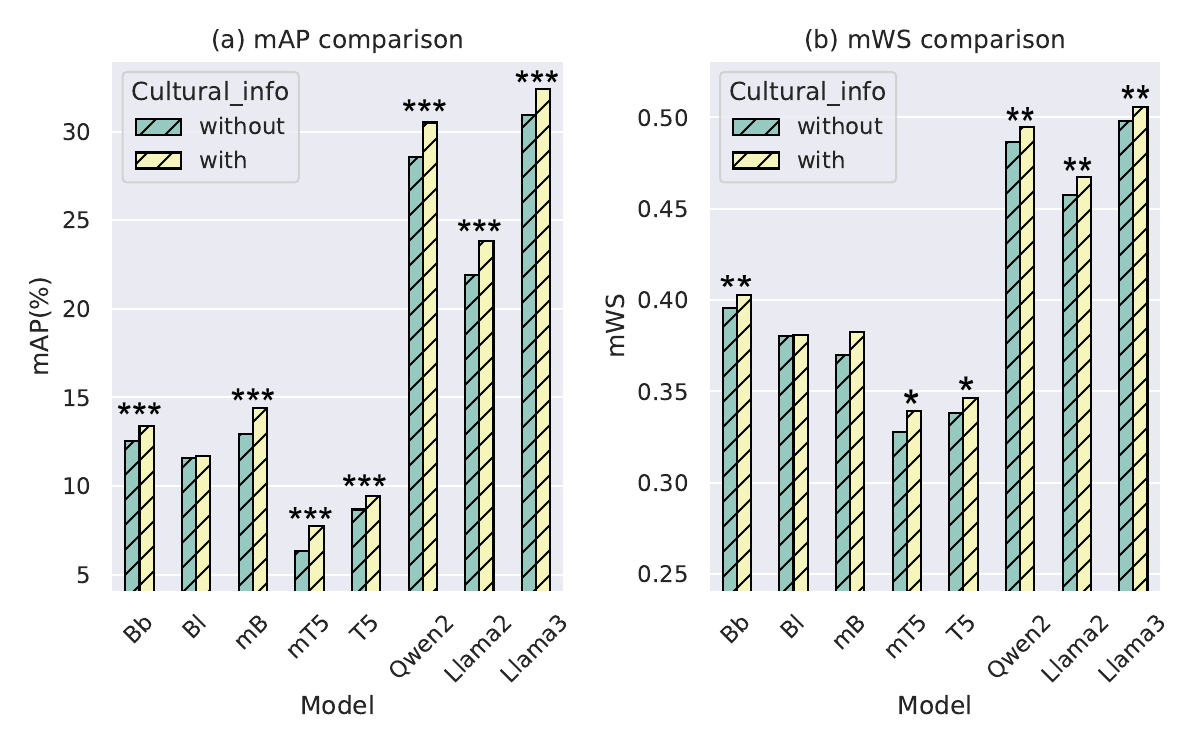}
    \caption{Comparative probing results on \textsc{FmLAMA}-\textit{en}: 
   incorporating cultural information about the origin of dishes into English prompts can enhance the probing of cultural knowledge. Significance levels are indicated by $*$ for $p < 0.1$, $**$ for $p < 0.05$, and $***$ for $p < 0.01$.
    }
    \label{fig:cultural_comparison}
\end{figure}

\subsection{Manual evaluation results}
Table~\ref{tab:en_without_main} shows that Llama3 achieve the best probing performance, so we conduct manual evaluation on its probing results. 
Table~\ref{tab:manual_evaluation} displays the results of the GPT-4o simulator and human evaluations, showing a high Pearson correlation of 0.88.
Both evaluation methods consistently show that the highest MES scores come from English-speaking cultural groups, including the U.S., U.K., and India, with the U.S. performing especially well in both manual evaluations. 
This aligns closely with the automatic evaluation results, reinforcing the argument that LLMs tend to exhibit cultural bias toward these groups.

\begin{figure}[t]
    \centering
    \includegraphics[width=1\linewidth, trim=0cm 1.0cm 0cm 0cm, clip]{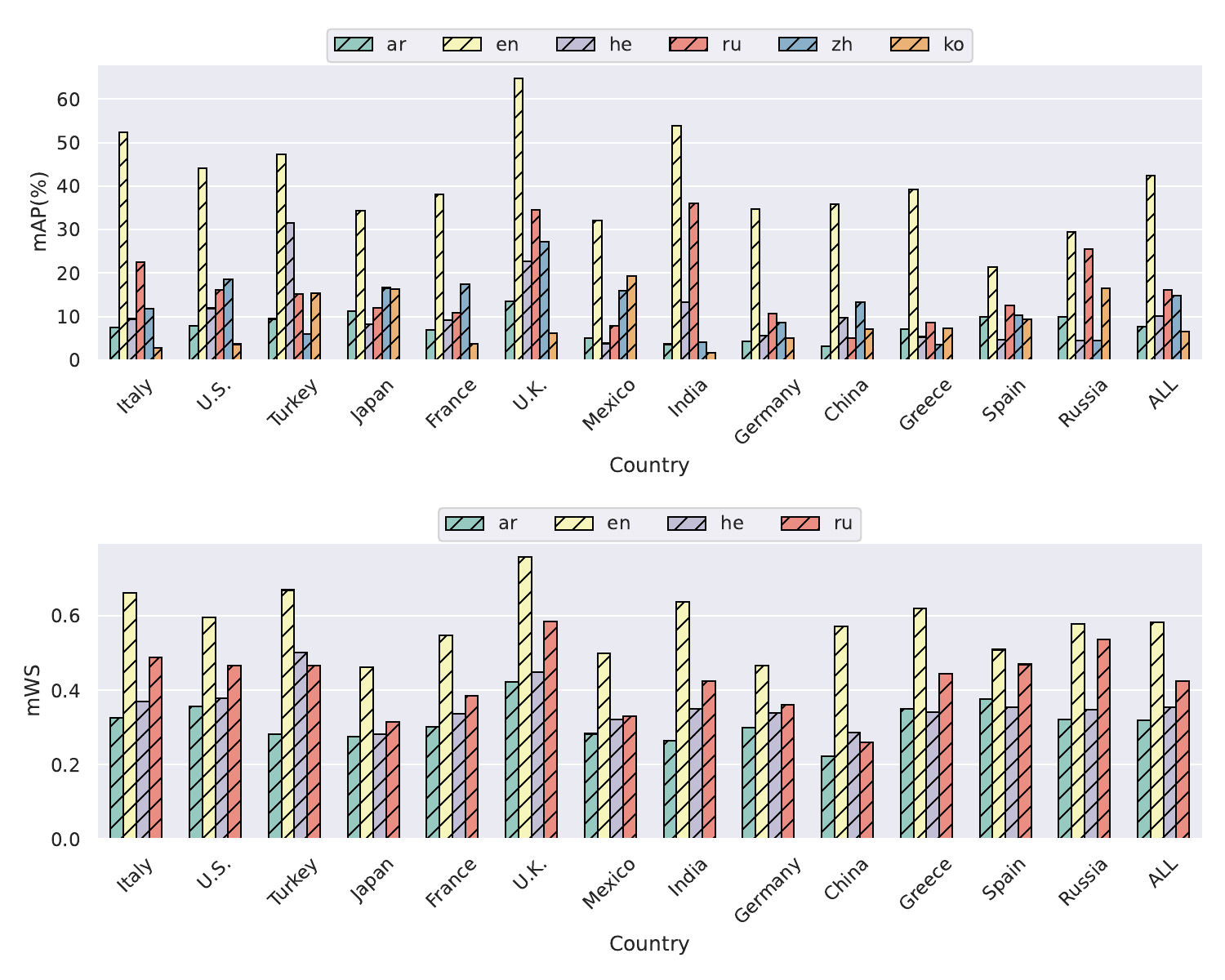}
    \caption{Comparison of probing results on Llama3 with prompts in different languages, showing that the English prompt exhibits the best performance.
    }
    \label{fig:language-comparison}
\end{figure}

\begin{figure*}[t]
    \centering
    \includegraphics[width=1\linewidth,  trim=0cm 0cm 0cm 0cm, clip]{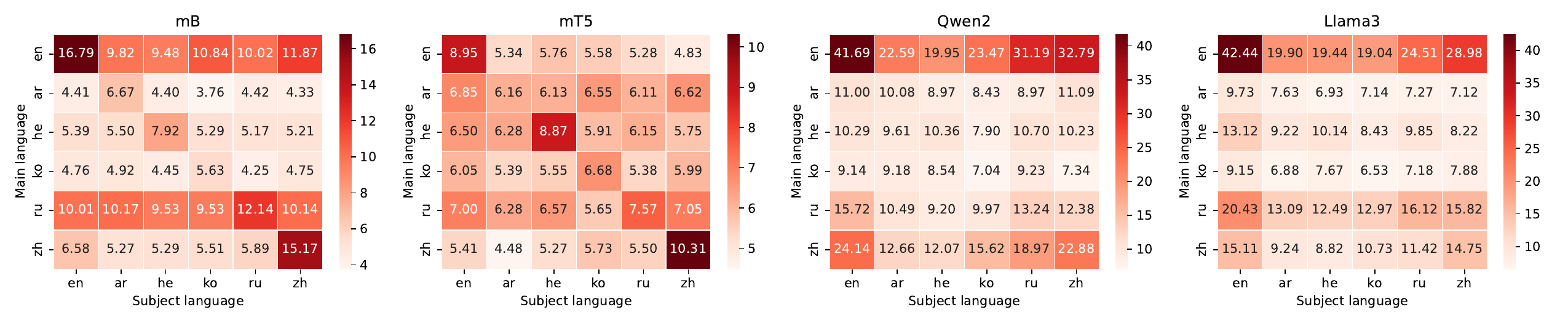}
    \caption{Probing results using code-switching prompts on Qwen2 and LLaMA3.}
    \label{fig:code_switching}
\end{figure*}

Upon reviewing the evaluation results for the GPT-4o simulator, we find that it treats synonymous expressions as \textit{Direct Matches}, e.g., \textit{cooked rice} vs. \textit{gohan} (the Japanese term) and \textit{cauliflower} vs. \textit{Brassica oleracea var. botrytis} (the botanical term). Additionally, reference labels from Wikidata often exhibit inconsistencies, with terms like "shrimp or prawn" and "fish as food" marked as ingredients. Furthermore, GPT-4o flags pre-made ingredients as \textit{Substitutable}; for example, \textit{flour} in Garganelli with pre-made \textit{pasta}.
For human evaluators, evaluation results may vary due to individual subjective biases. 
The average Inter-Rater Reliability (IRR) across country groups, measured by Fleiss' kappa, is 0.57,\footnote{IRR and the average evaluation time for each country group are shown in Appendix~\ref{app:manual-evaluation}.} indicating moderate agreement. 
While this indicates acceptable reliability for subjective tasks, it also highlights that evaluators from the same cultural background may still show variation in their judgments.
Although GPT-4o can partially simulate human evaluations, its hallucination issues~\cite{ji2023survey} raise concerns about the reliability of its assessments. 
These observations underscore the inherent challenges of using automatic metrics for culturally-related tasks.




\section{Analysis: Designing Better Prompts}
\label{sec:analysis}
We examine the impact of different prompt settings on cross-cultural knowledge exploration, focusing on cultural background references, multilingual prompts, and code-switching scenarios.

\paragraph{Cultural background analysis.}
Using the English prompts from Figure~\ref{fig:probing_template} as a basis, we incorporate information about the dish country of origin into the probing prompts for each specific dish (as described in \S\ref{sec:probing_templates}). Figure~\ref{fig:cultural_comparison} presents the comparative probing results on \textsc{FmLAMA}-\textit{en} across different LLMs, considering the inclusion of cultural information mentions in the prompts. 
We find that English prompts with cultural information achieve higher probing scores in capturing LLMs' knowledge within the food domain. This suggests the importance of emphasizing the cultural background when utilizing LLMs, especially in the exploration of culture-related topics.


\paragraph{Multilingual prompt analysis.} 
We utilize prompts in six different languages to conduct knowledge probing on the multilingual LLM, Llama3.\footnote{The probing results comparison on other multilingual LLMs are shown in Appendix~\ref{app:language}.} 
To ensure a fair comparison of probing results across different language prompts, we filter a sub-dataset where all food instances have both subject and ingredient labels available in all the languages involved.
Figure~\ref{fig:language-comparison} displays probing results on the filtered sub-dataset with prompts in different languages.\footnote{Some Chinese and Korean objects are missing Fasttext vectors. Therefore, mWS cannot be calculated for them.}
The English prompt performs significantly better than prompts in other languages across all cultural groups, probably due to more training data. 

\paragraph{Code-switching analysis.}
Figure~\ref{fig:code_switching} presents the probing results using code-switching prompts, for each pair out of six languages, applied to the two best-performing decoder-only LLMs, Qwen2 and Llama3. 
When English is the main language and the dish subject is presented in another language, probing performance significantly declines but still outperforms probing conducted solely in other languages. This suggests that the instruction language is important, not just the language of the dish name. 
Furthermore, when the main language of the prompt is non-English and the subject language is English, performance does not significantly decline and even shows a slight improvement. This further emphasizes the dominant role of English in multilingual LLMs.
The detailed probing results of each cultural group and other multilingual LLMs are in shown Appendix~\ref{app:cs}.

\begin{table}[t]
  \begin{subtable}{1\linewidth}
    \centering
    \scalebox{0.8}{
    \begin{tabular}{@{}l|p{7.5cm}}
\toprule
\textbf{Country} & \textbf{ Top 5 Most Falsely Predictions (Llama3)} \\ \midrule
\textbf{Italy}   & pasta, bread, meat, egg, pork                   \\
\textbf{Japan}   & rice, fish, soy, tofu, chicken                  \\
\textbf{France}  & butter, cream, meat, caramel, bread             \\
\textbf{Russia}  & meat, beef, cabbage, pork, beet                 \\ \bottomrule
\end{tabular}
}
    \caption{Coarse-grained confusions}
  \end{subtable}
  
  \medskip 
  
  \begin{subtable}{1\linewidth}
    \centering
    \scalebox{0.8}{

    \begin{tabular}{@{}l|p{7.5cm}}
\toprule
\textbf{Model} & \textbf{ Top 5 Most Frequent Predictions} \\ \midrule
\textbf{BERT}     & dried meat, ham, beef, pork, fish                              \\
\textbf{T5}       & vinaigrette, scrambled eggs, buckwheat flour, ricotta, sultana \\
\textbf{Llama2}   & rice, meat, bread, white meat, egg                             \\
\textbf{Llama3}   & rice, meat, sugar, chicken, bread                              \\ \bottomrule
\end{tabular}

}
    \caption{Fine-grained confusions}
  \end{subtable}
  
  \caption{Error Analysis: Coarse and fine-grained confusions in ingredient predictions by various cultural groups and models.}
  \label{tab:ingredients}
\end{table}

\section{Error Analysis} 
\label{sec:evaluation}
\label{subsec:case_study}

A manual inspection of a data sample\footnote{Conducted by one of the authors on 500 predictions by Llama3, Llama2, Bert-base-uncased and T5-base.} followed by an automated evaluation on the complete data reveals recurring mistake patterns: (1) \textbf{Coarse-grained confusions} (foreign ingredients, e.g., \textit{coffee} in a Chinese dish), where the main issue lies in the model's lack of cultural attribute awareness for both the dish and its ingredients.
(2) \textbf{Fine-grained confusions} (wrong local ingredients, e.g., \textit{Chinese wine} in \textit{Mapo tofu}), and (3) \textbf{Inconsistent confusions}, where the model predicts different incorrect ingredients across prompts for the same dish.

Table \ref{tab:ingredients} illustrates the described error patterns. In the results for Llama3 in English, $47\%$ of the incorrect predictions for Italian dishes are the ingredient \textit{pasta}, $45\%$ of the incorrect predictions for French dishes are \textit{butter}, and $40\%$ of the incorrect predictions for Japanese dishes are \textit{tofu}. These are examples of fine-grained confusions. Additionally, \textit{rice}, \textit{wheat} and \textit{flour} account for $11\%$ of the Llama3's total predictions, regardless of the dish's origin. This is a coarse confusion, where the model often predicts the same foreign ingredient. 
 
These errors reveal a lack of cultural knowledge about the dishes and suggest different types of guessing. While coarse and inconsistent confusions seem to reflect more general guesses, possibly influenced by the frequency of certain ingredients in the training data, fine-grained confusions indicate that the model has some knowledge of the dish’s origin and is attempting to guess a local ingredient accordingly (i.e. overfitting to the context). Specifically, for fine-grained errors, even when the cultural attributes of candidate ingredients align with the dish, LLMs still make incorrect predictions. While many studies have shown that LLMs can function as knowledge bases~\cite{nguyen2024cultural,pan2024unifying} and memorize factual information~\cite{fierro2024multilingual, fierro-etal-2024-mulan}, the familiarity with specific entities is crucial in shaping their predictive expectations~\cite{du-etal-2024-context}. 

\section{Conclusion}
This study presents an automated method for generating extensive cultural knowledge datasets, exemplified by the creation of \textsc{FmLAMA}, a diverse, food-centric dataset that spans multiple cultures and languages. We introduce novel metrics for cultural knowledge evaluation in LLMs, emphasizing the influence of cultural context and language in the probing process. Our findings reveal a predominant bias towards American culture in LLMs when using English prompts, a bias that diminishes with prompts in other languages. Interestingly, incorporating explicit cultural cues in prompts enhances LLMs' cultural knowledge access. The study also highlights the scarcity of culturally diverse knowledge across languages, pointing to a potential root of observed biases in LLMs.

\section{Limitations}
While this study provides valuable insights into cross-cultural knowledge probing in LLMs, it is essential to acknowledge several limitations.
Firstly, the food domain knowledge dataset utilized in this research is sourced from Wikidata, which may not offer comprehensive coverage. For example, the dish \textit{soy sauce chicken} may only include the ingredient \textit{chicken meat} while lacking the inclusion of \textit{soy sauce}.
Moreover, ingredient descriptions are not always detailed. For instance, the Wikidata gold label might be \textit{oil} when the recipe requires a specific type of oil, such as \textit{sesame oil}. This inconsistency underscores the motivation behind our mWS metric.
Furthermore, aside from well-known dishes, certain recipes lack standardization and may vary depending on individual preferences and cooking styles, posing challenges to precise probing.
Additionally, the fuzzy-match metric mWS, introduced in this study, relies on Fasttext for obtaining object representation vectors. However, for certain objects in Chinese and Korean, zero vectors may result, rendering similarity calculation impossible.
Furthermore, we employ manually crafted templates in this paper. However, research has shown that sampling templates from large corpora can also enhance knowledge-probing evaluation. This aspect is deferred to future work.
Despite our endeavors to construct comprehensive multilingual and multicultural knowledge repositories, the availability of aligned cross-cultural knowledge remains limited in multilingual settings. This constraint presents challenges in exploring the interaction between language and culture.




\section*{ACKNOWLEDGMENTS}
This research is supported by the project of Shenzhen Science and Technology Research Fund (Fundamental Research Key Project Grant No. JCYJ20220818103001002), Shenzhen Science and Technology Program (Grant No. ZDSYS20230626091302006), Key Project of Shenzhen Higher Education Stability Support Program (Grant No. 2024SC0009).


\bibliography{anthology,custom}

\begin{thebibliography}{48}
\expandafter\ifx\csname natexlab\endcsname\relax\def\natexlab#1{#1}\fi

\bibitem[{Adilazuarda et~al.(2024)Adilazuarda, Mukherjee, Lavania, Singh, Aji,
  O{'}Neill, Modi, and Choudhury}]{adilazuarda-etal-2024-towards}
Muhammad~Farid Adilazuarda, Sagnik Mukherjee, Pradhyumna Lavania,
  Siddhant~Shivdutt Singh, Alham~Fikri Aji, Jacki O{'}Neill, Ashutosh Modi, and
  Monojit Choudhury. 2024.
\newblock \href {https://doi.org/10.18653/v1/2024.emnlp-main.882} {Towards
  measuring and modeling {``}culture{''} in {LLM}s: A survey}.
\newblock In \emph{Proceedings of the 2024 Conference on Empirical Methods in
  Natural Language Processing}, pages 15763--15784, Miami, Florida, USA.
  Association for Computational Linguistics.

\bibitem[{Aguilar and Solorio(2020)}]{aguilar-solorio-2020-english}
Gustavo Aguilar and Thamar Solorio. 2020.
\newblock \href {https://doi.org/10.18653/v1/2020.acl-main.716} {From {E}nglish
  to code-switching: Transfer learning with strong morphological clues}.
\newblock In \emph{Proceedings of the 58th Annual Meeting of the Association
  for Computational Linguistics}, pages 8033--8044, Online. Association for
  Computational Linguistics.

\bibitem[{AI@Meta(2024)}]{llama3modelcard}
AI@Meta. 2024.
\newblock \href {https://github.com/meta-llama/llama3/blob/main/MODEL_CARD.md}
  {Llama 3 model card}.

\bibitem[{Bojanowski et~al.(2017)Bojanowski, Grave, Joulin, and
  Mikolov}]{bojanowski-etal-2017-enriching}
Piotr Bojanowski, Edouard Grave, Armand Joulin, and Tomas Mikolov. 2017.
\newblock \href {https://doi.org/10.1162/tacl_a_00051} {Enriching word vectors
  with subword information}.
\newblock \emph{Transactions of the Association for Computational Linguistics},
  5:135--146.

\bibitem[{Cao et~al.(2023)Cao, Zhou, Lee, Cabello, Chen, and
  Hershcovich}]{cao-etal-2023-assessing}
Yong Cao, Li~Zhou, Seolhwa Lee, Laura Cabello, Min Chen, and Daniel
  Hershcovich. 2023.
\newblock \href {https://doi.org/10.18653/v1/2023.c3nlp-1.7} {Assessing
  cross-cultural alignment between {C}hat{GPT} and human societies: An
  empirical study}.
\newblock In \emph{Proceedings of the First Workshop on Cross-Cultural
  Considerations in NLP (C3NLP)}, pages 53--67, Dubrovnik, Croatia. Association
  for Computational Linguistics.

\bibitem[{Deshpande et~al.(2022)Deshpande, Ruiter, Mosbach, and
  Klakow}]{deshpande-etal-2022-stereokg}
Awantee Deshpande, Dana Ruiter, Marius Mosbach, and Dietrich Klakow. 2022.
\newblock \href {https://doi.org/10.18653/v1/2022.woah-1.7} {{S}tereo{KG}:
  Data-driven knowledge graph construction for cultural knowledge and
  stereotypes}.
\newblock In \emph{Proceedings of the Sixth Workshop on Online Abuse and Harms
  (WOAH)}, pages 67--78, Seattle, Washington (Hybrid). Association for
  Computational Linguistics.

\bibitem[{Devlin et~al.(2019)Devlin, Chang, Lee, and
  Toutanova}]{devlin-etal-2019-bert}
Jacob Devlin, Ming-Wei Chang, Kenton Lee, and Kristina Toutanova. 2019.
\newblock \href {https://doi.org/10.18653/v1/N19-1423} {{BERT}: Pre-training of
  deep bidirectional transformers for language understanding}.
\newblock In \emph{Proceedings of the 2019 Conference of the North {A}merican
  Chapter of the Association for Computational Linguistics: Human Language
  Technologies, Volume 1 (Long and Short Papers)}, pages 4171--4186,
  Minneapolis, Minnesota. Association for Computational Linguistics.

\bibitem[{Do{\u{g}}ru{\"o}z et~al.(2021)Do{\u{g}}ru{\"o}z, Sitaram, Bullock,
  and Toribio}]{dogruoz-etal-2021-survey}
A.~Seza Do{\u{g}}ru{\"o}z, Sunayana Sitaram, Barbara~E. Bullock, and
  Almeida~Jacqueline Toribio. 2021.
\newblock \href {https://doi.org/10.18653/v1/2021.acl-long.131} {A survey of
  code-switching: Linguistic and social perspectives for language
  technologies}.
\newblock In \emph{Proceedings of the 59th Annual Meeting of the Association
  for Computational Linguistics and the 11th International Joint Conference on
  Natural Language Processing (Volume 1: Long Papers)}, pages 1654--1666,
  Online. Association for Computational Linguistics.

\bibitem[{Du et~al.(2024)Du, Sn{\ae}bjarnarson, Stoehr, White, Schein, and
  Cotterell}]{du-etal-2024-context}
Kevin Du, V{\'e}steinn Sn{\ae}bjarnarson, Niklas Stoehr, Jennifer White, Aaron
  Schein, and Ryan Cotterell. 2024.
\newblock \href {https://doi.org/10.18653/v1/2024.acl-long.714} {Context versus
  prior knowledge in language models}.
\newblock In \emph{Proceedings of the 62nd Annual Meeting of the Association
  for Computational Linguistics (Volume 1: Long Papers)}, pages 13211--13235,
  Bangkok, Thailand. Association for Computational Linguistics.

\bibitem[{Elazar et~al.(2021)Elazar, Kassner, Ravfogel, Ravichander, Hovy,
  Sch{\"u}tze, and Goldberg}]{elazar-etal-2021-measuring}
Yanai Elazar, Nora Kassner, Shauli Ravfogel, Abhilasha Ravichander, Eduard
  Hovy, Hinrich Sch{\"u}tze, and Yoav Goldberg. 2021.
\newblock \href {https://doi.org/10.1162/tacl_a_00410} {Measuring and improving
  consistency in pretrained language models}.
\newblock \emph{Transactions of the Association for Computational Linguistics},
  9:1012--1031.

\bibitem[{Fierro et~al.(2024{\natexlab{a}})Fierro, Foroutan, Elliott, and
  S{\o}gaard}]{fierro2024multilingual}
Constanza Fierro, Negar Foroutan, Desmond Elliott, and Anders S{\o}gaard.
  2024{\natexlab{a}}.
\newblock How do multilingual models remember? investigating multilingual
  factual recall mechanisms.
\newblock \emph{arXiv preprint arXiv:2410.14387}.

\bibitem[{Fierro et~al.(2024{\natexlab{b}})Fierro, Garneau, Bugliarello,
  Kementchedjhieva, and S{\o}gaard}]{fierro-etal-2024-mulan}
Constanza Fierro, Nicolas Garneau, Emanuele Bugliarello, Yova Kementchedjhieva,
  and Anders S{\o}gaard. 2024{\natexlab{b}}.
\newblock \href {https://doi.org/10.18653/v1/2024.naacl-short.67} {{M}u{L}an: A
  study of fact mutability in language models}.
\newblock In \emph{Proceedings of the 2024 Conference of the North American
  Chapter of the Association for Computational Linguistics: Human Language
  Technologies (Volume 2: Short Papers)}, pages 762--771, Mexico City, Mexico.
  Association for Computational Linguistics.

\bibitem[{Fierro and S{\o}gaard(2022)}]{fierro-sogaard-2022-factual}
Constanza Fierro and Anders S{\o}gaard. 2022.
\newblock \href {https://doi.org/10.18653/v1/2022.findings-acl.240} {Factual
  consistency of multilingual pretrained language models}.
\newblock In \emph{Findings of the Association for Computational Linguistics:
  ACL 2022}, pages 3046--3052, Dublin, Ireland. Association for Computational
  Linguistics.

\bibitem[{Gonz{\'a}lez et~al.(2021)Gonz{\'a}lez, Rogers, and
  S{\o}gaard}]{gonzalez-etal-2021-interaction}
Ana~Valeria Gonz{\'a}lez, Anna Rogers, and Anders S{\o}gaard. 2021.
\newblock \href {https://doi.org/10.18653/v1/2021.findings-acl.259} {On the
  interaction of belief bias and explanations}.
\newblock In \emph{Findings of the Association for Computational Linguistics:
  ACL-IJCNLP 2021}, pages 2930--2942, Online. Association for Computational
  Linguistics.

\bibitem[{Ji et~al.(2023)Ji, Lee, Frieske, Yu, Su, Xu, Ishii, Bang, Madotto,
  and Fung}]{ji2023survey}
Ziwei Ji, Nayeon Lee, Rita Frieske, Tiezheng Yu, Dan Su, Yan Xu, Etsuko Ishii,
  Ye~Jin Bang, Andrea Madotto, and Pascale Fung. 2023.
\newblock Survey of hallucination in natural language generation.
\newblock \emph{ACM Computing Surveys}, 55(12):1--38.

\bibitem[{Jiang et~al.(2020{\natexlab{a}})Jiang, Anastasopoulos, Araki, Ding,
  and Neubig}]{jiang-etal-2020-x}
Zhengbao Jiang, Antonios Anastasopoulos, Jun Araki, Haibo Ding, and Graham
  Neubig. 2020{\natexlab{a}}.
\newblock \href {https://doi.org/10.18653/v1/2020.emnlp-main.479} {{X}-{FACTR}:
  Multilingual factual knowledge retrieval from pretrained language models}.
\newblock In \emph{Proceedings of the 2020 Conference on Empirical Methods in
  Natural Language Processing (EMNLP)}, pages 5943--5959, Online. Association
  for Computational Linguistics.

\bibitem[{Jiang et~al.(2020{\natexlab{b}})Jiang, Xu, Araki, and
  Neubig}]{10.1162/tacl_a_00324}
Zhengbao Jiang, Frank~F. Xu, Jun Araki, and Graham Neubig. 2020{\natexlab{b}}.
\newblock \href {https://doi.org/10.1162/tacl_a_00324} {{How Can We Know What
  Language Models Know?}}
\newblock \emph{Transactions of the Association for Computational Linguistics},
  8:423--438.

\bibitem[{Joulin et~al.(2016)Joulin, Grave, Bojanowski, Douze, J{\'e}gou, and
  Mikolov}]{joulin2016fasttext}
Armand Joulin, Edouard Grave, Piotr Bojanowski, Matthijs Douze, H{\'e}rve
  J{\'e}gou, and Tomas Mikolov. 2016.
\newblock Fasttext. zip: Compressing text classification models.
\newblock \emph{arXiv preprint arXiv:1612.03651}.

\bibitem[{Kaneko et~al.(2022)Kaneko, Imankulova, Bollegala, and
  Okazaki}]{kaneko-etal-2022-gender}
Masahiro Kaneko, Aizhan Imankulova, Danushka Bollegala, and Naoaki Okazaki.
  2022.
\newblock \href {https://doi.org/10.18653/v1/2022.naacl-main.197} {Gender bias
  in masked language models for multiple languages}.
\newblock In \emph{Proceedings of the 2022 Conference of the North American
  Chapter of the Association for Computational Linguistics: Human Language
  Technologies}, pages 2740--2750, Seattle, United States. Association for
  Computational Linguistics.

\bibitem[{Kassner et~al.(2021)Kassner, Dufter, and
  Sch{\"u}tze}]{kassner-etal-2021-multilingual}
Nora Kassner, Philipp Dufter, and Hinrich Sch{\"u}tze. 2021.
\newblock \href {https://doi.org/10.18653/v1/2021.eacl-main.284} {Multilingual
  {LAMA}: Investigating knowledge in multilingual pretrained language models}.
\newblock In \emph{Proceedings of the 16th Conference of the European Chapter
  of the Association for Computational Linguistics: Main Volume}, pages
  3250--3258, Online. Association for Computational Linguistics.

\bibitem[{Keleg and Magdy(2023)}]{keleg-magdy-2023-dlama}
Amr Keleg and Walid Magdy. 2023.
\newblock \href {https://doi.org/10.18653/v1/2023.findings-acl.389} {{DLAMA}: A
  framework for curating culturally diverse facts for probing the knowledge of
  pretrained language models}.
\newblock In \emph{Findings of the Association for Computational Linguistics:
  ACL 2023}, pages 6245--6266, Toronto, Canada. Association for Computational
  Linguistics.

\bibitem[{Kuratov and Arkhipov(2019)}]{kuratov2019adaptation}
Yuri Kuratov and Mikhail Arkhipov. 2019.
\newblock Adaptation of deep bidirectional multilingual transformers for
  russian language.
\newblock \emph{arXiv preprint arXiv:1905.07213}.

\bibitem[{Lent and S{\o}gaard(2021)}]{lent-sogaard-2021-common}
Heather Lent and Anders S{\o}gaard. 2021.
\newblock \href {https://doi.org/10.18653/v1/2021.wnut-1.14} {Common sense bias
  in semantic role labeling}.
\newblock In \emph{Proceedings of the Seventh Workshop on Noisy User-generated
  Text (W-NUT 2021)}, pages 114--119, Online. Association for Computational
  Linguistics.

\bibitem[{Meng et~al.(2022)Meng, Liu, Shareghi, Su, Collins, and
  Collier}]{meng-etal-2022-rewire}
Zaiqiao Meng, Fangyu Liu, Ehsan Shareghi, Yixuan Su, Charlotte Collins, and
  Nigel Collier. 2022.
\newblock \href {https://doi.org/10.18653/v1/2022.acl-long.329}
  {Rewire-then-probe: A contrastive recipe for probing biomedical knowledge of
  pre-trained language models}.
\newblock In \emph{Proceedings of the 60th Annual Meeting of the Association
  for Computational Linguistics (Volume 1: Long Papers)}, pages 4798--4810,
  Dublin, Ireland. Association for Computational Linguistics.

\bibitem[{Minkov and Hofstede(2012)}]{minkov2012national}
Michael Minkov and Geert Hofstede. 2012.
\newblock Is national culture a meaningful concept? cultural values delineate
  homogeneous national clusters of in-country regions.
\newblock \emph{Cross-Cultural Research}, 46(2):133--159.

\bibitem[{Mukherjee et~al.(2024)Mukherjee, Caliskan, Zhu, and
  Anastasopoulos}]{mukherjee-etal-2024-global}
Anjishnu Mukherjee, Aylin Caliskan, Ziwei Zhu, and Antonios Anastasopoulos.
  2024.
\newblock \href {https://doi.org/10.18653/v1/2024.naacl-long.355} {Global
  gallery: The fine art of painting culture portraits through multilingual
  instruction tuning}.
\newblock In \emph{Proceedings of the 2024 Conference of the North American
  Chapter of the Association for Computational Linguistics: Human Language
  Technologies (Volume 1: Long Papers)}, pages 6398--6415, Mexico City, Mexico.
  Association for Computational Linguistics.

\bibitem[{Nguyen et~al.(2023)Nguyen, Razniewski, Varde, and
  Weikum}]{nguyen2023extracting}
Tuan-Phong Nguyen, Simon Razniewski, Aparna Varde, and Gerhard Weikum. 2023.
\newblock Extracting cultural commonsense knowledge at scale.
\newblock In \emph{Proceedings of the ACM Web Conference 2023}, pages
  1907--1917.

\bibitem[{Nguyen et~al.(2024)Nguyen, Razniewski, and
  Weikum}]{nguyen2024cultural}
Tuan-Phong Nguyen, Simon Razniewski, and Gerhard Weikum. 2024.
\newblock Cultural commonsense knowledge for intercultural dialogues.
\newblock In \emph{Proceedings of the 33rd ACM International Conference on
  Information and Knowledge Management}, pages 1774--1784.

\bibitem[{OpenAI et~al.(2024)OpenAI, Achiam, Adler, Agarwal, Ahmad, Akkaya,
  Aleman, Almeida, Altenschmidt, Altman, Anadkat, Avila, Babuschkin, Balaji,
  Balcom, Baltescu, Bao, Bavarian, Belgum, Bello, Berdine, Bernadett-Shapiro,
  Berner, Bogdonoff, Boiko, Boyd, Brakman, Brockman, Brooks, Brundage, Button,
  Cai, Campbell, Cann, Carey, Carlson, Carmichael, Chan, Chang, Chantzis, Chen,
  Chen, Chen, Chen, Chen, Chess, Cho, Chu, Chung, Cummings, Currier, Dai,
  Decareaux, Degry, Deutsch, Deville, Dhar, Dohan, Dowling, Dunning, Ecoffet,
  Eleti, Eloundou, Farhi, Fedus, Felix, Fishman, Forte, Fulford, Gao, Georges,
  Gibson, Goel, Gogineni, Goh, Gontijo-Lopes, Gordon, Grafstein, Gray, Greene,
  Gross, Gu, Guo, Hallacy, Han, Harris, He, Heaton, Heidecke, Hesse, Hickey,
  Hickey, Hoeschele, Houghton, Hsu, Hu, Hu, Huizinga, Jain, Jain, Jang, Jiang,
  Jiang, Jin, Jin, Jomoto, Jonn, Jun, Kaftan, Łukasz Kaiser, Kamali,
  Kanitscheider, Keskar, Khan, Kilpatrick, Kim, Kim, Kim, Kirchner, Kiros,
  Knight, Kokotajlo, Łukasz Kondraciuk, Kondrich, Konstantinidis, Kosic,
  Krueger, Kuo, Lampe, Lan, Lee, Leike, Leung, Levy, Li, Lim, Lin, Lin, Litwin,
  Lopez, Lowe, Lue, Makanju, Malfacini, Manning, Markov, Markovski, Martin,
  Mayer, Mayne, McGrew, McKinney, McLeavey, McMillan, McNeil, Medina, Mehta,
  Menick, Metz, Mishchenko, Mishkin, Monaco, Morikawa, Mossing, Mu, Murati,
  Murk, Mély, Nair, Nakano, Nayak, Neelakantan, Ngo, Noh, Ouyang, O'Keefe,
  Pachocki, Paino, Palermo, Pantuliano, Parascandolo, Parish, Parparita,
  Passos, Pavlov, Peng, Perelman, de~Avila Belbute~Peres, Petrov,
  de~Oliveira~Pinto, Michael, Pokorny, Pokrass, Pong, Powell, Power, Power,
  Proehl, Puri, Radford, Rae, Ramesh, Raymond, Real, Rimbach, Ross, Rotsted,
  Roussez, Ryder, Saltarelli, Sanders, Santurkar, Sastry, Schmidt, Schnurr,
  Schulman, Selsam, Sheppard, Sherbakov, Shieh, Shoker, Shyam, Sidor, Sigler,
  Simens, Sitkin, Slama, Sohl, Sokolowsky, Song, Staudacher, Such, Summers,
  Sutskever, Tang, Tezak, Thompson, Tillet, Tootoonchian, Tseng, Tuggle,
  Turley, Tworek, Uribe, Vallone, Vijayvergiya, Voss, Wainwright, Wang, Wang,
  Wang, Ward, Wei, Weinmann, Welihinda, Welinder, Weng, Weng, Wiethoff,
  Willner, Winter, Wolrich, Wong, Workman, Wu, Wu, Wu, Xiao, Xu, Yoo, Yu, Yuan,
  Zaremba, Zellers, Zhang, Zhang, Zhao, Zheng, Zhuang, Zhuk, and
  Zoph}]{openai2024gpt4technicalreport}
OpenAI, Josh Achiam, Steven Adler, Sandhini Agarwal, Lama Ahmad, Ilge Akkaya,
  Florencia~Leoni Aleman, Diogo Almeida, Janko Altenschmidt, Sam Altman,
  Shyamal Anadkat, Red Avila, Igor Babuschkin, Suchir Balaji, Valerie Balcom,
  Paul Baltescu, Haiming Bao, Mohammad Bavarian, Jeff Belgum, Irwan Bello, Jake
  Berdine, Gabriel Bernadett-Shapiro, Christopher Berner, Lenny Bogdonoff, Oleg
  Boiko, Madelaine Boyd, Anna-Luisa Brakman, Greg Brockman, Tim Brooks, Miles
  Brundage, Kevin Button, Trevor Cai, Rosie Campbell, Andrew Cann, Brittany
  Carey, Chelsea Carlson, Rory Carmichael, Brooke Chan, Che Chang, Fotis
  Chantzis, Derek Chen, Sully Chen, Ruby Chen, Jason Chen, Mark Chen, Ben
  Chess, Chester Cho, Casey Chu, Hyung~Won Chung, Dave Cummings, Jeremiah
  Currier, Yunxing Dai, Cory Decareaux, Thomas Degry, Noah Deutsch, Damien
  Deville, Arka Dhar, David Dohan, Steve Dowling, Sheila Dunning, Adrien
  Ecoffet, Atty Eleti, Tyna Eloundou, David Farhi, Liam Fedus, Niko Felix,
  Simón~Posada Fishman, Juston Forte, Isabella Fulford, Leo Gao, Elie Georges,
  Christian Gibson, Vik Goel, Tarun Gogineni, Gabriel Goh, Rapha Gontijo-Lopes,
  Jonathan Gordon, Morgan Grafstein, Scott Gray, Ryan Greene, Joshua Gross,
  Shixiang~Shane Gu, Yufei Guo, Chris Hallacy, Jesse Han, Jeff Harris, Yuchen
  He, Mike Heaton, Johannes Heidecke, Chris Hesse, Alan Hickey, Wade Hickey,
  Peter Hoeschele, Brandon Houghton, Kenny Hsu, Shengli Hu, Xin Hu, Joost
  Huizinga, Shantanu Jain, Shawn Jain, Joanne Jang, Angela Jiang, Roger Jiang,
  Haozhun Jin, Denny Jin, Shino Jomoto, Billie Jonn, Heewoo Jun, Tomer Kaftan,
  Łukasz Kaiser, Ali Kamali, Ingmar Kanitscheider, Nitish~Shirish Keskar,
  Tabarak Khan, Logan Kilpatrick, Jong~Wook Kim, Christina Kim, Yongjik Kim,
  Jan~Hendrik Kirchner, Jamie Kiros, Matt Knight, Daniel Kokotajlo, Łukasz
  Kondraciuk, Andrew Kondrich, Aris Konstantinidis, Kyle Kosic, Gretchen
  Krueger, Vishal Kuo, Michael Lampe, Ikai Lan, Teddy Lee, Jan Leike, Jade
  Leung, Daniel Levy, Chak~Ming Li, Rachel Lim, Molly Lin, Stephanie Lin,
  Mateusz Litwin, Theresa Lopez, Ryan Lowe, Patricia Lue, Anna Makanju, Kim
  Malfacini, Sam Manning, Todor Markov, Yaniv Markovski, Bianca Martin, Katie
  Mayer, Andrew Mayne, Bob McGrew, Scott~Mayer McKinney, Christine McLeavey,
  Paul McMillan, Jake McNeil, David Medina, Aalok Mehta, Jacob Menick, Luke
  Metz, Andrey Mishchenko, Pamela Mishkin, Vinnie Monaco, Evan Morikawa, Daniel
  Mossing, Tong Mu, Mira Murati, Oleg Murk, David Mély, Ashvin Nair, Reiichiro
  Nakano, Rajeev Nayak, Arvind Neelakantan, Richard Ngo, Hyeonwoo Noh, Long
  Ouyang, Cullen O'Keefe, Jakub Pachocki, Alex Paino, Joe Palermo, Ashley
  Pantuliano, Giambattista Parascandolo, Joel Parish, Emy Parparita, Alex
  Passos, Mikhail Pavlov, Andrew Peng, Adam Perelman, Filipe de~Avila
  Belbute~Peres, Michael Petrov, Henrique~Ponde de~Oliveira~Pinto, Michael,
  Pokorny, Michelle Pokrass, Vitchyr~H. Pong, Tolly Powell, Alethea Power,
  Boris Power, Elizabeth Proehl, Raul Puri, Alec Radford, Jack Rae, Aditya
  Ramesh, Cameron Raymond, Francis Real, Kendra Rimbach, Carl Ross, Bob
  Rotsted, Henri Roussez, Nick Ryder, Mario Saltarelli, Ted Sanders, Shibani
  Santurkar, Girish Sastry, Heather Schmidt, David Schnurr, John Schulman,
  Daniel Selsam, Kyla Sheppard, Toki Sherbakov, Jessica Shieh, Sarah Shoker,
  Pranav Shyam, Szymon Sidor, Eric Sigler, Maddie Simens, Jordan Sitkin,
  Katarina Slama, Ian Sohl, Benjamin Sokolowsky, Yang Song, Natalie Staudacher,
  Felipe~Petroski Such, Natalie Summers, Ilya Sutskever, Jie Tang, Nikolas
  Tezak, Madeleine~B. Thompson, Phil Tillet, Amin Tootoonchian, Elizabeth
  Tseng, Preston Tuggle, Nick Turley, Jerry Tworek, Juan Felipe~Cerón Uribe,
  Andrea Vallone, Arun Vijayvergiya, Chelsea Voss, Carroll Wainwright,
  Justin~Jay Wang, Alvin Wang, Ben Wang, Jonathan Ward, Jason Wei, CJ~Weinmann,
  Akila Welihinda, Peter Welinder, Jiayi Weng, Lilian Weng, Matt Wiethoff, Dave
  Willner, Clemens Winter, Samuel Wolrich, Hannah Wong, Lauren Workman, Sherwin
  Wu, Jeff Wu, Michael Wu, Kai Xiao, Tao Xu, Sarah Yoo, Kevin Yu, Qiming Yuan,
  Wojciech Zaremba, Rowan Zellers, Chong Zhang, Marvin Zhang, Shengjia Zhao,
  Tianhao Zheng, Juntang Zhuang, William Zhuk, and Barret Zoph. 2024.
\newblock \href {http://arxiv.org/abs/2303.08774} {Gpt-4 technical report}.

\bibitem[{Palta and Rudinger(2023)}]{palta-rudinger-2023-fork}
Shramay Palta and Rachel Rudinger. 2023.
\newblock \href {https://doi.org/10.18653/v1/2023.findings-acl.631} {{FORK}: A
  bite-sized test set for probing culinary cultural biases in commonsense
  reasoning models}.
\newblock In \emph{Findings of the Association for Computational Linguistics:
  ACL 2023}, pages 9952--9962, Toronto, Canada. Association for Computational
  Linguistics.

\bibitem[{Pan et~al.(2024)Pan, Luo, Wang, Chen, Wang, and Wu}]{pan2024unifying}
Shirui Pan, Linhao Luo, Yufei Wang, Chen Chen, Jiapu Wang, and Xindong Wu.
  2024.
\newblock Unifying large language models and knowledge graphs: A roadmap.
\newblock \emph{IEEE Transactions on Knowledge and Data Engineering}.

\bibitem[{Park et~al.(2021)Park, Moon, Kim, Cho, Han, Park, Song, Kim, Song,
  Oh, Lee, Oh, Lyu, Jeong, Lee, Seo, Lee, Kim, Lee, Jang, Do, Kim, Lim, Lee,
  Park, Shin, Kim, Park, Oh, Ha, and Cho}]{park2021klue}
Sungjoon Park, Jihyung Moon, Sungdong Kim, Won~Ik Cho, Jiyoon Han, Jangwon
  Park, Chisung Song, Junseong Kim, Yongsook Song, Taehwan Oh, Joohong Lee,
  Juhyun Oh, Sungwon Lyu, Younghoon Jeong, Inkwon Lee, Sangwoo Seo, Dongjun
  Lee, Hyunwoo Kim, Myeonghwa Lee, Seongbo Jang, Seungwon Do, Sunkyoung Kim,
  Kyungtae Lim, Jongwon Lee, Kyumin Park, Jamin Shin, Seonghyun Kim, Lucy Park,
  Alice Oh, Jungwoo Ha, and Kyunghyun Cho. 2021.
\newblock \href {http://arxiv.org/abs/2105.09680} {Klue: Korean language
  understanding evaluation}.

\bibitem[{Peterson et~al.(2018)Peterson, S{\o}ndergaard, and
  Kara}]{peterson2018traversing}
Mark~F Peterson, Mikael S{\o}ndergaard, and Aycan Kara. 2018.
\newblock Traversing cultural boundaries in ib: The complex relationships
  between explicit country and implicit cultural group boundaries at multiple
  levels.
\newblock \emph{Journal of International Business Studies}, 49:1081--1099.

\bibitem[{Petroni et~al.(2019)Petroni, Rockt{\"a}schel, Riedel, Lewis, Bakhtin,
  Wu, and Miller}]{petroni-etal-2019-language}
Fabio Petroni, Tim Rockt{\"a}schel, Sebastian Riedel, Patrick Lewis, Anton
  Bakhtin, Yuxiang Wu, and Alexander Miller. 2019.
\newblock \href {https://doi.org/10.18653/v1/D19-1250} {Language models as
  knowledge bases?}
\newblock In \emph{Proceedings of the 2019 Conference on Empirical Methods in
  Natural Language Processing and the 9th International Joint Conference on
  Natural Language Processing (EMNLP-IJCNLP)}, pages 2463--2473, Hong Kong,
  China. Association for Computational Linguistics.

\bibitem[{Raffel et~al.(2020)Raffel, Shazeer, Roberts, Lee, Narang, Matena,
  Zhou, Li, and Liu}]{raffel2020exploring}
Colin Raffel, Noam Shazeer, Adam Roberts, Katherine Lee, Sharan Narang, Michael
  Matena, Yanqi Zhou, Wei Li, and Peter~J Liu. 2020.
\newblock Exploring the limits of transfer learning with a unified text-to-text
  transformer.
\newblock \emph{The Journal of Machine Learning Research}, 21(1):5485--5551.

\bibitem[{Safaya et~al.(2020)Safaya, Abdullatif, and
  Yuret}]{safaya-etal-2020-kuisail}
Ali Safaya, Moutasem Abdullatif, and Deniz Yuret. 2020.
\newblock \href {https://doi.org/10.18653/v1/2020.semeval-1.271} {{KUISAIL} at
  {S}em{E}val-2020 task 12: {BERT}-{CNN} for offensive speech identification in
  social media}.
\newblock In \emph{Proceedings of the Fourteenth Workshop on Semantic
  Evaluation}, pages 2054--2059, Barcelona (online). International Committee
  for Computational Linguistics.

\bibitem[{Savoldi et~al.(2021)Savoldi, Gaido, Bentivogli, Negri, and
  Turchi}]{savoldi-etal-2021-gender}
Beatrice Savoldi, Marco Gaido, Luisa Bentivogli, Matteo Negri, and Marco
  Turchi. 2021.
\newblock \href {https://doi.org/10.1162/tacl_a_00401} {Gender bias in machine
  translation}.
\newblock \emph{Transactions of the Association for Computational Linguistics},
  9:845--874.

\bibitem[{Shin et~al.(2020)Shin, Razeghi, Logan~IV, Wallace, and
  Singh}]{shin-etal-2020-autoprompt}
Taylor Shin, Yasaman Razeghi, Robert~L. Logan~IV, Eric Wallace, and Sameer
  Singh. 2020.
\newblock \href {https://doi.org/10.18653/v1/2020.emnlp-main.346}
  {{A}uto{P}rompt: {E}liciting {K}nowledge from {L}anguage {M}odels with
  {A}utomatically {G}enerated {P}rompts}.
\newblock In \emph{Proceedings of the 2020 Conference on Empirical Methods in
  Natural Language Processing (EMNLP)}, pages 4222--4235, Online. Association
  for Computational Linguistics.

\bibitem[{Singh et~al.(2024)Singh, Romanou, Fourrier, Adelani, Ngui,
  Vila-Suero, Limkonchotiwat, Marchisio, Leong, Susanto
  et~al.}]{singh2024global}
Shivalika Singh, Angelika Romanou, Cl{\'e}mentine Fourrier, David~I Adelani,
  Jian~Gang Ngui, Daniel Vila-Suero, Peerat Limkonchotiwat, Kelly Marchisio,
  Wei~Qi Leong, Yosephine Susanto, et~al. 2024.
\newblock Global mmlu: Understanding and addressing cultural and linguistic
  biases in multilingual evaluation.
\newblock \emph{arXiv preprint arXiv:2412.03304}.

\bibitem[{S{\o}gaard(2021)}]{sogaard-2021-lockes}
Anders S{\o}gaard. 2021.
\newblock \href {https://doi.org/10.18653/v1/2021.emnlp-main.649} {Locke{'}s
  holiday: Belief bias in machine reading}.
\newblock In \emph{Proceedings of the 2021 Conference on Empirical Methods in
  Natural Language Processing}, pages 8240--8245, Online and Punta Cana,
  Dominican Republic. Association for Computational Linguistics.

\bibitem[{Touvron et~al.(2023)Touvron, Martin, Stone, Albert, Almahairi,
  Babaei, Bashlykov, Batra, Bhargava, Bhosale et~al.}]{touvron2023llama}
Hugo Touvron, Louis Martin, Kevin Stone, Peter Albert, Amjad Almahairi, Yasmine
  Babaei, Nikolay Bashlykov, Soumya Batra, Prajjwal Bhargava, Shruti Bhosale,
  et~al. 2023.
\newblock Llama 2: Open foundation and fine-tuned chat models.
\newblock \emph{arXiv preprint arXiv:2307.09288}.

\bibitem[{Wang et~al.(2023)Wang, Haddow, Birch, and Peng}]{wang2023assessing}
Weixuan Wang, Barry Haddow, Alexandra Birch, and Wei Peng. 2023.
\newblock Assessing the reliability of large language model knowledge.
\newblock \emph{arXiv preprint arXiv:2310.09820}.

\bibitem[{Wu et~al.(2023)Wu, Jiang, Jiang, Xie, and Tu}]{wu-etal-2023-plms}
Weiqi Wu, Chengyue Jiang, Yong Jiang, Pengjun Xie, and Kewei Tu. 2023.
\newblock \href {https://doi.org/10.18653/v1/2023.acl-long.173} {Do {PLM}s know
  and understand ontological knowledge?}
\newblock In \emph{Proceedings of the 61st Annual Meeting of the Association
  for Computational Linguistics (Volume 1: Long Papers)}, pages 3080--3101,
  Toronto, Canada. Association for Computational Linguistics.

\bibitem[{Xue et~al.(2020)Xue, Constant, Roberts, Kale, Al-Rfou, Siddhant,
  Barua, and Raffel}]{xue2020mt5}
Linting Xue, Noah Constant, Adam Roberts, Mihir Kale, Rami Al-Rfou, Aditya
  Siddhant, Aditya Barua, and Colin Raffel. 2020.
\newblock mt5: A massively multilingual pre-trained text-to-text transformer.
\newblock \emph{arXiv preprint arXiv:2010.11934}.

\bibitem[{Yang et~al.(2024)Yang, Yang, Hui, Zheng, Yu, Zhou, Li, Li, Liu, Huang
  et~al.}]{yang2024qwen2}
An~Yang, Baosong Yang, Binyuan Hui, Bo~Zheng, Bowen Yu, Chang Zhou, Chengpeng
  Li, Chengyuan Li, Dayiheng Liu, Fei Huang, et~al. 2024.
\newblock Qwen2 technical report.
\newblock \emph{arXiv preprint arXiv:2407.10671}.

\bibitem[{Yin et~al.(2022)Yin, Bansal, Monajatipoor, Li, and
  Chang}]{yin-etal-2022-geomlama}
Da~Yin, Hritik Bansal, Masoud Monajatipoor, Liunian~Harold Li, and Kai-Wei
  Chang. 2022.
\newblock \href {https://doi.org/10.18653/v1/2022.emnlp-main.132}
  {{G}eo{MLAMA}: Geo-diverse commonsense probing on multilingual pre-trained
  language models}.
\newblock In \emph{Proceedings of the 2022 Conference on Empirical Methods in
  Natural Language Processing}, pages 2039--2055, Abu Dhabi, United Arab
  Emirates. Association for Computational Linguistics.

\bibitem[{Zhong et~al.(2021)Zhong, Friedman, and
  Chen}]{zhong-etal-2021-factual}
Zexuan Zhong, Dan Friedman, and Danqi Chen. 2021.
\newblock \href {https://doi.org/10.18653/v1/2021.naacl-main.398} {Factual
  probing is [{MASK}]: Learning vs. learning to recall}.
\newblock In \emph{Proceedings of the 2021 Conference of the North American
  Chapter of the Association for Computational Linguistics: Human Language
  Technologies}, pages 5017--5033, Online. Association for Computational
  Linguistics.

\bibitem[{Zhou et~al.(2023)Zhou, Karamolegkou, Chen, and
  Hershcovich}]{zhou-etal-2023-cultural}
Li~Zhou, Antonia Karamolegkou, Wenyu Chen, and Daniel Hershcovich. 2023.
\newblock \href {https://doi.org/10.18653/v1/2023.findings-emnlp.845} {Cultural
  compass: Predicting transfer learning success in offensive language detection
  with cultural features}.
\newblock In \emph{Findings of the Association for Computational Linguistics:
  EMNLP 2023}, pages 12684--12702, Singapore. Association for Computational
  Linguistics.

\end{thebibliography}

\appendix

\begin{figure*}[t]
    \centering
    \includegraphics[width=1\linewidth, trim=0cm 6.0cm 0cm 0cm, clip]{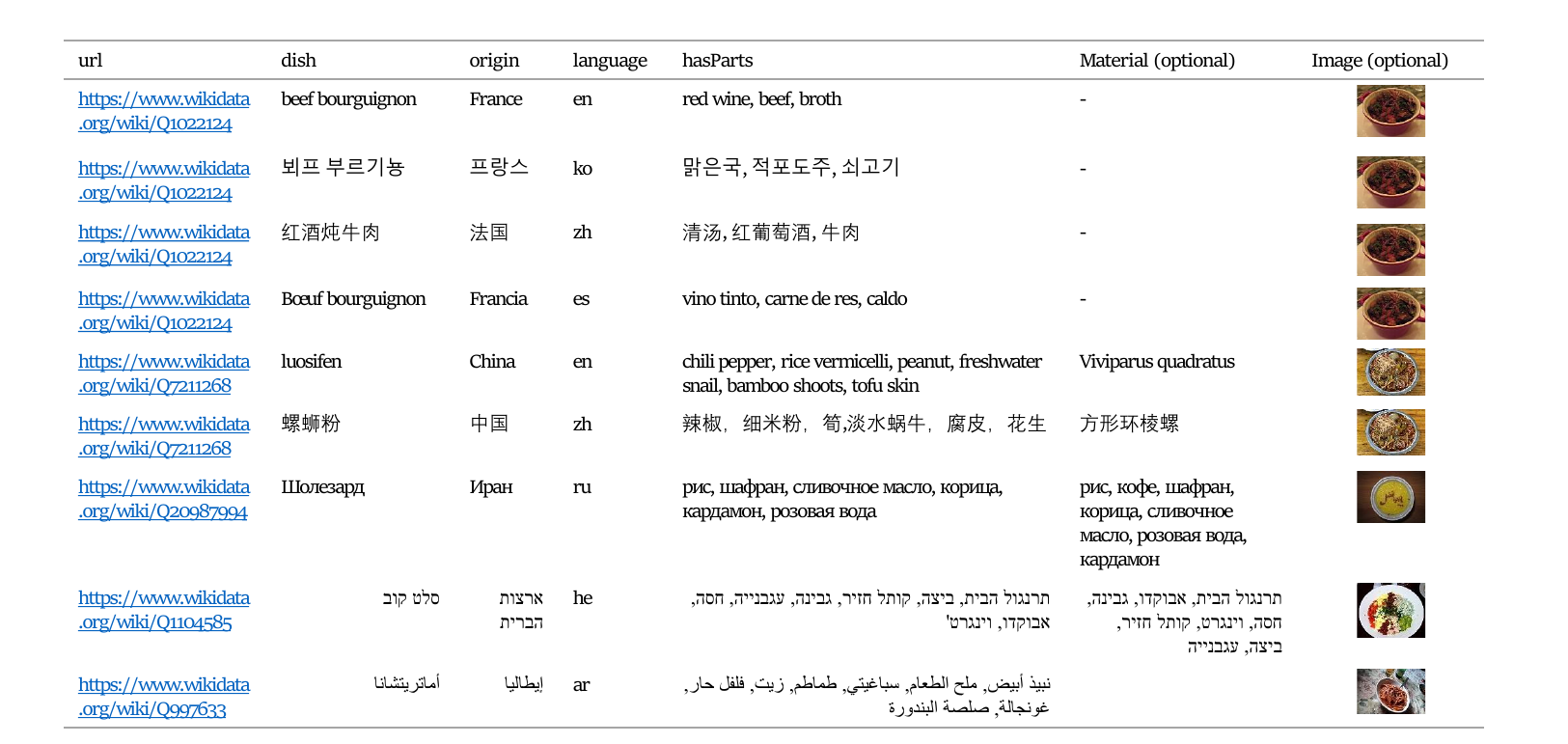}
    \caption{Examples of \textsc{FmLAMA}.}
    \label{fig:FmLAMA_example}
\end{figure*}

\section{\textsc{FmLAMA} details}
\label{app:dataset}

\begin{table*}[t]
\centering
    \centering
    \scalebox{0.65}{
    \begin{tabular}{lllll}
    \toprule
     \textbf{Datasets}    & \textbf{Format}  & \textbf{Topic} & \textbf{Construction method}  & \textbf{Size} \\
     \midrule
     \textsc{GeoMLama} \small~\cite{yin-etal-2022-geomlama} & Manual Template & Geo-Diverse Concept & Manually curated & 3125 \\
     FORK\small~\cite{palta-rudinger-2023-fork} & CommonsenseQA  & Culinary culture & Manually curated   & 184 \\
     StereoKG \small~\cite{deshpande-etal-2022-stereokg}  &  Triplet knowledge   &  Stereotypes about religion and ethnicity &  Automatically constructed  & 4722 \\
     \textsc{Candle} \small~\cite{nguyen2023extracting}  & Sentences  & \multirow{2}{7cm}{Several
cultural facets (food, drinks, clothing, traditions, rituals, behaviors)} & Automatically constructed & 47360 \\  \\
     
     FmLAMA (ours) & Triplet knowledge  &  Food domain   &  Automatically constructed    & 33600   \\
     \bottomrule
    \end{tabular}}
    \caption{Comparison of cultural knowledge datasets. Size is in number of instances.}
    \label{tab:culturalKB}
\end{table*}

\subsection{Examples}
\label{app-sec:example}

Figure~\ref{fig:FmLAMA_example} illustrates examples from dataset \textsc{FmLAMA}.
Each dish instance in \textsc{FmLAMA} is defined as: $\left( url ,na ,cou,la,pa,\left[ ma,im \right] \right) $, the elements in $[\cdot]$ indicate optional:
\begin{itemize}
\setlength{\itemsep}{-0.1\baselineskip}
    \item $url$: the link in Wikidata;
    \item $na$: the name of the dish; 
    \item $cou$: the country of origin of the dish;
    \item $la$: the language used in this entry; 
    \item $pa$: the ingredients of this dish;
    \item $ma$: the material used in the dish;
    \item $im$: the image of this dish.
\end{itemize}
Particularly, in each instance, $la=\mathrm{LANG}\left( na \right) =\mathrm{LANG}\left( pa \right) =\mathrm{LANG}\left( ma \right)$, which indicates the descriptive language for all attributes remains consistent.
For a dish with the same $url$, there may be several instances with different languages. Table~\ref{tab:culturalKB} contrasts our dataset with prior cultural knowledge collections.

\subsection{Statistics}
\begin{figure*}[t]
  \centering
  \begin{subfigure}{0.45\linewidth}
    \centering
    \includegraphics[width=\linewidth]{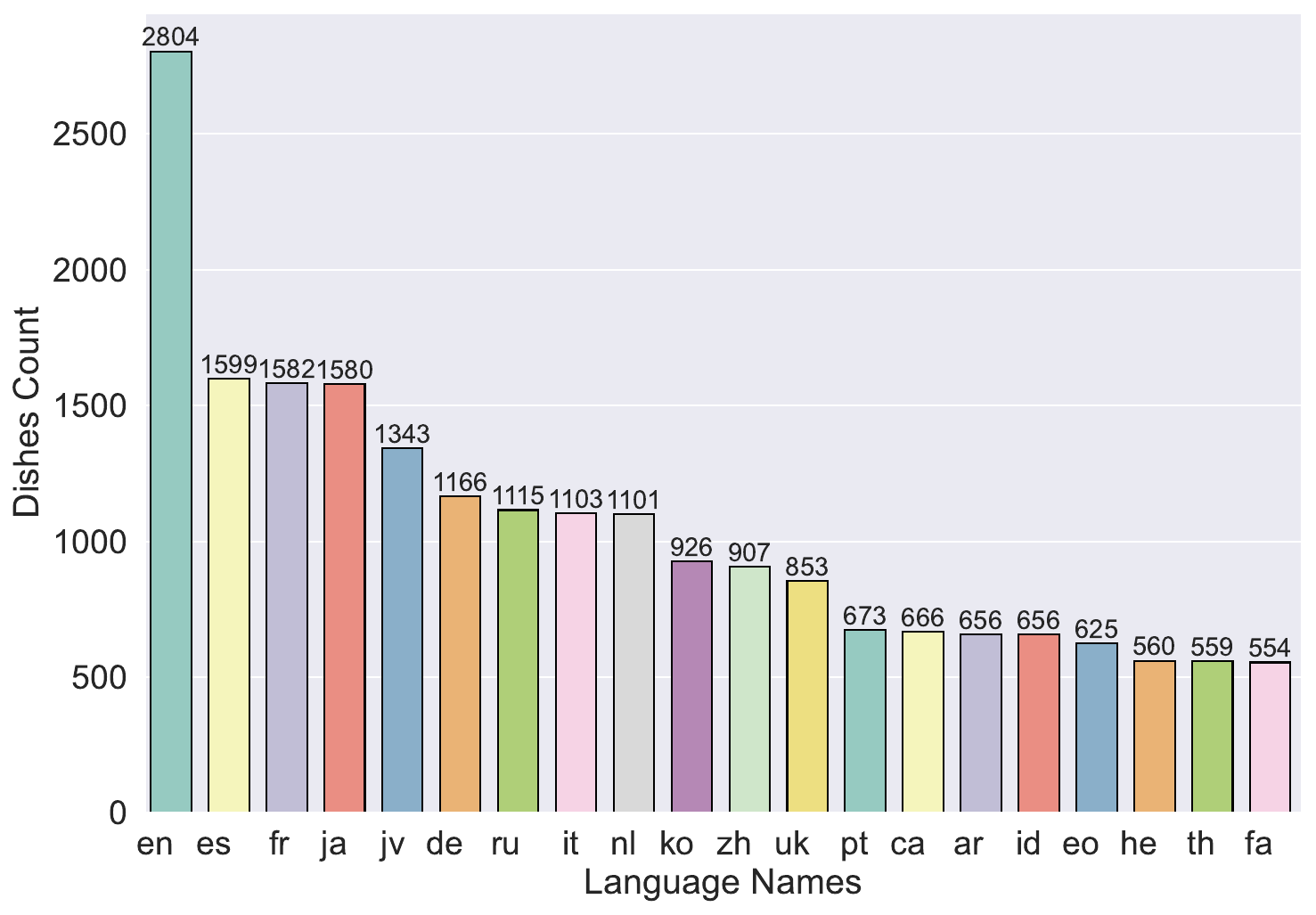}
    \caption{The Top20 languages ranked by the dishes count}
    \label{fig:top20_lang}
  \end{subfigure}
  \begin{subfigure}{0.45\linewidth}
    \centering
    \includegraphics[width=\linewidth]{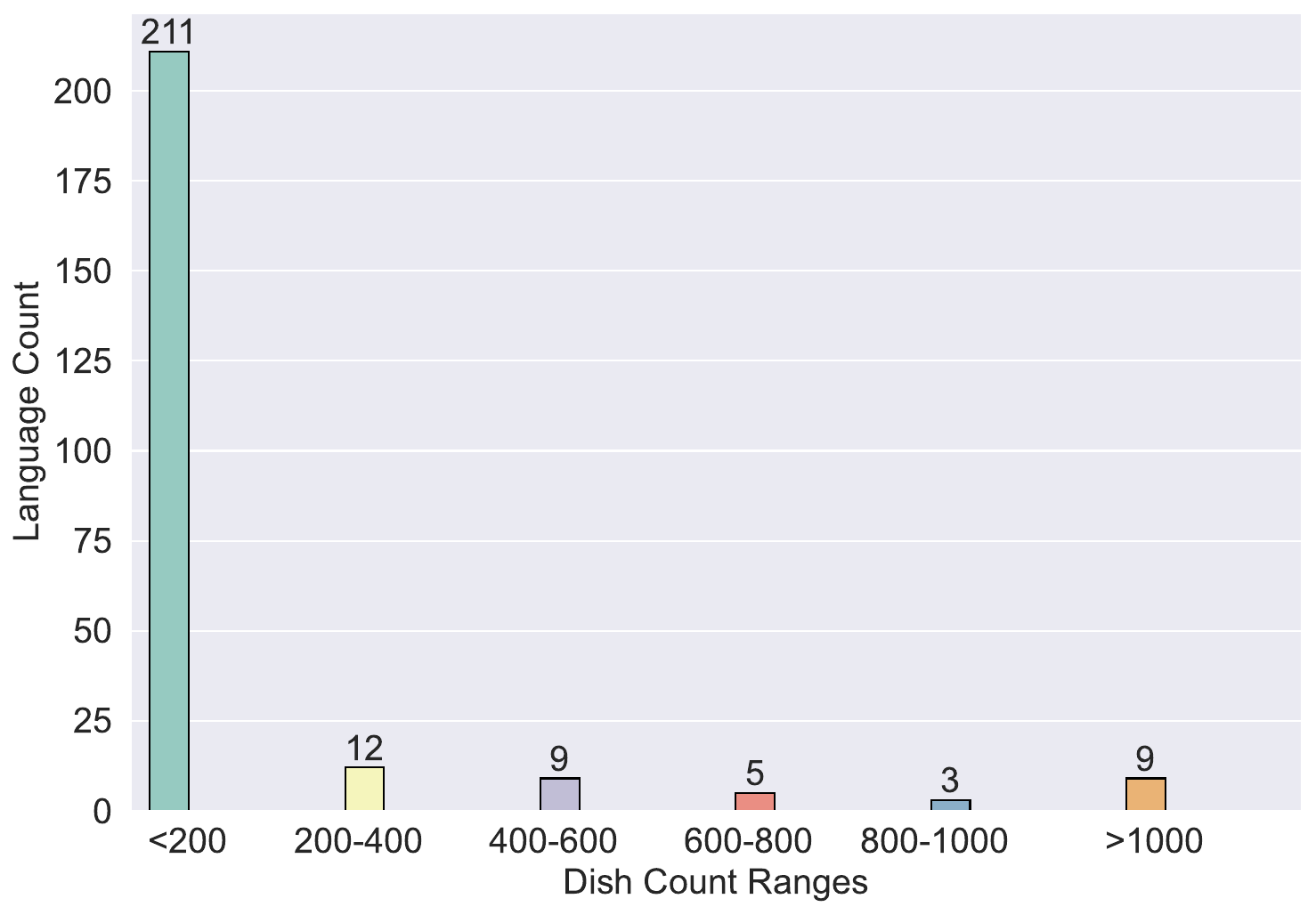}
    \caption{Languages distributed across various dish count ranges}
    \label{fig:distributation_lang}
  \end{subfigure}
  \caption{Dish statistics by languages}
  \label{fig:top20_lang_dis}
\end{figure*}

\begin{figure*}[t]
  \centering
  \begin{subfigure}{0.45\linewidth}
    \centering
    \includegraphics[width=\linewidth]{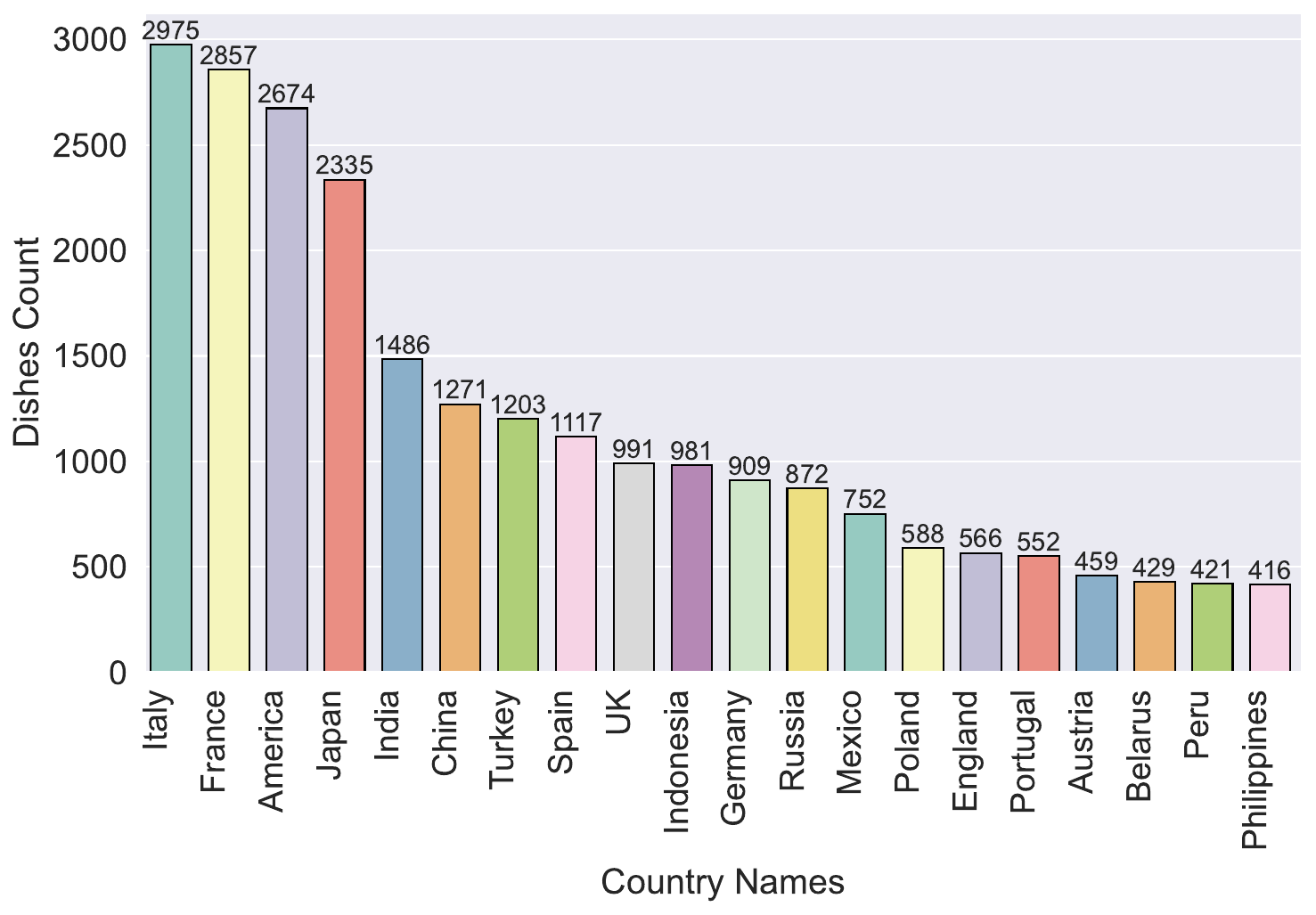}
    \caption{The Top20 countries ranked by the dishes count}
    \label{fig:top20_country}
  \end{subfigure}
  \begin{subfigure}{0.45\linewidth}
    \centering
    \includegraphics[width=\linewidth]{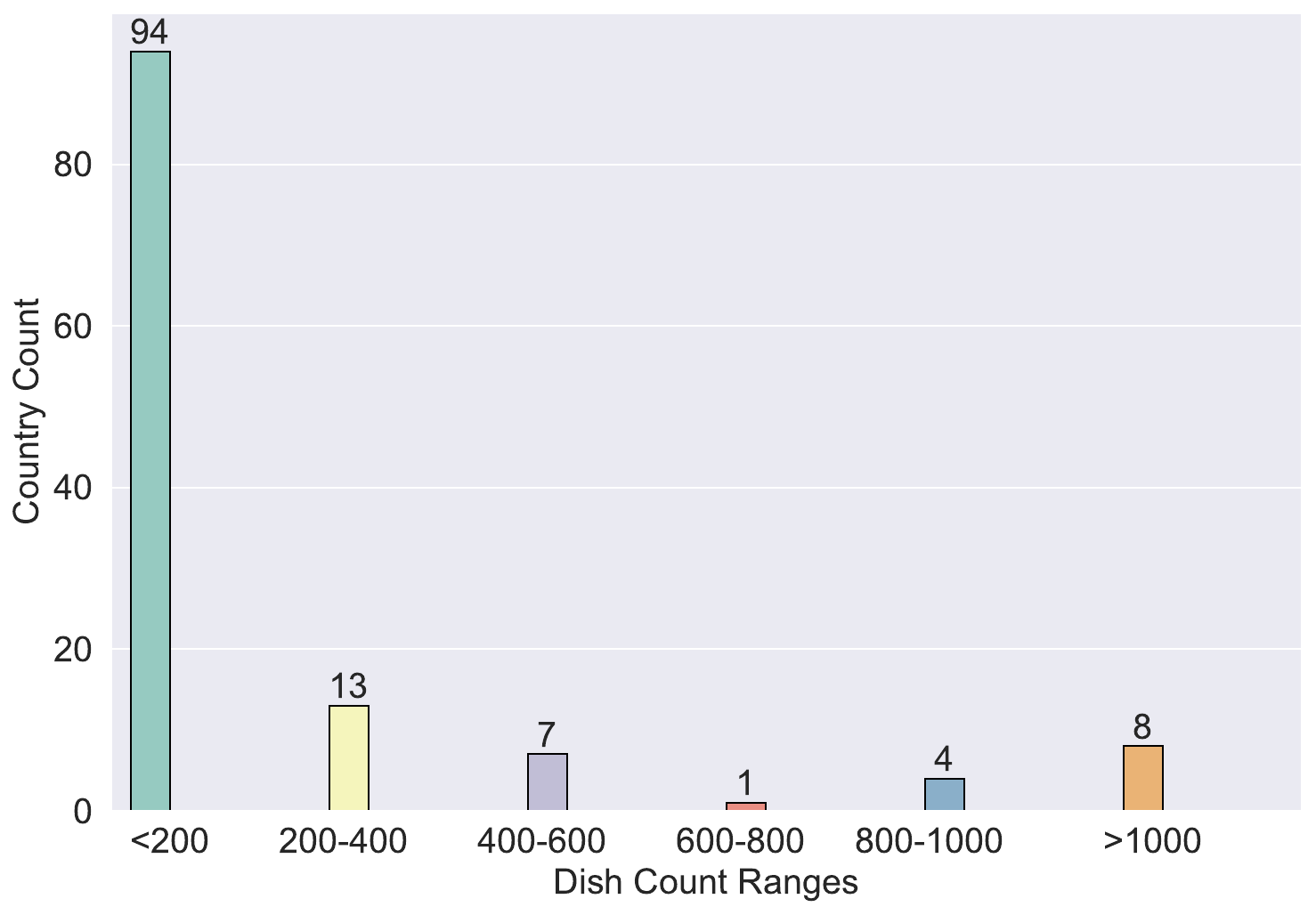}
    \caption{Countries distributed across various dish count ranges}
    \label{fig:distributation_country}
  \end{subfigure}
  \caption{Dish statistics by countries}
  \label{fig:top20_country_dis}
\end{figure*}

\begin{figure}[t]
    \centering
    \includegraphics[width=0.9\linewidth]{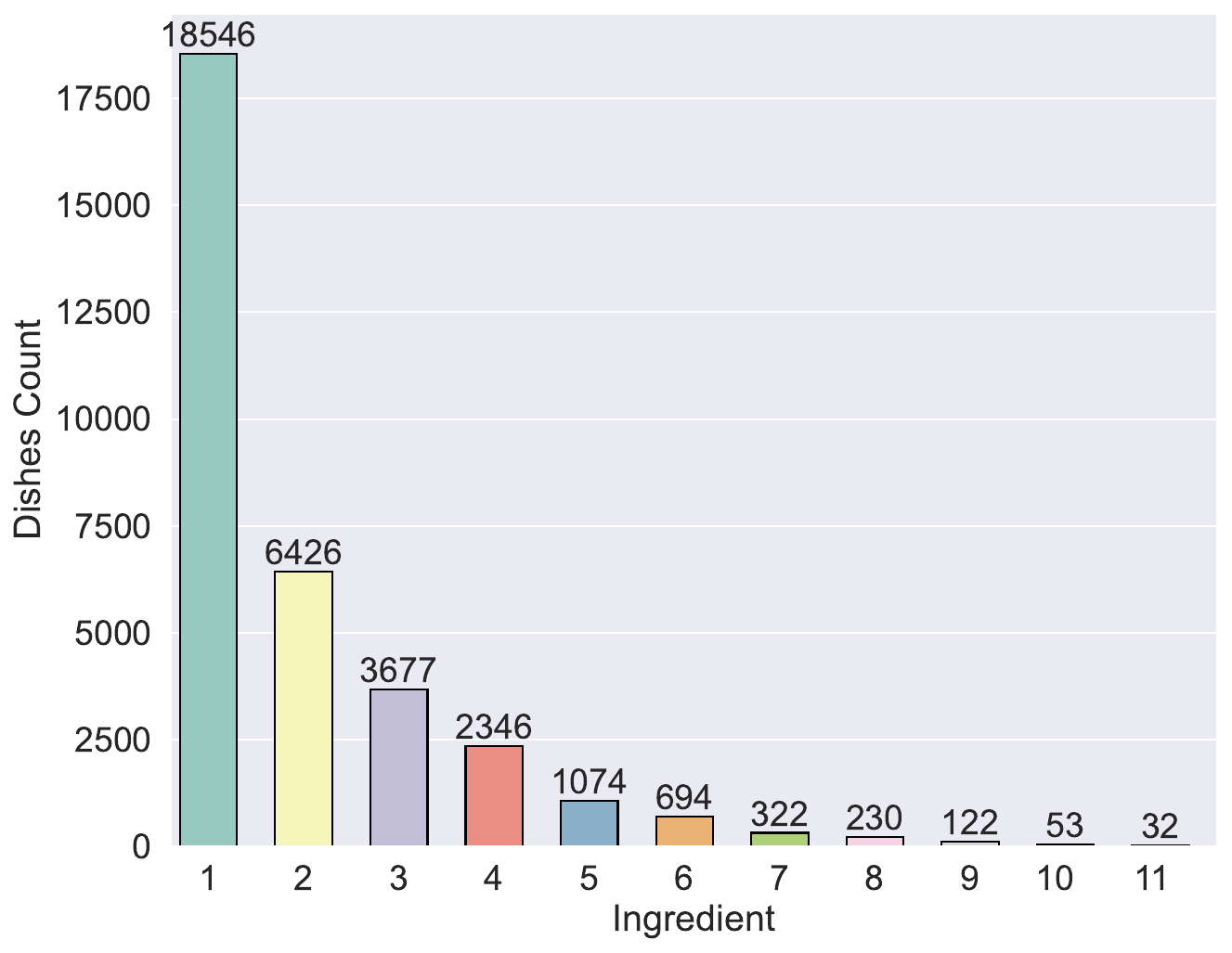}
    \caption{Dish statistics by ingredient count}
    \label{fig:ingredient}
\end{figure}

There are a total of 33,600 dishes in \textsc{FmLAMA}. 
We count the number of dishes corresponding to each language and country in the dataset, as shown in Figures~\ref{fig:top20_lang_dis} and \ref{fig:top20_country_dis}. 
Figure~\ref{fig:top20_lang} shows the top 20 languages ranked by the number of dishes, with English (en) having the highest count at 2804, followed by Spanish (es), French (fr), and Japanese (ja). Figure~\ref{fig:distributation_lang} displays the distribution of dish counts across different intervals for all languages, indicating that most languages have fewer than 200 dishes.
Similarly, Figure~\ref{fig:top20_country} shows the top 20 countries ranked by the number of dishes, with Italy having the highest count at 2975, followed by France, America, and Japan. Figure~\ref{fig:distributation_country} shows the distribution of dish counts across various intervals for all countries, indicating that only 8 countries have more than 200 dishes.
Figure~\ref{fig:ingredient} shows the dish statistics by ingredient count. 
The majority of dishes contain only one ingredient, totaling 18,546. Dishes with two ingredients are the second most common, with a count of 6,426. The counts decrease as the number of ingredients increases, with dishes containing eleven ingredients being the least common, at 32. 

\section{Probing Templates}
\label{app:templates}

\begin{figure*}[t]
    \centering
    \includegraphics[width=1\linewidth]{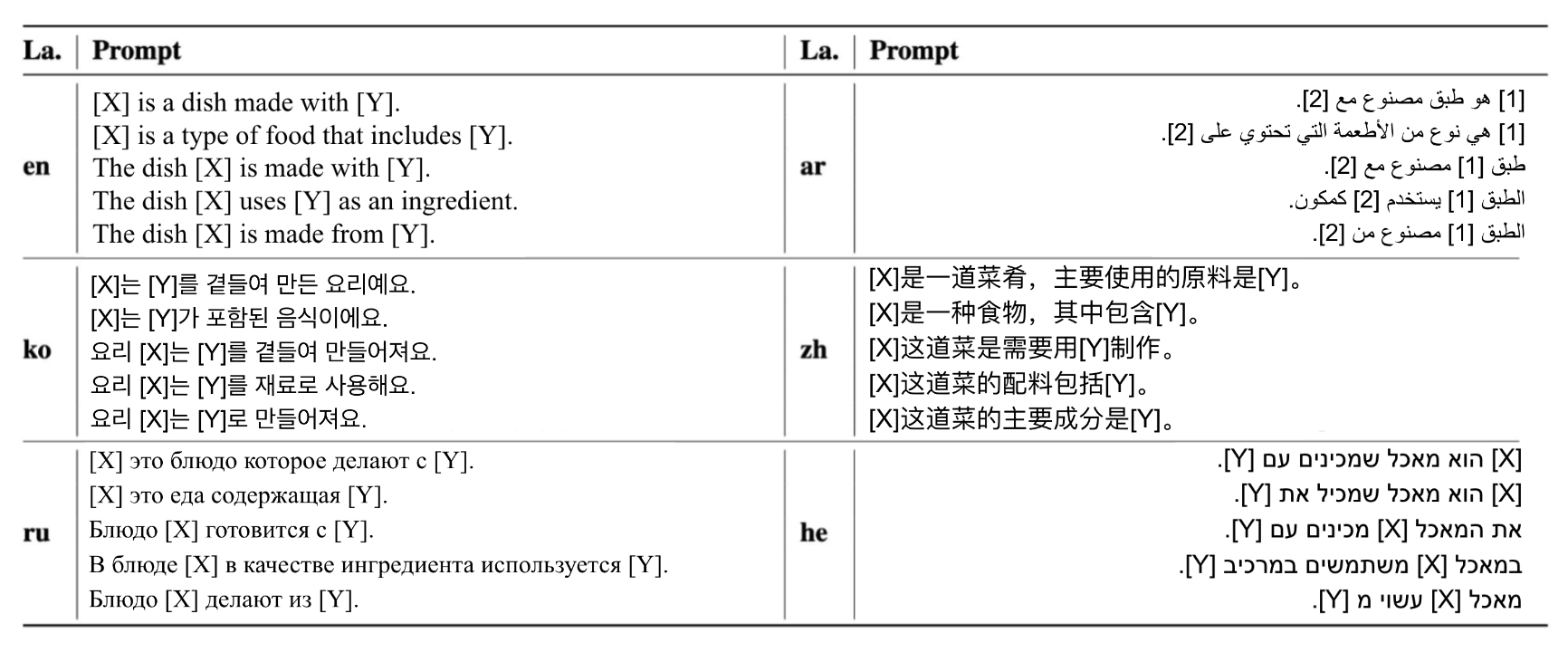}
    \caption{Probing templates in six involved languages, with [X] representing the subject and [Y] indicating the object that can be substituted.}
    \label{fig:probing_template}
\end{figure*}

\subsection{Prompt illustration}

Probing templates in six involved languages are shown in Figure~\ref{fig:probing_template}, in which [X] represents the subject (dish) and [Y] indicates the object (ingredient).

\subsection{Language comparison}
\label{app:languages_prompt}
\subsubsection{Geographic differences}
In this section, we discuss the representative countries where each selected language is spoken and their geographic distribution:
(1) \textit{English}: Mainly spoken in  U.S., Canada, Australia, Guyana, etc., spanning North and South America, Europe, Australia, and Africa.
(2) \textit{Chinese}: Mainly spoken in  China, Singapore, etc., covering huge area of Asia.
(3) \textit{Arabic}: Mainly spoken in  Saudi Arabia, Egypt, United Arab Emirates, etc., covering the Arabian Peninsula, North Africa, and some sub-Saharan African countries.
(4) \textit{Korean}: Mainly spoken in South Korea, North Korea.
(5) \textit{Russian}: Mainly spoken in  Russia, Belarus, Kazakhstan, etc., covering Eastern Europe, Central Asia, and Northern Asia.
(6) \textit{Hebrew}: Mainly spoken in Israel, with Hebrew-speaking communities worldwide.
These languages represent cultural diversity, being spoken on different continents by groups with rich and distinct cultural backgrounds, demonstrating the broad geographic and cultural diversity represented by these languages.

\subsubsection{Grammatical differences}
In addition to geographical differences in usage, these six languages also display the following grammatical and reading distinctions:
(1) \textit{Word Order}: English and Chinese follow an 
Subject-Verb-Object word order, Arabic typically uses Verb-Subject-Object, Korean follows Subject-Object-Verb, while Russian and Hebrew are relatively flexible but generally adhere to Subject-Verb-Object.
(2) \textit{Tense}: English, Arabic, Russian, and Hebrew possess complex tense systems. Chinese lacks a strict tense structure, whereas Korean expresses tense through verb conjugation.
(3) \textit{Gender}: Arabic, Russian, and Hebrew have grammatical gender, while English distinguishes gender only in pronouns. Chinese and Korean do not have grammatical gender.
(4) \textit{Articles}: English, Arabic, and Hebrew have an article system, whereas Chinese, Korean, and Russian do not use articles.
(5) \textit{Reading Direction}: English, modern Chinese, modern Korean, and Russian are read from left to right, while Arabic and Hebrew are read from right to left. Traditional Chinese and traditional Korean are read vertically from top to bottom, with columns arranged from right to left.
(6) \textit{Case System}: Arabic and Russian have complex case systems, while Korean employs particles to indicate grammatical relations. English, Chinese, and Hebrew do not use case inflections.

\section{Manual Evaluation Setup}
\label{app:manual-evaluation}
\subsection{GPT-4o simulator}
The evaluation prompts for the GPT-4o simulator are shown in Table~\ref{tab:gpt4o-simulator}.

\subsection{Human evaluator}

\begin{figure*}[t]
    \centering
    \includegraphics[width=0.8\linewidth, trim=0 10 0 0, clip] {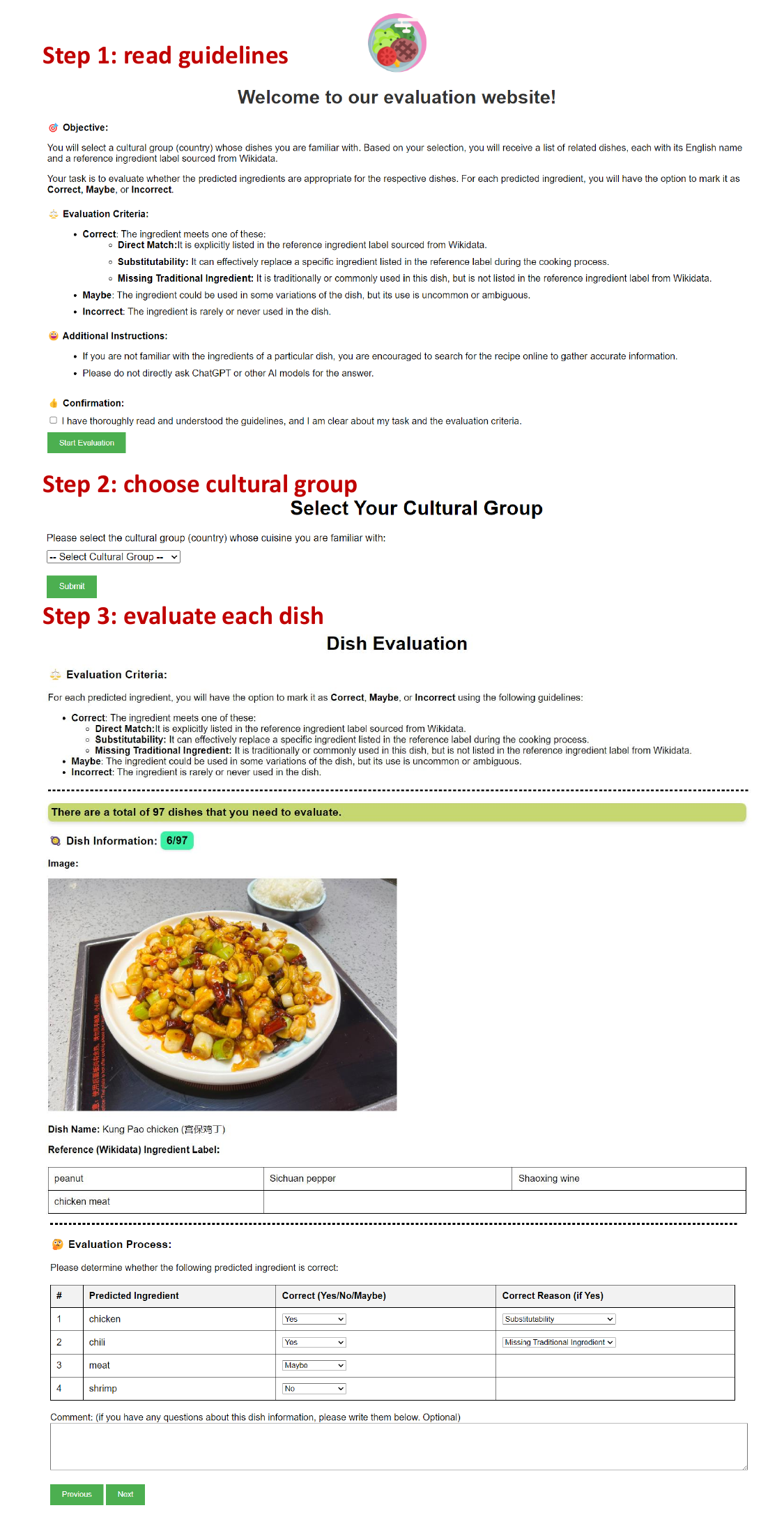}
    \caption{Human evaluation website}
    \label{fig:website}
\end{figure*}

Figure~\ref{fig:website} shows the human evaluation website. 
The Inter-Rater Reliability (IRR) and the average evaluation time for the three evaluators from each country group are provided in Table~\ref{tab:IRR}.

\begin{table*}[t]
\centering
\begin{tabular}{|p{16cm}|}
\hline
\textbf{Instructions} \\
\hline
You are a professional evaluator. I will provide you with a dish, its reference ingredient label sourced from Wikidata, and a list of predicted ingredients.
Your task is to evaluate whether each predicted ingredient is appropriate for the respective dish. \\
\\
\textbf{\#\#\# Evaluation Criteria:} \\
For each predicted ingredient, mark it as \textbf{Correct}, \textbf{Maybe}, or \textbf{Incorrect} using the following guidelines:\\

\begin{itemize}
    \item \textbf{Correct}: The ingredient meets one of these:
    \begin{itemize}
        \item \textbf{Direct Match}: It is explicitly listed in the reference ingredient label from Wikidata.
        \item \textbf{Substitutability}: It can replace a specific ingredient in the reference label during cooking.
        \item \textbf{Missing Traditional Ingredient}: It is traditionally or commonly used in the dish, but not listed in the reference label.
    \end{itemize}
    \item \textbf{Maybe}: The ingredient could be used in some variations of the dish, but its use is uncommon or ambiguous.
    \item \textbf{Incorrect}: The ingredient is rarely or never used in the dish.
\end{itemize}

\textbf{\#\#\# Output Format:} \\
Please directly return the evaluation as a Python dictionary with JSON format where the key is the predicted ingredient, and the value is a tuple consisting of:\\
1. The evaluation ("Correct", "Maybe", "Incorrect")\\
2. The reason ("Direct Match", "Substitutability", "Missing Traditional Ingredient", or a brief explanation)\\

\\
\textbf{\#\#\# Example:} \\

\texttt{Input:}\\
\texttt{Dish: Spaghetti Bolognese}\\
\texttt{Reference ingredient label: spaghetti, ground beef, onion, garlic, Carrot}\\
\texttt{Predicted ingredients: spaghetti, basil, chicken, tomato, beef}\\
\\
\texttt{Output}

\begin{verbatim}
{
    'ground beef': ('Correct', 'Substitutability'),
    'meat': ('Maybe', 'Ambiguous'),
    'beef': ('Correct', 'Substitutability'),
    'pasta': ('Incorrect', 'Pasta is not part of the sauce itself'),
    'turkey meat': ('Maybe', 'It can be used in some variations'),
}
\end{verbatim} \\
\\
\hline 
\end{tabular}
\caption{GPT-4o simulator: Evaluation Instructions}
\label{tab:gpt4o-simulator}
\end{table*}

\begin{table}[t]
\centering
\begin{tabular}{@{}lrrr@{}}
\toprule
\textbf{Group}   & \textbf{Dishes} & \textbf{IRR} & \textbf{Time(m)} \\ \midrule
\textbf{Italy}   & 215             & 0.60         & 74.93            \\
\textbf{U.S.}    & 285             & 0.48         & 104.25           \\
\textbf{Turkey}  & 98              & 0.78         & 32.05            \\
\textbf{Japan}   & 186             & 0.64         & 49.25            \\
\textbf{France}  & 175             & 0.63         & 71.83            \\
\textbf{U.K.}    & 83              & 0.46         & 30.33            \\
\textbf{Mexico}  & 57              & 0.44         & 22.25            \\
\textbf{India}   & 132             & 0.58         & 40.25            \\
\textbf{Germany} & 57              & 0.69         & 20.03            \\
\textbf{China}   & 97              & 0.59         & 48.50            \\
\textbf{Iran}    & 21              & 0.61         & 13,33            \\
\textbf{Greece}  & 21              & 0.35         & 11.30            \\
\textbf{Spain}   & 95              & 0.57         & 41.50            \\
\textbf{Russia}  & 27              & 0.51         & 26.50            \\ \midrule
\textbf{Mean}    & 111             & 0.57         & -                \\ \bottomrule
\end{tabular}
\caption{Human evaluator: IRR and average evaluation time for the three evaluators from each country group.}
\label{tab:IRR}
\end{table}

\section{Supplementary Experiments and Analyses}
\subsection{Probing results in each \textsc{FmLAMA}-\textit{la}}
\label{app:language_results}


In addition to the multilingual LLMs discussed in \S\ref{sec:experiments}, configure probing for monolingual, encoder-only LLMs in Arabic, Russian, Korean, and Chinese, including \texttt{asafaya/bert-base-arabic}~\cite{safaya-etal-2020-kuisail}, \texttt{DeepPavlov/rubert-base-cased}~\cite{kuratov2019adaptation}, \texttt{kykim/bert-kor-base}, \texttt{klue/bert-base}~\cite{park2021klue} and \texttt{bert-base-chinese}.

The probing results with prompts in the other five languages (Arabic, Chinese, Hebrew, Korean, and Russian) on the corresponding filtered sub-datasets (\textsc{FmLAMA}-\textit{ar}, \textsc{FmLAMA}-\textit{zh}, \textsc{FmLAMA}-\textit{he}, \textsc{FmLAMA}-\textit{ko}, and \textsc{FmLAMA}-\textit{ru}) are depicted in Table~\ref{tab:ar_without},~\ref{tab:zh_main_results},~\ref{tab:he_without},~\ref{tab:ko_without}, and~\ref{tab:ru_without}, respectively.
Because certain objects in Chinese and Korean have representation vectors that result in all zeros when obtained through Fasttext, calculating cosine similarity was not feasible. Consequently, mWS evaluation was not conducted for prompts in Chinese and Korean prompts.



\begin{table*}[]
\centering
\scalebox{0.7}{
\begin{tabular}{@{}l|rrr|rr|rrr|r@{}}
\toprule
\multirow{2}{*}{\textbf{Origin}} & \multicolumn{3}{c|}{\textbf{Encoder-only LLMs}}                                                                                & \multicolumn{2}{c|}{\textbf{Encoder-Decoder LLMs}}                                  & \multicolumn{3}{c|}{\textbf{Decoder-only LLMs}}                                                                                & \multirow{2}{*}{\textbf{Avg.}} \\ \cmidrule(lr){2-9}
                                 & \textbf{Bb}                              & \textbf{Bl}                              & \textbf{mB}                              & \textbf{mT5}                             & \textbf{T5}                              & \textbf{Qwen2}                           & \textbf{Llama2}                          & \textbf{Llama3}                          &                                   \\ \midrule
\textbf{Italy}                   & 0.7393±0.01                              & 0.7425±0.01                              & 0.7438±0.00                              & 0.6865±0.02                              & 0.6579±0.01                              & 0.7749±0.01                              & 0.7658±0.01                              & 0.7869±0.01                              & 0.7372                            \\
\textbf{U.S.}                    & \textbf{0.7748±0.01}    & \textbf{0.7763±0.01}    & \textbf{0.7815±0.01}    & 0.7071±0.02                              & \textbf{0.7178±0.03}    & \underline{0.7968±0.01} & 0.7839±0.02                              & \underline{0.8107±0.01} & 0.7686                            \\
\textbf{Turkey}                  & 0.7357±0.01                              & 0.7461±0.01                              & 0.7545±0.01                              & 0.7015±0.02                              & 0.6703±0.01                              & 0.7681±0.01                              & 0.7571±0.01                              & 0.7830±0.01                              & 0.7395                            \\
\textbf{Japan}                   & 0.7113±0.00                              & 0.7144±0.00                              & 0.7218±0.01                              & 0.6788±0.02                              & 0.6442±0.01                              & 0.7446±0.00                              & 0.7446±0.01                              & 0.7516±0.00                              & 0.7139                            \\
\textbf{France}                  & 0.7398±0.01                              & 0.7520±0.02                              & 0.7513±0.01                              & 0.6755±0.02                              & 0.6642±0.01                              & 0.7941±0.01                              & 0.7796±0.01                              & 0.7846±0.00                              & 0.7426                            \\
\textbf{U.K.}                    & 0.7460±0.01                              & 0.7500±0.01                              & 0.7550±0.00                              & 0.6697±0.02                              & 0.6745±0.02                              & 0.7678±0.01                              & 0.7701±0.01                              & 0.7790±0.01                              & 0.7390                            \\
\textbf{Mexico}                  & 0.7273±0.01                              & 0.7313±0.01                              & 0.7291±0.01                              & 0.7171±0.01                              & 0.6734±0.01                              & 0.7605±0.01                              & 0.7752±0.02                              & 0.7852±0.02                              & 0.7374                            \\
\textbf{India}                   & 0.7477±0.01                              & 0.7523±0.00                              & 0.7568±0.00                              & 0.6998±0.01                              & 0.6619±0.01                              & 0.7867±0.00                              & \textbf{0.7944±0.01}    & 0.7935±0.01                              & 0.7491                            \\
\textbf{Germany}                 & 0.7348±0.01                              & 0.7305±0.03                              & 0.7493±0.00                              & 0.6755±0.01                              & 0.6535±0.01                              & 0.7835±0.01                              & 0.7769±0.02                              & 0.7849±0.01                              & 0.7361                            \\
\textbf{China}                   & 0.7273±0.01                              & 0.7319±0.01                              & 0.7317±0.01                              & 0.7024±0.01                              & \underline{0.7054±0.01} & 0.7698±0.01                              & 0.7593±0.01                              & 0.7793±0.01                              & 0.7384                            \\
\textbf{Iran}                    & 0.7064±0.01                              & 0.7013±0.01                              & 0.7310±0.01                              & 0.7042±0.02                              & 0.6686±0.01                              & 0.7905±0.03                              & \underline{0.7914±0.04} & 0.8026±0.01                              & 0.7370                            \\
\textbf{Greece}                  & \underline{0.7488±0.01} & \underline{0.7618±0.02} & 0.7566±0.02                              & \underline{0.7190±0.03} & 0.6369±0.00                              & \textbf{0.8137±0.04}    & 0.7739±0.01                              & \textbf{0.8247±0.01}    & 0.7544                            \\
\textbf{Spain}                   & 0.7402±0.00                              & 0.7465±0.01                              & 0.7472±0.00                              & 0.6883±0.01                              & 0.6391±0.01                              & 0.7748±0.01                              & 0.7663±0.01                              & 0.7765±0.01                              & 0.7349                            \\
\textbf{Russia}                  & 0.7467±0.01                              & 0.7571±0.01                              & \underline{0.7573±0.00} & \textbf{0.7216±0.02}    & 0.6710±0.01                              & 0.7716±0.01                              & 0.7680±0.01                              & 0.7721±0.00                              & 0.7457                            \\ \midrule
\textbf{ALL}                     & 0.7419±0.00                              & 0.7466±0.01                              & 0.7507±0.00                              & 0.6924±0.01                              & 0.6723±0.01                              & 0.7778±0.01                              & 0.7711±0.01                              & 0.7859±0.00                              & 0.7423                            \\ \bottomrule
\textbf{Corr.}                     & 0.35                              & 0.59                              & 0.38                              & 0.83                              & 0.57                             & 0.86                              & 0.71                             & 0.96                              & -                            \\ \bottomrule
\end{tabular}}
\caption{Probing performance comparison with \textbf{English} prompts and \textbf{\textsc{FmLAMA}-\textit{ar}} sub-dataset. The results are evaluated using mWS with \textbf{BERT base}. \textit{Corr.} denotes the correlation of mWS evaluations using BERT Base and FastText for each LLM, which reveal a strong correlation on decoder-only LLMs,}
\label{tab: en_without_BERT_mWS}
\end{table*}

\begin{table*}[t]
  \begin{subtable}{1\linewidth}
    \centering
    \scalebox{0.8}{
\begin{tabular}{lr|rrrrrr|r}
\hline
\textbf{Origin}  & \textbf{Count} & \textbf{Bb-ar}                         & \textbf{mB}                            & \textbf{mT5}                           & \textbf{Qwen2}                      & \textbf{Llama2}                    & \textbf{Llama3}                    & \textbf{Avg.} \\ \hline
\textbf{Italy}   & 76 (19.2\%)    & 2.56±0.69                              & 3.07±1.02                              & 5.12±1.20                              & 5.13±3.42                           & 3.30±1.13                          & 4.05±0.81                          & 4.03          \\
\textbf{U.S.}    & 54 (13.7\%)    & 5.29±1.18                              & 7.22±3.91                              & 5.68±1.53                              & 9.41±4.12                           & 3.93±1.00                          & 5.95±2.38                          & 5.09          \\
\textbf{Turkey}  & 52 (13.2\%)    & 4.07±0.88                              & 6.66±3.09                              & 3.30±0.53                              & 9.20±5.71                           & 3.33±1.04                          & \textbf{9.54±3.48}                 & 5.47          \\
\textbf{Japan}   & 44 (11.11\%)   & 2.69±1.50                              & 2.56±0.50                              & 2.13±0.78                              & 9.91±3.73                           & 3.75±1.12                          & 5.43±1.62                          & 5.20          \\
\textbf{France}  & 37 (9.4\%)     & \underline{7.51±2.43} & 6.50±4.55                              & 4.51±1.21                              & 7.16±3.23                           & 5.71±0.82                          & 4.63±1.77                          & 4.70          \\
\textbf{U.K.}    & 26 (6.6\%)     & 5.80±2.15                              & \underline{9.17±6.85} & 6.72±1.67                              & \underline{11.77±6.35} & \underline{5.72±4.86} & \underline{8.62±6.52} & 5.75          \\
\textbf{Mexico}  & 20 (4.6\%)     & 3.52±1.57                              & 7.37±1.01                              & 3.43±0.61                              & 3.60±0.55                           & 2.38±0.48                          & 3.72±1.68                          & 2.82          \\
\textbf{India}   & 18 (3.2\%)     & 4.39±1.09                              & 8.86±5.19                              & 3.08±1.17                              & 10.82±9.44                          & 3.04±1.56                          & 3.03±1.38                          & 7.29          \\
\textbf{Germany} & 13 (3.3\%)     & 7.39±2.52                              & \textbf{10.65±4.95}                    & 5.87±2.04                              & 9.53±4.72                           & 2.75±2.12                          & 7.67±3.29                          & 4.69          \\
\textbf{China}   & 13 (3.3\%)     & 2.07±0.53                              & 4.86±4.36                              & 2.43±0.93                              & 6.79±2.17                           & 2.75±0.70                          & 6.54±2.51                          & 4.58          \\
\textbf{Iran}    & 11 (2.8\%)     & \textbf{9.30±4.85}                     & 8.19±5.41                              & \underline{7.38±2.56} & 11.23±7.82                          & \textbf{6.31±2.85}                 & 6.67±5.02                          & 5.90          \\
\textbf{Greece}  & 11 (2.8\%)     & 3.12±0.98                              & 5.21±3.26                              & 4.25±1.23                              & 9.25±4.78                           & 3.13±1.09                          & 4.54±3.95                          & 4.63          \\
\textbf{Spain}   & 10 (2.5\%)     & 5.84±1.24                              & 4.13±2.32                              & \textbf{8.01±1.35}                     & \textbf{12.74±3.84}                 & 3.41±1.18                          & 6.90±4.08                          & 7.34          \\
\textbf{Russia}  & 10 (2.5\%)     & 3.62±0.80                              & 3.13±0.79                              & 2.48±0.26                              & 4.20±0.98                           & 4.88±2.78                          & 4.76±0.94                          & 4.34          \\ \hline
\textbf{ALL}     & 395 (100.0\%)  & 4.41±0.35                              & 5.80±2.56                              & 4.48±0.60                              & 8.24±3.65                           & 3.85±1.00                          & 5.85±1.47                          & 4.96          \\ \hline
\end{tabular}
}
    \caption{Performance results evaluated on \textbf{mAP} (\%).}
  \end{subtable}
  
  \medskip 
  
  \begin{subtable}{1\linewidth}
    \centering
    \scalebox{0.8}{
\begin{tabular}{@{}l|rrrrrr|r@{}}
\toprule
   \textbf{Origin}      & \textbf{Bb-ar}                           & \textbf{mB}                              & \textbf{mT5}                             & \textbf{Qwen2}                           & \textbf{Llama2}                          & \textbf{Llama3}                          & \textbf{Avg.} \\ \midrule
\textbf{Italy}   & 0.3280±0.01                              & 0.2259±0.02                              & 0.3547±0.05                              & 0.3213±0.06                              & 0.2252±0.05                              & 0.2802±0.05                              & 0.2892        \\
\textbf{U.S.}    & 0.3355±0.02                              & 0.2746±0.05                              & 0.3581±0.03                              & 0.3261±0.05                              & 0.2419±0.03                              & 0.3194±0.02                              & 0.3093        \\
\textbf{Turkey}  & 0.3141±0.01                              & 0.2434±0.04                              & 0.3154±0.03                              & 0.3111±0.06                              & 0.2210±0.03                              & 0.3164±0.06                              & 0.2869        \\
\textbf{Japan}   & 0.2876±0.02                              & 0.2411±0.04                              & 0.3150±0.05                              & 0.3030±0.05                              & 0.1834±0.04                              & 0.2553±0.04                              & 0.2642        \\
\textbf{France}  & \underline{0.3599±0.02} & 0.2984±0.04                              & 0.3462±0.03                              & 0.3129±0.04                              & 0.2608±0.02                              & 0.2844±0.01                              & 0.3104        \\
\textbf{U.K.}    & 0.3450±0.02                              & \underline{0.3118±0.08} & \underline{0.3651±0.04} & 0.3450±0.06                              & 0.2406±0.07                              & \underline{0.3435±0.05} & 0.3252        \\
\textbf{Mexico}  & 0.3098±0.02                              & 0.2891±0.04                              & 0.3335±0.07                              & 0.2687±0.03                              & 0.2163±0.02                              & 0.2665±0.02                              & 0.2807        \\
\textbf{India}   & 0.2850±0.03                              & 0.2474±0.07                              & 0.3051±0.03                              & 0.3355±0.09                              & 0.2217±0.05                              & 0.2455±0.03                              & 0.2734        \\
\textbf{Germany} & 0.3300±0.02                              & 0.2859±0.07                              & 0.3530±0.04                              & 0.2772±0.06                              & 0.2046±0.04                              & 0.3172±0.05                              & 0.2946        \\
\textbf{China}   & 0.2938±0.02                              & 0.2836±0.06                              & 0.3255±0.07                              & 0.2889±0.03                              & 0.1907±0.04                              & 0.2774±0.04                              & 0.2767        \\
\textbf{Iran}    & 0.3453±0.04                              & 0.2644±0.06                              & 0.3634±0.03                              & \underline{0.3488±0.04} & 0.2405±0.06                              & 0.3257±0.02                              & 0.3147        \\
\textbf{Greece}  & 0.3241±0.02                              & 0.2559±0.04                              & 0.3502±0.05                              & 0.3358±0.06                              & 0.2325±0.03                              & 0.3052±0.04                              & 0.3006        \\
\textbf{Spain}   & \textbf{0.3659±0.02}    & \textbf{0.3213±0.03}    & \textbf{0.4071±0.06}    & \textbf{0.3626±0.06}    & \underline{0.2869±0.05} & \textbf{0.3576±0.05}    & 0.3502        \\
\textbf{Russia}  & 0.3323±0.02                              & 0.2772±0.02                              & 0.3347±0.02                              & 0.3451±0.02                              & \textbf{0.2964±0.03}    & 0.3145±0.00                              & 0.3167        \\ \midrule
\textbf{ALL}     & 0.3243±0.01                              & 0.2625±0.03                              & 0.3421±0.04                              & 0.3176±0.05                              & 0.2282±0.04                              & 0.2957±0.03                              & 0.2951        \\ \bottomrule
\end{tabular}
}
    \caption{Performance results evaluated on \textbf{mWS}.}
  \end{subtable}
  
  \caption{Probing performance comparison with \textbf{Arabic} prompts and \textbf{\textsc{FmLAMA}-\textit{ar}} sub-dataset. 
  }
  \label{tab:ar_without}
\end{table*}

\begin{table*}[t]
\centering
\scalebox{0.8}{
\begin{tabular}{@{}lr|rrrrrr|r@{}}
\toprule
\textbf{Origin}  & \textbf{Count} & \textbf{Bb-zh}                          & \textbf{mB}                             & \textbf{mT5}                            & \textbf{Qwen2}                          & \textbf{Llama2}                         & \textbf{Llama3}                         & \textbf{Avg.} \\ \midrule
\textbf{Italy}   & 49 (8.7\%)     & 17.18±1.67                              & 13.07±4.92                              & 10.55±0.84                              & 24.54±2.87                              & 9.32±4.48                               & 13.58±4.08                              & 14.71         \\
\textbf{U.S.}    & 101 (17.9\%)   & 19.32±1.47                              & 13.79±3.56                              & \underline{12.37±1.09} & 26.50±2.94                              & 12.91±3.61                              & 13.17±6.45                              & 16.34         \\
\textbf{Turkey}  & 12 (2.1\%)     & 8.92±3.86                               & 6.30±2.12                               & 5.22±0.18                               & 18.16±3.49                              & 8.18±3.58                               & 11.00±6.39                              & 9.63          \\
\textbf{Japan}   & 114 (20.2\%)   & 15.79±2.32                              & 11.18±4.41                              & 7.85±0.60                               & 17.67±2.73                              & 11.15±2.66                              & 11.78±4.54                              & 12.57         \\
\textbf{France}  & 70 (12.4\%)    & 19.45±4.13                              & 16.04±5.31                              & 9.14±0.98                               & 20.96±1.45                              & 13.36±3.62                              & 14.71±5.89                              & 15.61         \\
\textbf{U.K.}    & 38 (6.7\%)     & \textbf{27.88±2.65}    & \textbf{19.55±7.36}    & 10.64±0.68                              & \underline{28.43±4.16} & \textbf{18.02±8.33}    & \textbf{21.46±6.75}    & 21.00         \\
\textbf{Mexico}  & 19 (3.4\%)     & 21.12±0.86                              & 13.40±3.51                              & 12.05±0.79                              & 16.25±4.70                              & 10.60±1.32                              & 8.21±3.83                               & 13.61         \\
\textbf{India}   & 24 (4.3\%)     & 5.47±1.65                               & 5.02±2.68                               & 1.13±0.17                               & 12.37±3.79                              & 4.57±2.20                               & 6.58±5.13                               & 5.86          \\
\textbf{Germany} & 16 (2.8\%)     & 10.31±2.53                              & 12.46±3.54                              & 6.81±1.75                               & 10.41±2.12                              & 4.08±2.38                               & 7.62±4.63                               & 8.62          \\
\textbf{China}   & 87 (15.4\%)    & \underline{22.55±4.86} & \underline{18.32±6.75} & 7.88±0.91                               & \textbf{33.48±6.15}    & \underline{13.83±3.72} & \underline{19.96±9.42} & 19.34         \\
\textbf{Iran}    & 7 (1.2\%)      & 7.72±1.44                               & 8.19±4.58                               & 3.54±0.20                               & 8.39±2.08                               & 3.40±0.38                               & 8.82±3.50                               & 6.68          \\
\textbf{Greece}  & 7 (1.2\%)      & 4.11±0.68                               & 2.88±1.08                               & 5.43±0.43                               & 14.81±3.05                              & 1.40±0.37                               & 3.46±1.23                               & 5.35          \\
\textbf{Spain}   & 12 (2.1\%)     & 13.00±1.83                              & 10.21±1.58                              & \textbf{12.72±1.17}    & 21.02±0.93                              & 5.53±3.77                               & 12.41±6.05                              & 12.48         \\
\textbf{Russia}  & 8 (1.4\%)      & 5.58±0.93                               & 3.98±0.76                               & 2.29±0.38                               & 5.89±1.09                               & 4.06±2.84                               & 5.14±1.07                               & 4.49          \\ \midrule
\textbf{ALL}     & 564 (100\%)    & 17.84±2.37                              & 13.56±4.36                              & 8.96±0.44                               & 22.71±2.37                              & 11.46±3.32                              & 13.76±5.56                              & 14.71         \\ \bottomrule
\end{tabular}}
\caption{Probing performance comparison with \textbf{Chinese} prompts and \textbf{\textsc{FmLAMA}-\textit{zh}} sub-dataset. 
}
\label{tab:zh_main_results}
\end{table*}

\begin{table*}[t]
  \begin{subtable}{1\linewidth}
    \centering
    \scalebox{0.8}{
    \begin{tabular}{@{}lr|rrrrr|r@{}}
\toprule
\textbf{Origin}  & \textbf{Count} & \textbf{mB}                            & \textbf{mT5}                           & \textbf{Qwen2}                          & \textbf{Llama2}                        & \textbf{Llama3}                         & \textbf{Average} \\ \midrule
\textbf{Italy}   & 59 (18.6\%)    & 4.18±2.54                              & 4.30±1.49                              & 7.78±2.37                               & 3.97±1.47                              & 6.29±3.14                               & 5.30             \\
\textbf{U.S.}    & 67 (21.1\%)    & 8.62±7.22                              & \underline{8.13±2.02} & 9.09±4.52                               & 3.62±0.81                              & 10.17±4.44                              & 7.93             \\
\textbf{Turkey}  & 12 (3.9\%)     & \underline{8.99±8.65} & 6.33±4.01                              & 9.21±5.92                               & 2.93±1.18                              & \underline{10.68±6.33} & 7.63             \\
\textbf{Japan}   & 19 (6.0\%)     & 6.69±5.23                              & 5.99±3.64                              & 6.60±4.19                               & \underline{6.92±3.85} & 4.39±3.12                               & 6.12             \\
\textbf{France}  & 65 (20.5\%)    & 5.43±3.01                              & 4.54±0.92                              & 6.40±2.74                               & 2.72±0.54                              & 5.50±1.47                               & 4.92             \\
\textbf{U.K.}    & 23 (7.3\%)     & \textbf{9.43±6.79}    & 6.95±2.26                              & 9.79±5.06                               & 3.97±2.49                              & 10.08±5.66                              & 8.04             \\
\textbf{Mexico}  & 11 (3.5\%)     & 2.22±0.75                              & 2.26±0.74                              & 4.33±2.44                               & 2.87±1.99                              & 2.14±2.01                               & 2.76             \\
\textbf{India}   & 18 (5.7\%)     & 8.12±3.72                              & 3.31±1.07                              & 2.80±2.77                               & 3.26±1.95                              & 4.73±2.46                               & 4.44             \\
\textbf{Germany} & 11 (3.5\%)     & 7.57±5.98                              & \textbf{14.84±6.23}   & 8.59±3.46                               & 5.92±1.83                              & 9.38±5.42                               & 9.26             \\
\textbf{China}   & 9 (2.8\%)      & 3.61±3.34                              & 2.12±0.78                              & 3.75±3.62                               & 1.64±1.29                              & 4.05±1.62                               & 3.03             \\
\textbf{Iran}    & 6 (1.9\%)      & 8.58±5.37                              & 5.05±2.90                              & \textbf{19.40±9.90}    & 2.95±0.57                              & \textbf{20.63±10.96}   & 11.32            \\
\textbf{Greece}  & 6 (1.9\%)      & 2.76±0.60                              & 4.31±0.49                              & \underline{12.96±9.55} & \textbf{7.67±2.85}    & 3.34±2.19                               & 6.21             \\
\textbf{Spain}   & 7 (2.2\%)      & 4.58±1.81                              & 7.02±1.87                              & 3.86±1.66                               & 4.04±1.62                              & 2.82±1.26                               & 4.46             \\
\textbf{Russia}  & 4 (1.3\%)      & 4.31±3.65                              & 3.35±0.07                              & 2.10±0.28                               & 2.56±0.46                              & 2.56±0.44                               & 2.98             \\ \midrule
\textbf{ALL}     & 317 (100\%)    & 6.41±4.30                              & 5.77±1.59                              & 7.57±2.66                               & 3.74±0.85                              & 7.17±2.63                               & 6.13             \\ \bottomrule
\end{tabular}
}
    \caption{Performance results evaluated on \textbf{mAP} (\%).}
  \end{subtable}
  
  \medskip 
  
  \begin{subtable}{1\linewidth}
    \centering
    \scalebox{0.8}{
\begin{tabular}{@{}l|rrrrr|r@{}}
\toprule
\textbf{Origin}  & \textbf{mB}                              & \textbf{mT5}                             & \textbf{Qwen2}                           & \textbf{Llama2}                          & \textbf{Llama3}                          & \textbf{Average} \\ \midrule
\textbf{Italy}   & 0.3102±0.03                              & 0.3320±0.03                              & 0.3197±0.03                              & 0.2973±0.04                              & 0.3265±0.05                              & 0.3171           \\
\textbf{U.S.}    & 0.3435±0.06                              & 0.3626±0.03                              & 0.3524±0.03                              & 0.3042±0.03                              & 0.3517±0.04                              & 0.3429           \\
\textbf{Turkey}  & 0.2976±0.06                              & 0.3225±0.05                              & 0.3110±0.04                              & 0.2621±0.05                              & 0.3296±0.10                              & 0.3046           \\
\textbf{Japan}   & 0.3134±0.05                              & 0.3167±0.04                              & 0.3114±0.05                              & 0.2839±0.03                              & 0.2899±0.02                              & 0.3031           \\
\textbf{France}  & 0.3082±0.03                              & 0.3509±0.04                              & 0.3334±0.02                              & 0.2838±0.04                              & 0.3134±0.03                              & 0.3179           \\
\textbf{U.K.}    & 0.3400±0.06                              & 0.3418±0.04                              & 0.3414±0.04                              & 0.2930±0.02                              & 0.3167±0.06                              & 0.3266           \\
\textbf{Mexico}  & 0.2959±0.05                              & \underline\{0.3829±0.06\} & 0.3420±0.06                              & 0.2847±0.06                              & 0.2865±0.04                              & 0.3184           \\
\textbf{India}   & \textbf\{0.3614±0.05\}    & 0.3437±0.06                              & 0.3354±0.01                              & 0.3133±0.04                              & 0.3192±0.05                              & 0.3346           \\
\textbf{Germany} & \underline\{0.3595±0.04\} & 0.3763±0.02                              & \underline\{0.3756±0.03\} & \underline\{0.3339±0.05\} & \underline\{0.3586±0.06\} & 0.3608           \\
\textbf{China}   & 0.2863±0.03                              & 0.3131±0.06                              & 0.2714±0.06                              & 0.2683±0.03                              & 0.2432±0.04                              & 0.2765           \\
\textbf{Iran}    & 0.3435±0.05                              & 0.3521±0.07                              & \textbf\{0.4569±0.07\}    & 0.2902±0.02                              & \textbf\{0.4416±0.09\}    & 0.3769           \\
\textbf{Greece}  & 0.3014±0.02                              & 0.3423±0.02                              & 0.3722±0.08                              & 0.3332±0.03                              & 0.2951±0.04                              & 0.3288           \\
\textbf{Spain}   & 0.3493±0.03                              & \textbf\{0.4229±0.07\}    & 0.3547±0.02                              & \textbf\{0.3635±0.06\}    & 0.3260±0.03                              & 0.3633           \\
\textbf{Russia}  & 0.3276±0.03                              & 0.3748±0.06                              & 0.3319±0.04                              & 0.3308±0.01                              & 0.3179±0.04                              & 0.3366           \\ \midrule
\textbf{ALL}     & 0.3237±0.04                              & 0.3484±0.03                              & 0.3369±0.02                              & 0.2969±0.03                              & 0.3248±0.04                              & 0.3261           \\ \bottomrule
\end{tabular}
    
}
    \caption{Performance results evaluated on \textbf{mWS}.}
  \end{subtable}
  
  \caption{Probing performance comparison with \textbf{Hebrew} prompts and \textbf{\textsc{FmLAMA}-\textit{he}} sub-dataset. 
  }
  \label{tab:he_without}
\end{table*}

\begin{table*}[t]
\centering
\scalebox{0.8}{
\begin{tabular}{@{}lr|rrrrrrr|r@{}}
\toprule
\textbf{Origin}  & \textbf{Count} & \textbf{Bb-ky}                          & \textbf{Bb-kl}                          & \textbf{mB-u}                          & \textbf{mT5}                           & \textbf{Qwen2}                         & \textbf{Llama2}                        & \textbf{Llama3}                        & \textbf{Avg.} \\ \midrule
\textbf{Italy}   & 88 (16.7\%)    & 6.90±1.57                               & 7.75±2.34                               & 3.85±2.54                              & 6.02±0.64                              & 3.50±1.10                              & 1.19±0.22                              & 2.23±0.75                              & 4.49          \\
\textbf{U.S.}    & 64 (12.2\%)    & \underline{11.37±2.66} & \underline{10.87±3.86} & \textbf{6.32±5.20}    & \textbf{10.15±0.34}   & 2.46±0.69                              & 1.68±0.12                              & 2.95±0.93                              & 6.54          \\
\textbf{Turkey}  & 19 (3.6\%)     & 2.75±2.47                               & 4.13±2.62                               & 4.07±3.17                              & 1.82±0.44                              & 3.86±1.00                              & 2.16±0.12                              & \textbf{7.47±1.45}    & 3.75          \\
\textbf{Japan}   & 101 (19.2\%)   & 7.34±2.12                               & 7.74±1.86                               & 1.93±1.24                              & 5.25±0.99                              & 6.56±5.34                              & 2.19±0.22                              & 2.62±1.15                              & 4.80          \\
\textbf{France}  & 67 (12.7\%)    & 2.33±0.43                               & 3.09±0.44                               & 3.15±0.17                              & 2.79±0.30                              & 3.10±0.68                              & 1.40±0.09                              & 2.41±1.27                              & 2.61          \\
\textbf{U.K.}    & 27 (5.1\%)     & \textbf{14.07±2.83}    & \textbf{15.89±4.43}    & \underline{6.00±4.42} & \underline{7.55±1.34} & 3.21±1.02                              & 1.79±0.31                              & \underline{6.89±5.70} & 7.91          \\
\textbf{Mexico}  & 13 (2.5\%)     & 5.18±3.69                               & 8.37±2.85                               & 4.89±3.29                              & 2.31±0.40                              & \underline{7.12±3.13} & 2.75±1.99                              & 4.18±1.61                              & 4.97          \\
\textbf{India}   & 32 (6.1\%)     & 7.98±2.29                               & 5.74±1.32                               & 3.28±2.09                              & 2.96±0.61                              & \textbf{7.69±6.42}    & 2.01±0.32                              & 2.36±0.67                              & 4.57          \\
\textbf{Germany} & 15 (2.9\%)     & 2.50±0.94                               & 5.08±1.25                               & 2.42±1.68                              & 3.20±1.18                              & 2.63±0.95                              & 1.33±0.48                              & 1.91±0.21                              & 2.72          \\
\textbf{China}   & 54 (10.3\%)    & 9.02±2.76                               & 10.12±3.04                              & 2.36±2.22                              & 4.05±1.50                              & 4.43±1.74                              & 2.16±0.48                              & 3.06±1.48                              & 5.03          \\
\textbf{Iran}    & 8 (1.5\%)      & 7.99±5.23                               & 3.76±1.41                               & 1.62±0.67                              & 1.52±0.24                              & 1.50±0.14                              & \underline{2.87±2.49} & 2.12±1.83                              & 3.05          \\
\textbf{Greece}  & 9 (1.7\%)      & 2.32±0.73                               & 6.76±3.73                               & 2.74±0.76                              & 5.76±4.22                              & 4.59±1.31                              & \textbf{3.67±2.94}    & 3.35±1.87                              & 4.17          \\
\textbf{Spain}   & 21 (4.0\%)     & 4.05±2.25                               & 5.90±2.70                               & 1.79±1.24                              & 5.82±1.33                              & 3.99±1.38                              & 2.50±0.97                              & 3.85±0.99                              & 3.99          \\
\textbf{Russia}  & 8 (1.5\%)      & 3.70±0.99                               & 4.22±1.83                               & 2.68±1.12                              & 2.12±0.05                              & 3.01±1.41                              & 1.72±0.20                              & 3.07±1.46                              & 2.93          \\ \midrule
\textbf{ALL}     & 526 (100\%)    & 7.05±1.47                               & 7.68±2.00                               & 3.45±2.10                              & 5.19±0.42                              & 4.32±1.76                              & 1.86±0.18                              & 3.07±0.95                              & 4.66          \\ \bottomrule
\end{tabular}}
\caption{Probing performance comparison on \textbf{mAP} (\%) with \textbf{Korean} prompts and \textbf{\textsc{FmLAMA}-\textit{ko}} sub-dataset. 
}
\label{tab:ko_without}
\end{table*}

\begin{table*}[t]
  \begin{subtable}{1\linewidth}
    \centering
    \scalebox{0.8}{
   \begin{tabular}{@{}lr|rrrrrr|r@{}}
\toprule
\textbf{Origin}  & \textbf{Count} & \textbf{Bb-ru}                          & \textbf{mB}                             & \textbf{mT5}                           & \textbf{Qwen2}                          & \textbf{Llama2}                         & \textbf{Llama3}                         & \textbf{Avg.} \\ \midrule
\textbf{Italy}   & 101 (15.9\%)   & 4.29±2.04                               & 6.57±3.51                               & 4.52±1.27                              & 7.99±2.61                               & 4.30±2.32                               & 10.61±2.60                              & 6.38          \\
\textbf{U.S.}    & 79 (12.5\%)    & 9.76±4.95                               & 9.99±5.25                               & 5.51±0.71                              & 12.01±3.53                              & 9.04±4.33                               & 11.97±3.75                              & 9.71          \\
\textbf{Turkey}  & 28 (4.4\%)     & 5.75±2.12                               & 9.46±1.68                               & 3.58±1.61                              & 11.08±8.01                              & 10.49±2.90                              & 16.37±2.97                              & 9.46          \\
\textbf{Japan}   & 68 (10.7\%)    & 7.89±4.76                               & 7.13±2.34                               & 1.76±0.56                              & 6.10±2.74                               & 8.50±5.79                               & 14.11±0.74                              & 7.58          \\
\textbf{France}  & 93 (14.7\%)    & 7.08±3.04                               & 6.83±2.85                               & 5.03±1.86                              & 7.03±2.48                               & 8.85±4.96                               & 10.55±3.43                              & 7.56          \\
\textbf{U.K.}    & 33 (5.2\%)     & 8.42±5.82                               & 10.03±5.01                              & \underline{5.54±3.74} & 11.89±6.16                              & \textbf{13.17±8.89}    & 16.82±8.31                              & 10.98         \\
\textbf{Mexico}  & 22 (3.5\%)     & 4.08±0.89                               & 2.01±0.82                               & 2.08±0.95                              & 4.70±1.40                               & 5.64±3.25                               & 4.21±1.34                               & 3.79          \\
\textbf{India}   & 21 (3.3\%)     & \textbf{11.46±9.78}    & \textbf{19.02±2.71}    & 4.27±3.11                              & \underline{16.80±6.73} & \underline{12.95±5.70} & \underline{22.93±2.60} & 14.57         \\
\textbf{Germany} & 45 (7.1\%)     & 7.53±5.26                               & 7.95±3.59                               & 4.39±2.03                              & 9.41±2.54                               & 11.74±4.92                              & 11.57±4.32                              & 8.77          \\
\textbf{China}   & 30 (4.7\%)     & 9.67±4.64                               & \underline{11.86±3.63} & 3.28±2.19                              & 10.44±4.98                              & 12.63±6.07                              & 12.82±4.43                              & 10.12         \\
\textbf{Iran}    & 13 (2.1\%)     & \underline{10.86±6.84} & 11.06±5.99                              & \textbf{15.51±4.53}   & \textbf{22.88±6.61}    & 10.77±8.30                              & \textbf{23.88±6.87}    & 15.83         \\
\textbf{Greece}  & 18 (2.8\%)     & 3.45±2.42                               & 1.74±0.43                               & 3.38±1.24                              & 8.56±2.57                               & 4.91±2.85                               & 4.77±1.28                               & 4.47          \\
\textbf{Spain}   & 38 (6.0\%)     & 3.35±1.51                               & 6.51±2.61                               & 3.37±1.37                              & 5.27±2.52                               & 3.88±2.07                               & 5.61±1.84                               & 4.67          \\
\textbf{Russia}  & 45 (7.1\%)     & 4.79±2.05                               & 3.75±0.77                               & 2.80±0.92                              & 5.21±2.94                               & 5.31±2.18                               & 5.47±2.27                               & 4.55          \\ \midrule
\textbf{ALL}     & 634 (100\%)    & 6.85±3.14                               & 7.76±1.83                               & 4.28±1.11                              & 8.84±2.67                               & 8.19±4.04                               & 11.52±2.82                              & 7.91          \\ \bottomrule
\end{tabular}
    
}
    \caption{Performance results evaluated on \textbf{mAP} (\%).}
  \end{subtable}
  
  \medskip 
  
  \begin{subtable}{1\linewidth}
    \centering
    \scalebox{0.8}{
    \begin{tabular}{@{}l|rrrrrr|r@{}}
\toprule
\textbf{Origin}  & \textbf{Bb-ru}                           & \textbf{mB}                              & \textbf{mT5}                             & \textbf{Qwen2}                           & \textbf{Llama2}                          & \textbf{Llama3}                          & \textbf{Avg.} \\ \midrule
\textbf{Italy}   & 0.3319±0.04                              & 0.3217±0.06                              & 0.3633±0.03                              & 0.4073±0.04                              & 0.3774±0.02                              & 0.3779±0.05                              & 0.3633        \\
\textbf{U.S.}    & \textbf{0.3796±0.04}    & \underline{0.3556±0.05} & 0.3692±0.02                              & \underline{0.4393±0.03} & 0.3938±0.05                              & 0.4068±0.05                              & 0.3907        \\
\textbf{Turkey}  & 0.3199±0.06                              & 0.3317±0.04                              & 0.3300±0.04                              & 0.3966±0.06                              & 0.4012±0.02                              & 0.4178±0.04                              & 0.3662        \\
\textbf{Japan}   & 0.2955±0.03                              & 0.2796±0.04                              & 0.2740±0.02                              & 0.3085±0.04                              & 0.3110±0.05                              & 0.3319±0.01                              & 0.3001        \\
\textbf{France}  & 0.3438±0.05                              & 0.3132±0.03                              & 0.3652±0.04                              & 0.3875±0.02                              & 0.3954±0.04                              & 0.3752±0.04                              & 0.3634        \\
\textbf{U.K.}    & \underline{0.3647±0.08} & 0.3442±0.05                              & 0.3771±0.04                              & 0.4327±0.04                              & \underline{0.4149±0.08} & 0.4179±0.09                              & 0.3919        \\
\textbf{Mexico}  & 0.3150±0.01                              & 0.3081±0.04                              & 0.3604±0.04                              & 0.3688±0.03                              & 0.3909±0.02                              & 0.3225±0.04                              & 0.3443        \\
\textbf{India}   & 0.3475±0.10                              & \textbf{0.3808±0.05}    & 0.3062±0.03                              & 0.4375±0.06                              & 0.3665±0.07                              & \underline{0.4236±0.04} & 0.3770        \\
\textbf{Germany} & 0.3500±0.08                              & 0.3330±0.05                              & 0.3402±0.04                              & 0.3968±0.03                              & 0.4053±0.04                              & 0.3743±0.07                              & 0.3666        \\
\textbf{China}   & 0.3318±0.04                              & 0.3348±0.05                              & 0.3276±0.03                              & 0.3741±0.07                              & 0.3819±0.06                              & 0.3520±0.05                              & 0.3504        \\
\textbf{Iran}    & 0.3252±0.06                              & 0.3358±0.03                              & \textbf{0.4221±0.06}    & \textbf{0.4916±0.06}    & \textbf{0.4418±0.04}    & \textbf{0.4818±0.09}    & 0.4164        \\
\textbf{Greece}  & 0.3360±0.05                              & 0.3015±0.04                              & 0.3573±0.03                              & 0.4045±0.01                              & 0.4012±0.03                              & 0.3365±0.05                              & 0.3562        \\
\textbf{Spain}   & 0.3240±0.03                              & 0.3283±0.04                              & 0.3625±0.04                              & 0.3880±0.04                              & 0.3638±0.03                              & 0.3313±0.04                              & 0.3497        \\
\textbf{Russia}  & 0.3583±0.03                              & 0.3199±0.04                              & \underline{0.3899±0.03} & 0.3732±0.04                              & 0.3907±0.01                              & 0.3591±0.03                              & 0.3652        \\ \midrule
\textbf{ALL}     & 0.3395±0.04                              & 0.3247±0.04                              & 0.3515±0.03                              & 0.3940±0.03                              & 0.3824±0.03                              & 0.3750±0.04                              & 0.3612        \\ \bottomrule
\end{tabular}
}
    \caption{Performance results evaluated on \textbf{mWS}.}
  \end{subtable}
  
  \caption{Probing performance comparison with \textbf{Russian} prompts and \textbf{\textsc{FmLAMA}-\textit{ru}} sub-dataset. 
  }
  \label{tab:ru_without}
\end{table*}

\subsection{Language analysis on other multilingual LLMs}
\label{app:language}
\begin{figure*}[t]
    \centering
    \includegraphics[width=0.8\linewidth]{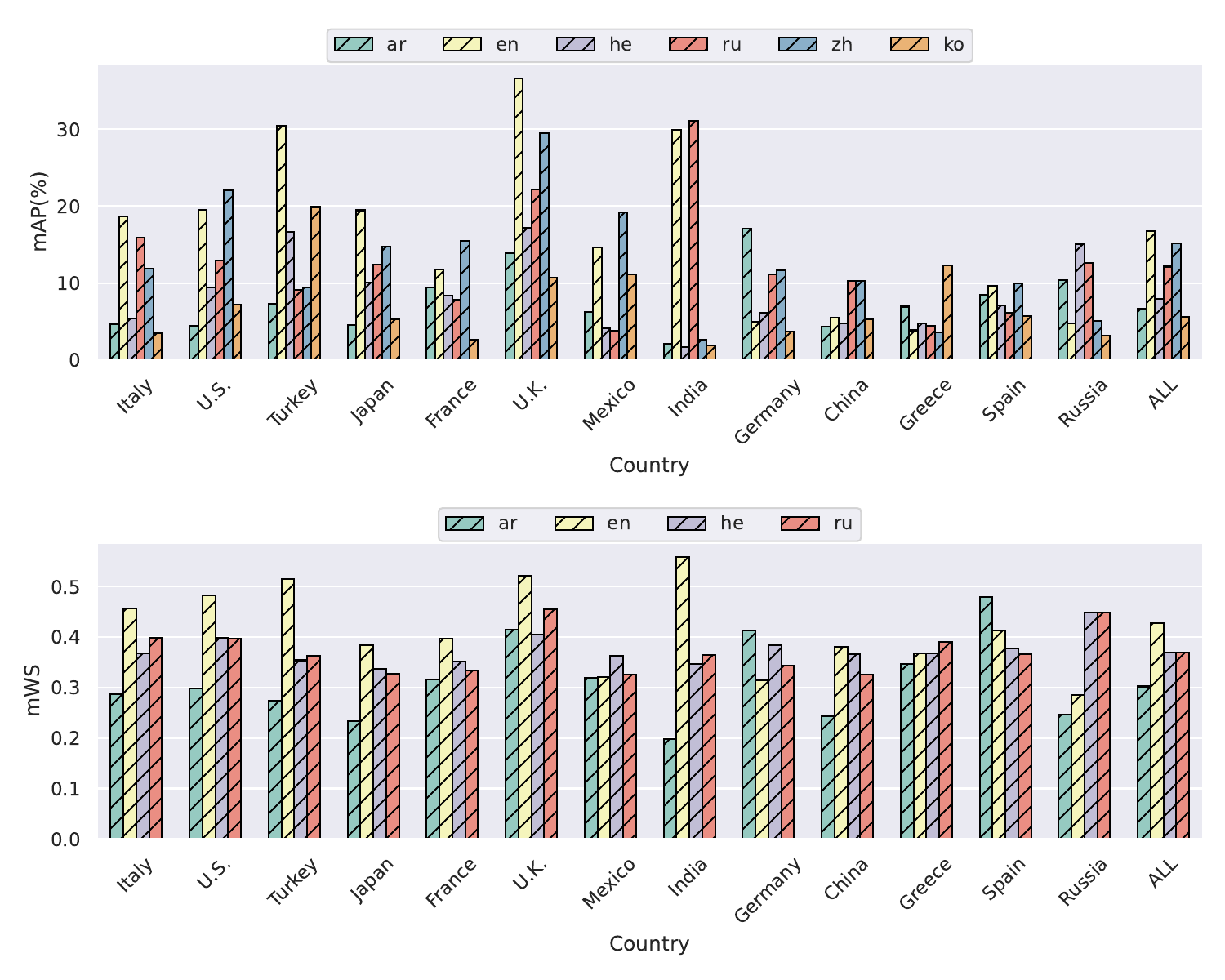}
    \caption{Comparison of probing results on mBERT with prompts in different languages,}
    \label{fig:mB}
\end{figure*}

\begin{figure*}[t]
    \centering
    \includegraphics[width=0.8\linewidth]{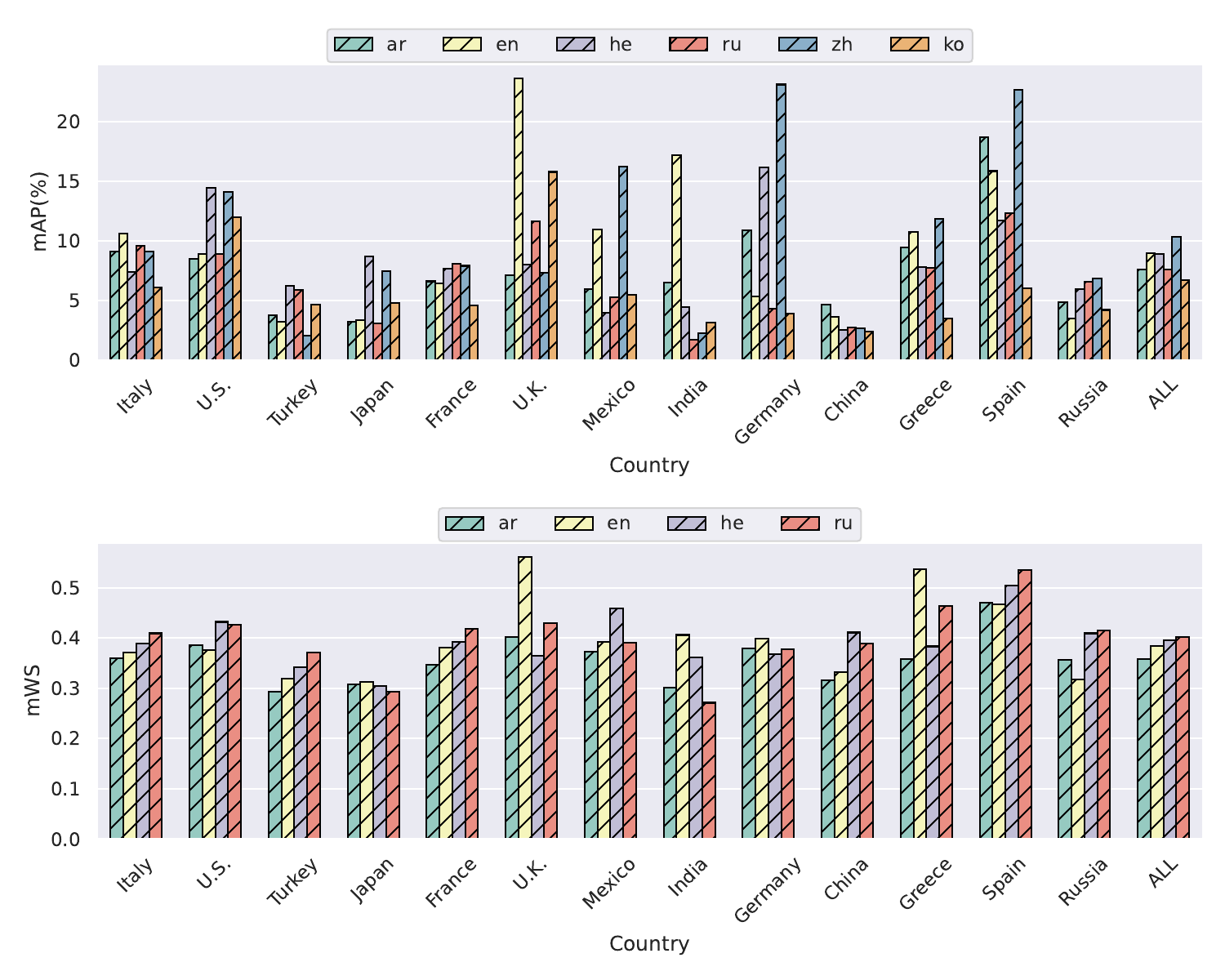}
    \caption{Comparison of probing results on mT5 with prompts in different languages,}
    \label{fig:mT5}
\end{figure*}

\begin{figure*}[t]
    \centering
    \includegraphics[width=0.8\linewidth]{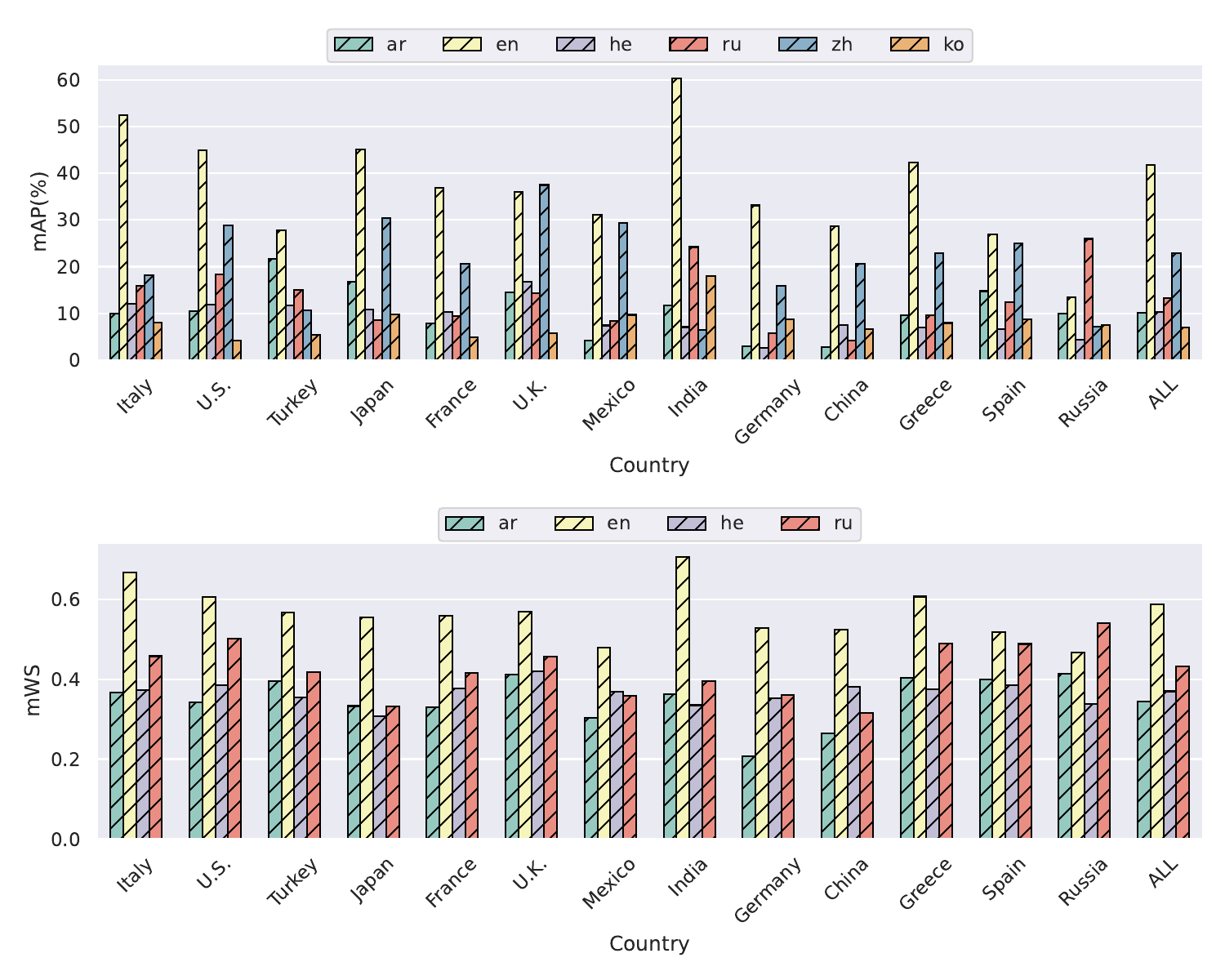}
    \caption{Comparison of probing results on Qwen2 with prompts in different languages,}
    \label{fig:Qwen2}
\end{figure*}

\begin{figure*}[t]
    \centering
    \includegraphics[width=0.8\linewidth]{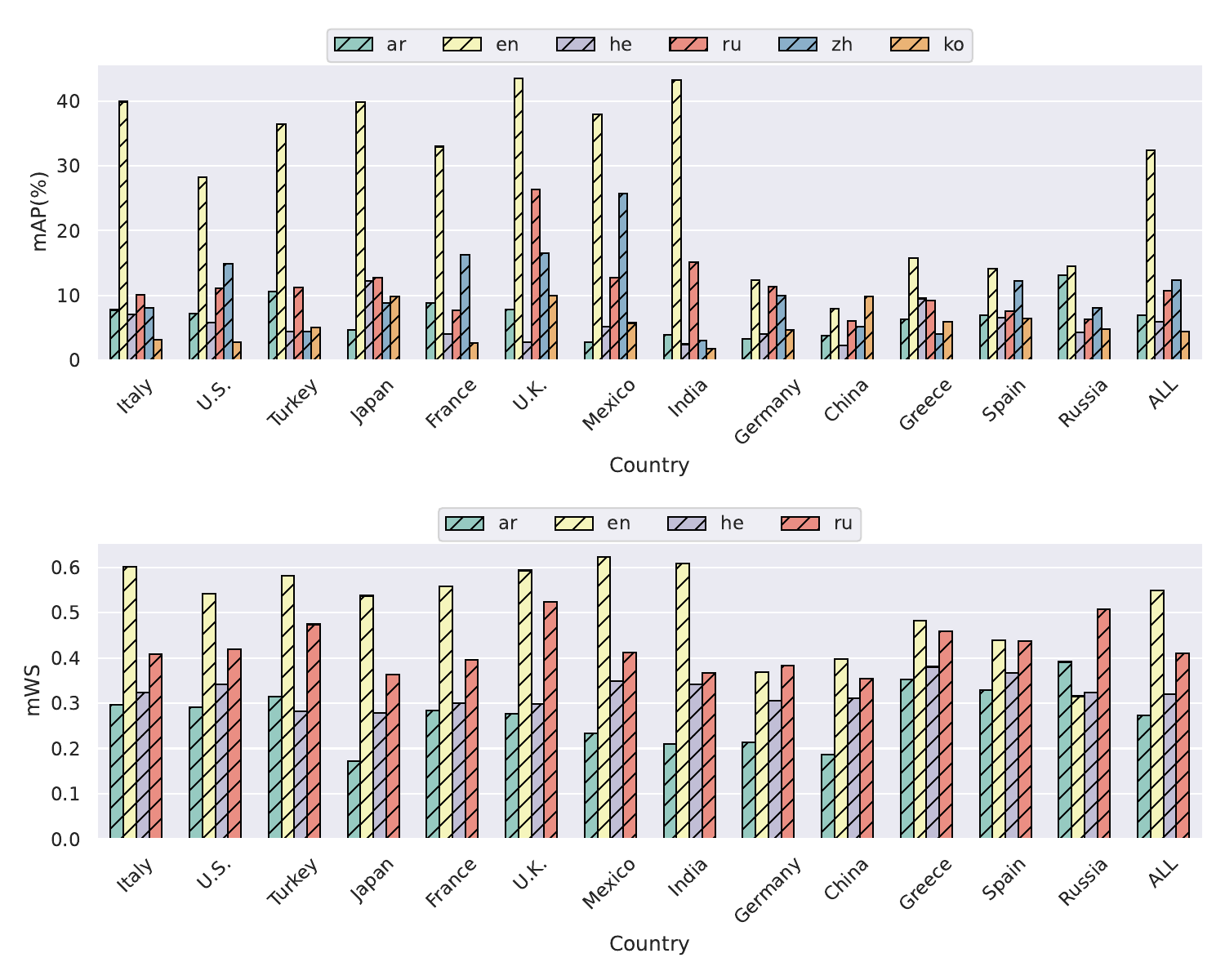}
    \caption{Comparison of probing results on Llama2 with prompts in different languages,}
    \label{fig:Llama2}
\end{figure*}

Figures~\ref{fig:mB},~\ref{fig:mT5},~\ref{fig:Qwen2}, and~\ref{fig:Llama2} present a comparison of probing results for the multilingual LLMs—mBERT, mT5, Qwen2, and Llama2—using prompts in various languages on the corresponding filtered sub-datasets.
Decoder-only LLMs, such as Qwen2, Llama2, and Llama3, while representing the cutting edge of current LLM technology, exhibit greater performance variation under multilingual prompt settings.

\subsection{Code-switching probing results in each cultural group}
\label{app:cs}
\begin{figure*}[t]
    \centering
    \includegraphics[width=1.0\linewidth]{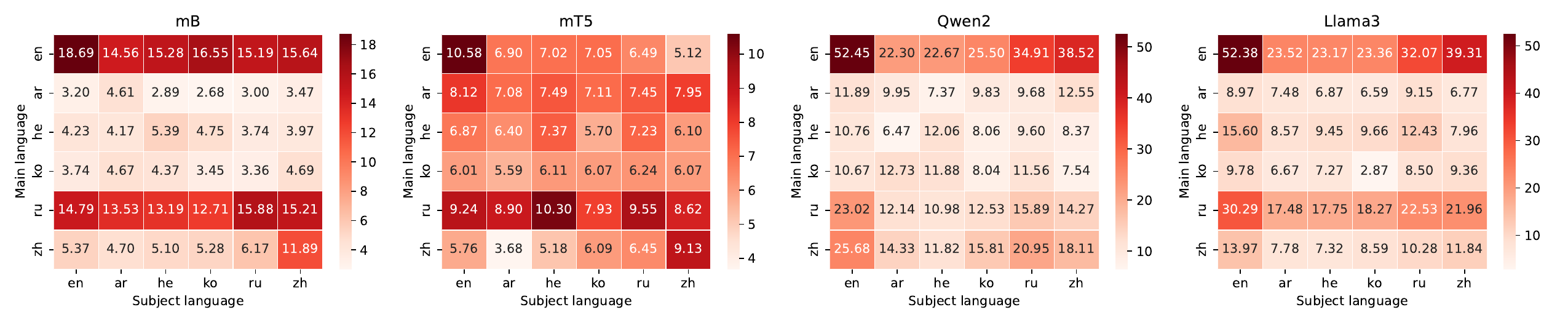}
    \caption{\textbf{Italy}: Probing results of code-switching prompts.}
    \label{fig:cs_Italy}
\end{figure*}

\begin{figure*}[t]
    \centering
    \includegraphics[width=1.0\linewidth]{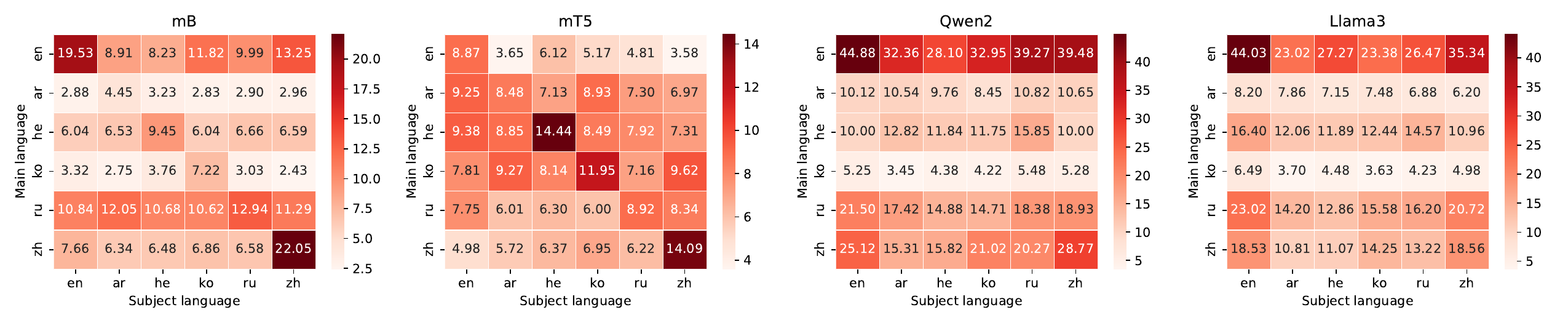}
    \caption{\textbf{U.S.}: Probing results of code-switching prompts.}
    \label{fig:cs_US}
\end{figure*}

\begin{figure*}[t]
    \centering
    \includegraphics[width=1.0\linewidth]{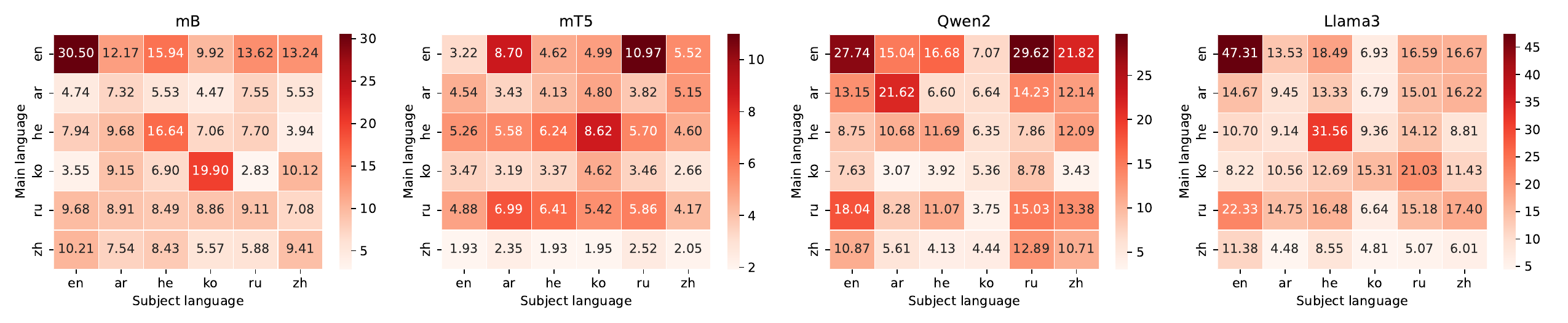}
    \caption{\textbf{Turkey}: Probing results of code-switching prompts.}
    \label{fig:cs_Turkey}
\end{figure*}

\begin{figure*}[t]
    \centering
    \includegraphics[width=1.0\linewidth]{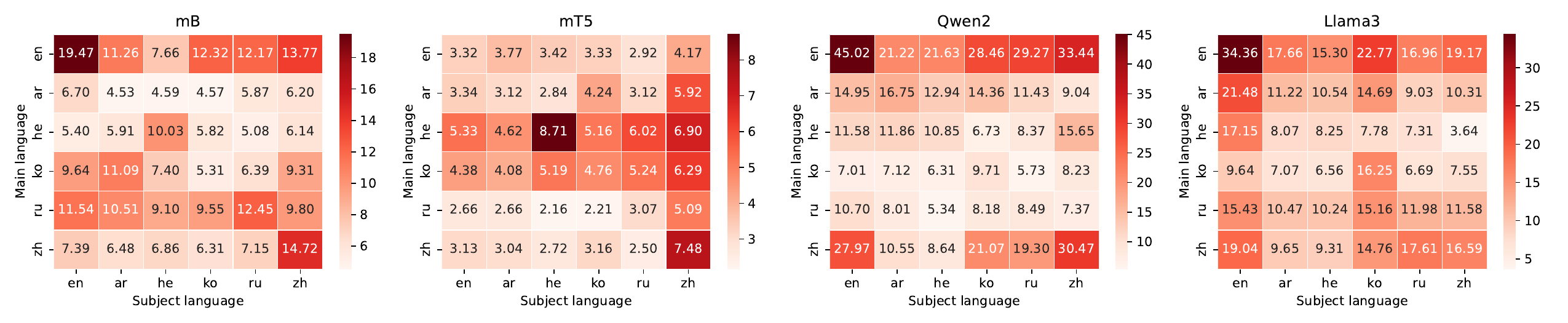}
    \caption{\textbf{Japan}: Probing results of code-switching prompts.}
    \label{fig:cs_Japan}
\end{figure*}

\begin{figure*}[t]
    \centering
    \includegraphics[width=1.0\linewidth]{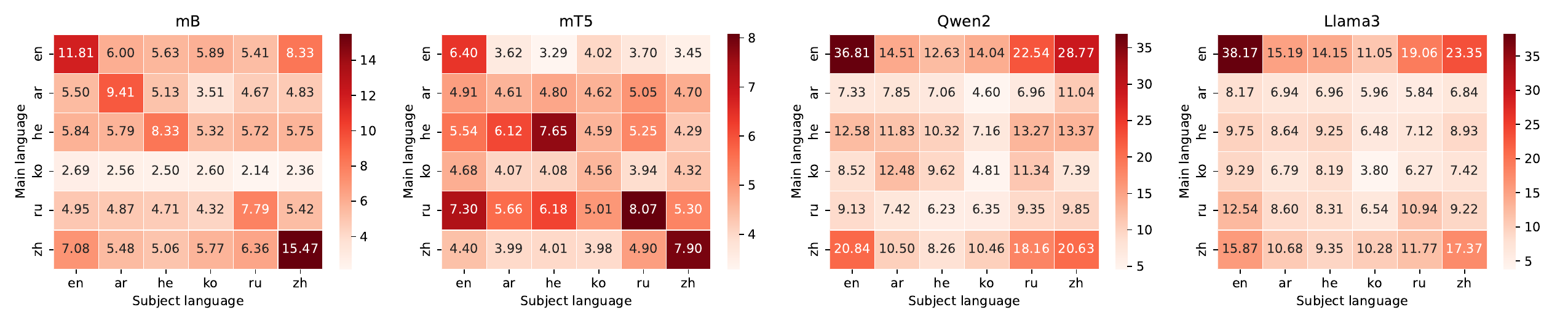}
    \caption{\textbf{France}: Probing results of code-switching prompts.}
    \label{fig:cs_France}
\end{figure*}

\begin{figure*}[t]
    \centering
    \includegraphics[width=1.0\linewidth]{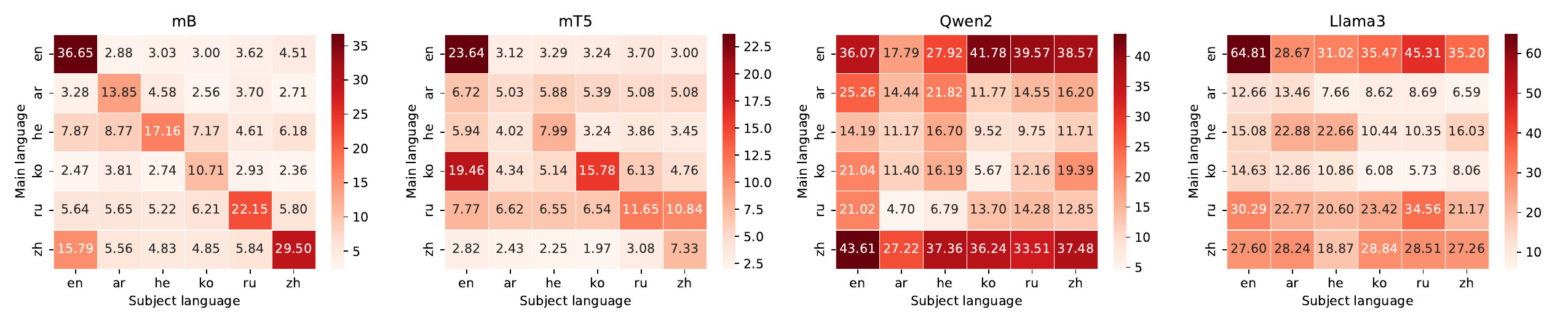}
    \caption{\textbf{U.K.}: Probing results of code-switching prompts.}
    \label{fig:cs_UK}
\end{figure*}

\begin{figure*}[t]
    \centering
    \includegraphics[width=1.0\linewidth]{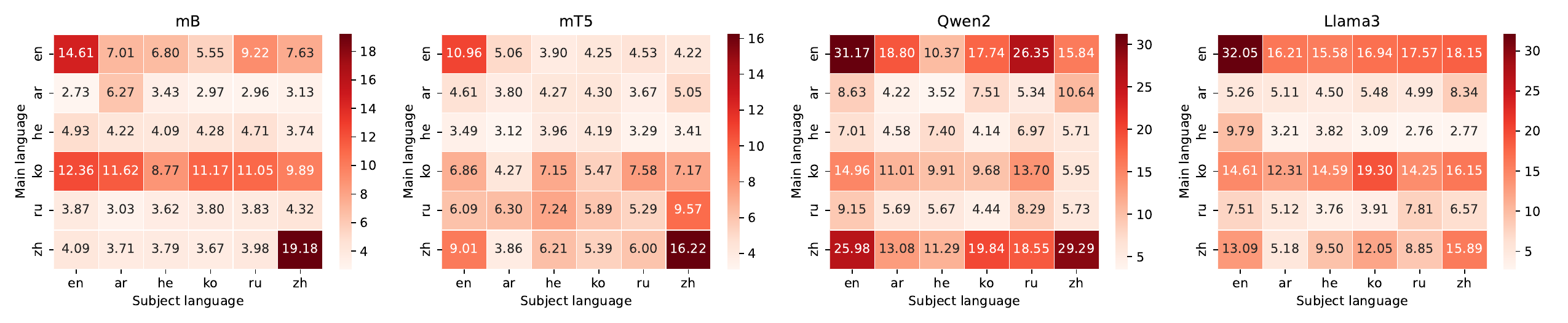}
    \caption{\textbf{Mexico}: Probing results of code-switching prompts.}
    \label{fig:cs_Mexico}
\end{figure*}

\begin{figure*}[t]
    \centering
    \includegraphics[width=1.0\linewidth]{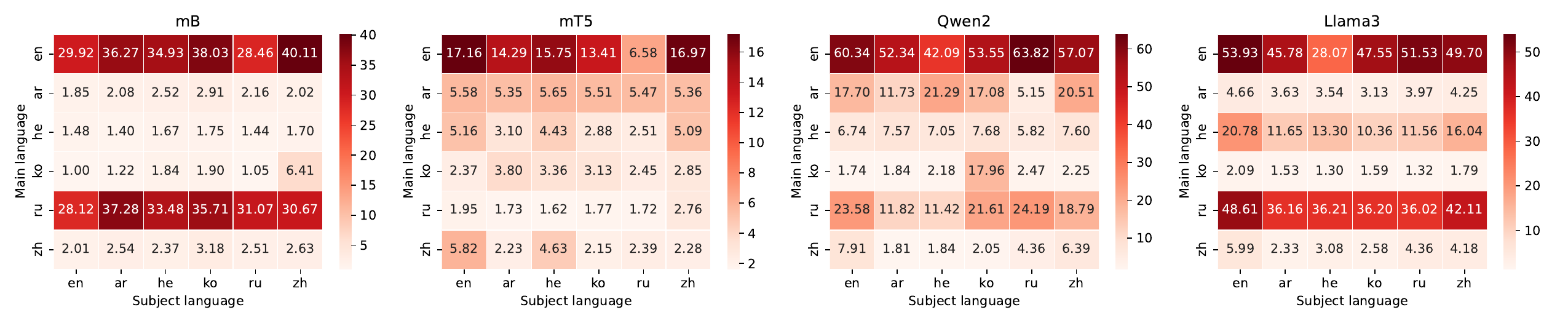}
    \caption{\textbf{India}: Probing results of code-switching prompts.}
    \label{fig:cs_India}
\end{figure*}

\begin{figure*}[t]
    \centering
    \includegraphics[width=1.0\linewidth]{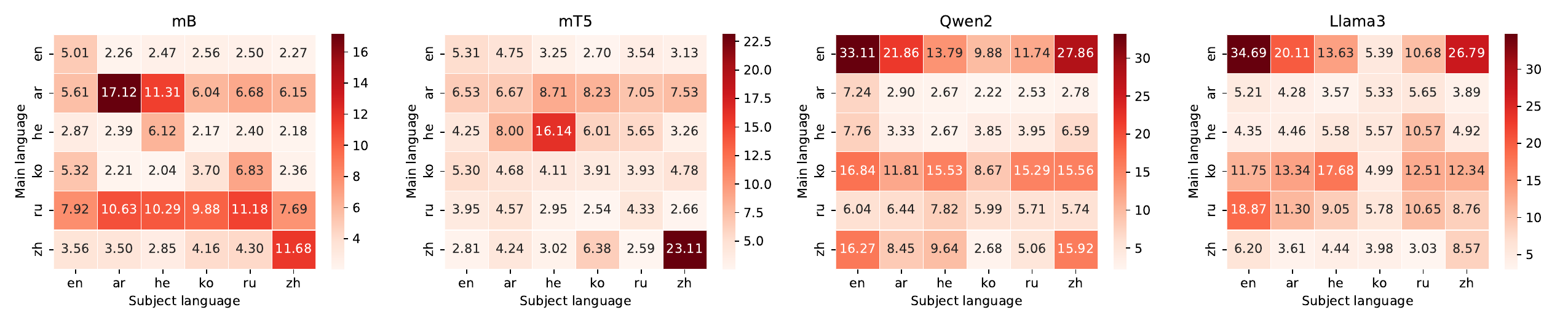}
    \caption{\textbf{Germany}: Probing results of code-switching prompts.}
    \label{fig:cs_Germany}
\end{figure*}

\begin{figure*}[t]
    \centering
    \includegraphics[width=1.0\linewidth]{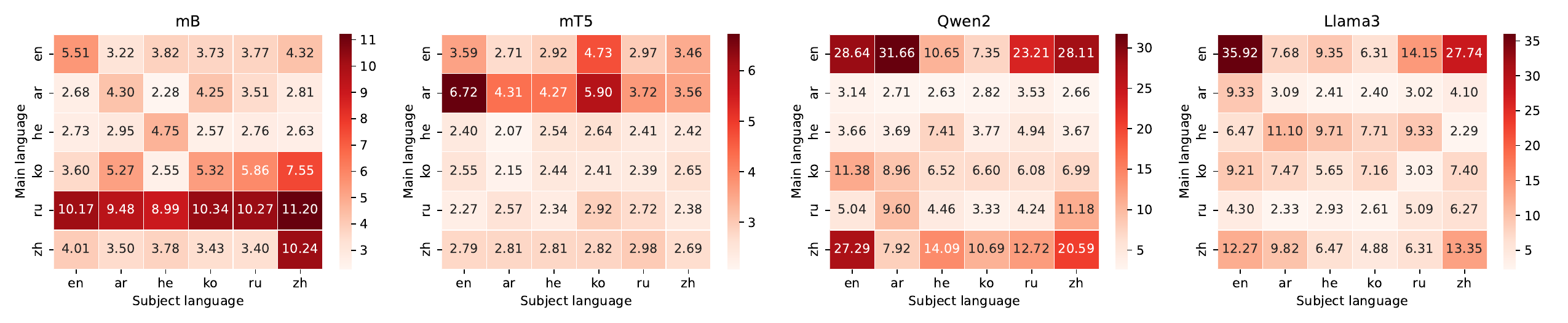}
    \caption{\textbf{China}: Probing results of code-switching prompts.}
    \label{fig:cs_China}
\end{figure*}

\begin{figure*}[t]
    \centering
    \includegraphics[width=1.0\linewidth]{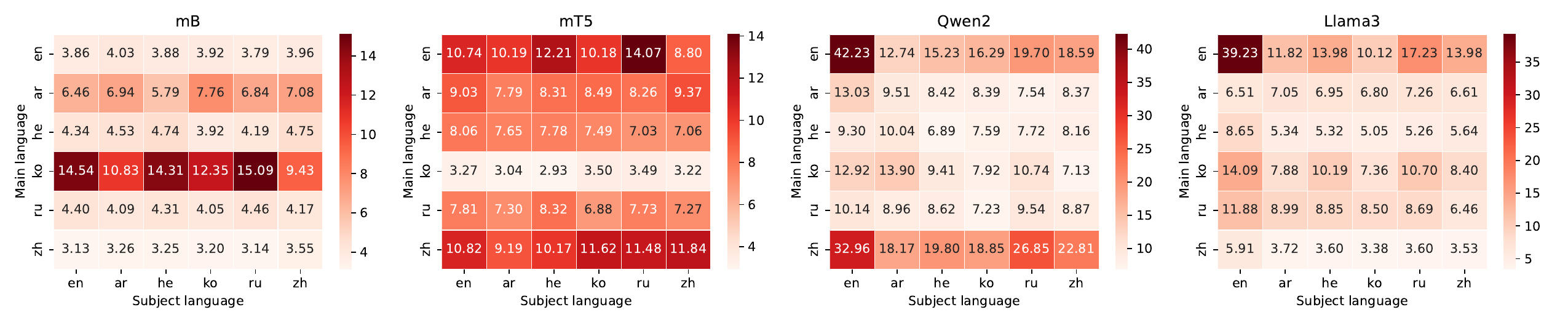}
    \caption{\textbf{Greece}: Probing results of code-switching prompts.}
    \label{fig:cs_Greece}
\end{figure*}

\begin{figure*}[t]
    \centering
    \includegraphics[width=1.0\linewidth]{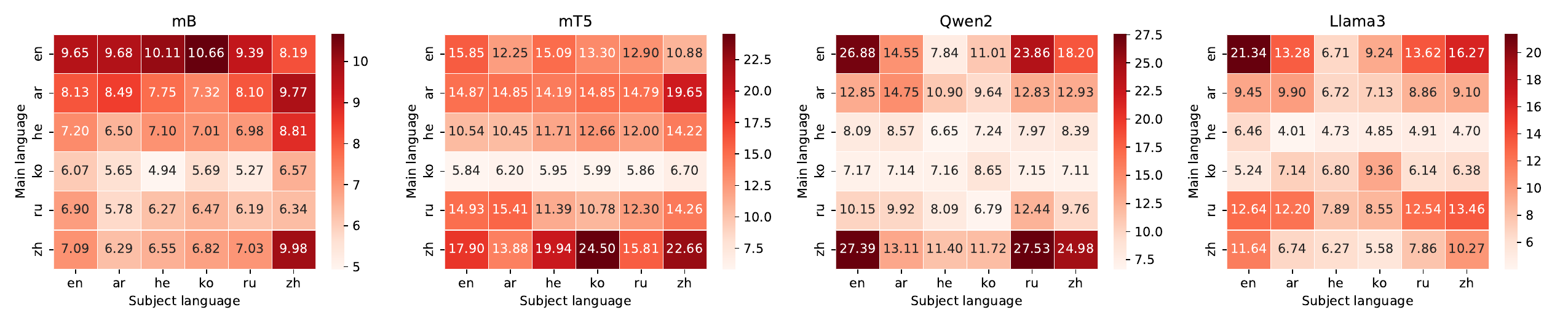}
    \caption{\textbf{Spain}: Probing results of code-switching prompts.}
    \label{fig:cs_Spain}
\end{figure*}

\begin{figure*}[t]
    \centering
    \includegraphics[width=1.0\linewidth]{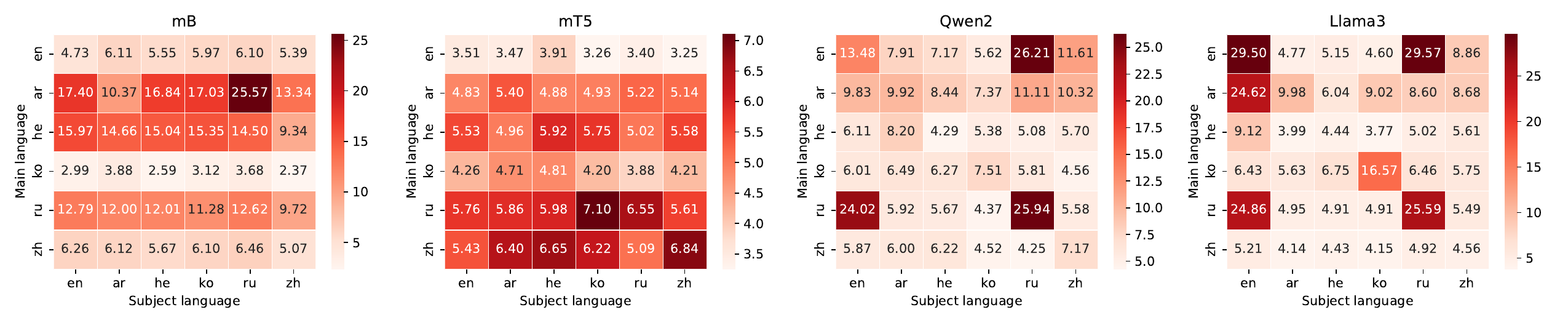}
    \caption{\textbf{Russia}: Probing results of code-switching prompts.}
    \label{fig:cs_Russia}
\end{figure*}

\begin{figure*}[t]
    \centering
    \includegraphics[width=1.0\linewidth]{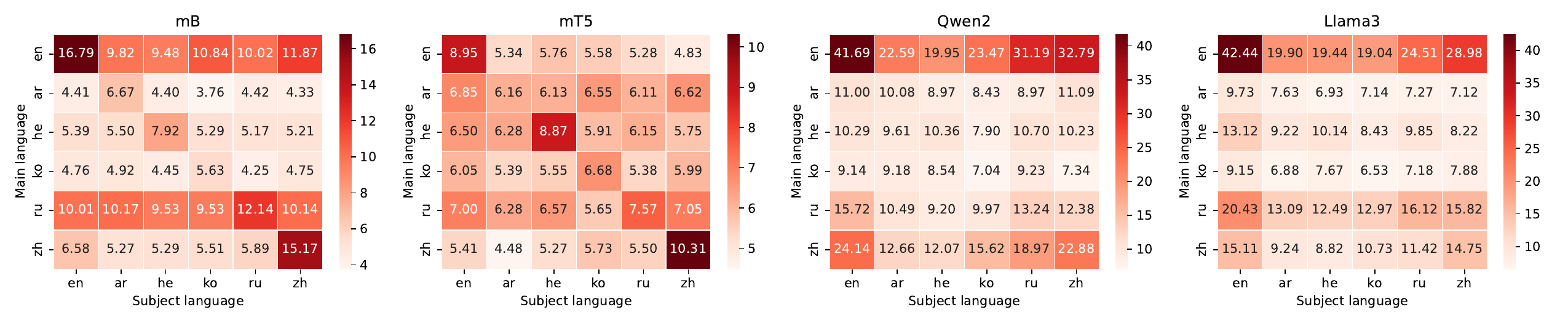}
    \caption{\textbf{ALL}: Probing results of code-switching prompts.}
    \label{fig:cs_ALL}
\end{figure*}

Figure~\ref{fig:cs_Italy}-~\ref{fig:cs_ALL} presents a detailed comparison of the probing results for each cultural group and other multilingual LLMs.

\section{Cultural bias \textit{vs} Dish popularity }
\label{app:Iran}
Table~\ref{tab:en_without_main} shows that some LLMs perform better in probing tasks for low-resource language groups. Specifically, the decoder-only LLM performs well in the Iran group (both mAP and mWS metric), while Greece shows decent results with both T5 and decoder-only LLMs (only mWS metric).

To understand the strong performance of the Iran group across different metrics, we analyzed the results for the 21 dishes. Table~\ref{tab:Iran_ingredients} shows 17 involved dishes where the predicted ingredients ranked in the top 10 (Llama3). Since the top-ranked predicted ingredients in the table are not all commonly used, we can rule out the possibility that the strong performance of the Iran group is due to the LLM's common ingredient prediction errors. We hypothesize two possible reasons for this performance:
(1) LLM Cultural Ability: Decoder-only LLMs may be more familiar with the Iran cultural group.
(2) Data Bias: The dishes in the Iran group are likely well-known and representative of Iranian cuisine, so the LLM might show stronger memory performance for familiar entities~\cite{du-etal-2024-context}.

\begin{table*}[ht]
\centering
\begin{tabular}{@{}llll@{}}
\toprule
\textbf{ID} & \textbf{Subject (dish)} & \textbf{Involved Object (ingredients)} & \textbf{Rank} \\ \midrule
0           & pomegranate soup        & pomegranate                            & 0             \\
1           & Albaloo polo            & rice                                   & 1             \\
2           & Kofta                   & minced meat                            & 0             \\
3           & Sabzi polo              & herb                                   & 5             \\
4           & falooda                 & milk, vermicelli                       & 1, 0          \\
5           & ghormeh sabzi           & parsley, herb                          & 0, 4          \\
6           & Bastani                 & milk, sugar, egg                       & 8, 7, 6       \\
7           & zeytoon parvardeh       & olive                                  & 8             \\
8           & rock candy              & sugar                                  & 0             \\
9           & Faloodeh                & vermicelli                             & 0             \\
10          & muhallebi               & milk, rice flour                       & 2, 1          \\
11          & rogan josh              & lamb meat                              & 1             \\
12          & ash-e doogh             & yogurt                                 & 0             \\
13          & Mahyawa                 & fish                                   & 4             \\
14          & sholezard               & rice                                   & 1             \\
15          & Eggplant Caviar         & eggplant                               & 0             \\
16          & nân-e panjere           & flour                                  & 6             \\ \bottomrule
\end{tabular}
\caption{Top 17 dishes from the Iran group with ingredients ranked in the top 10 by Llama3, excluding common prediction errors.}
\label{tab:Iran_ingredients}
\end{table*}

\end{document}